\documentclass[12pt]{article}
\usepackage{amsmath}
\usepackage{graphicx,psfrag,epsf}
\usepackage{enumerate}
\usepackage[usenames,dvipsnames]{color}
\usepackage{natbib}
\usepackage{url} 
\usepackage{bbm}
\usepackage[hyperfootnotes=false]{hyperref}
\usepackage{bm}
\usepackage{amsmath,amssymb,amsthm}
\usepackage{algpseudocode}
\usepackage{amsthm}
\usepackage[ruled,vlined,linesnumbered]{algorithm2e}
\usepackage{color}
\usepackage{float} 
\newtheorem{theorem}{Theorem}[section]
\newtheorem{lemma}[theorem]{Lemma}

\newtheorem{definition}[theorem]{Definition}
\newtheorem{condition}[theorem]{Condition}
\newtheorem{proposition}[theorem]{Proposition}

\newcommand{\R}{\mathbb{R}}                     

\newcommand{\bbeta}{\bm{\beta}}

\def\E{\mathbb{E}}

\def\bbeta{{\boldsymbol{\beta}}}

\newcommand{\Pro}{{\mathbb{P}}}

\hypersetup{
 colorlinks=false,
 citecolor=Violet,
 linkcolor=Red,
 urlcolor=Blue}

\newcommand{\blind}{0}

\begin{document}

\def\spacingset#1{\renewcommand{\baselinestretch}%
{#1}\small\normalsize} \spacingset{1}


\if0\blind
{
  \title{\bf High-Dimensional Differentially-Private EM Algorithm: Methods and Near-Optimal Statistical Guarantees}
  \author{Zhe Zhang and Linjun Zhang\\
    Department of Statistics, Rutgers University\\}
  \maketitle
} \fi

\if1\blind
{
  \bigskip
  \bigskip
  \bigskip
  \begin{center}
    {\LARGE\bf High-Dimensional Differentially-Private EM Algorithm: Methods and Near Optimal Statistical Guarantees}
\end{center}
  \medskip
} \fi

\bigskip
\begin{abstract}
In this paper, we develop a general framework to design differentially private expectation-maximization (EM) algorithms in high-dimensional latent variable models,  based on the noisy iterative hard-thresholding. We derive the statistical guarantees of the proposed framework and apply it to three specific models: Gaussian mixture, mixture of regression, and regression with missing covariates. In each model, we establish the near-optimal rate of convergence with differential privacy constraints, and show the proposed algorithm is minimax rate optimal up to logarithm factors. The technical tools developed for the high-dimensional setting are then extended to the classic low-dimensional latent variable models, and we propose a near rate-optimal EM algorithm with differential privacy guarantees in this setting. Simulation studies and real data analysis are conducted to support our results.
\end{abstract}

\noindent%
\vfill

\newpage
\spacingset{1.4} 

\section{Introduction}\label{sec:intro}
 In the era of big data, there is an unprecedented number of large data sets becoming available for researchers/industries to retrieve important information. In the meantime, these large data sets often include some sensitive personal information, urgently generating demands for privacy-preserving algorithms that can protect the individual information during the data analysis. One widely adopted criterion for privacy-preserving algorithms is differential privacy (DP) \citep{dwork2006calibrating,dwork2006our}.  This notion has been widely developed and used nowadays in Microsoft \citep{erlingsson2014rappor}, Google \citep{ding2017collecting}, Facebook \citep{kifer2020guidelines} and the U.S. Census Bureau \citep{abowd2016challenge} to help protect individual privacy including user habits, browsing history, social connections, and health records. {The basic idea behind differential privacy is that the information of a single individual in the training data should appear as hidden so that given the outcome, it is almost impossible to distinguish if a certain individual is in the dataset.}   
 
The attractiveness of differential privacy  could  possibly be attributed to the ease of building privacy-preserving algorithms. Many traditional algorithms and statistical methods have been extended to their  private counterparts, including top-$k$ selection \citep{bafna2017price, steinke2017tight}, multiple testing \citep{dwork2018differentially}, decision trees \citep{jagannathan2009practical}, and random forests \citep{rana2015differentially}. However, many existing works only focus on designing privacy-preserving algorithms while lacking the sharp analysis of accuracy in terms of minimax optimality. 
At the high-level, these privatized algorithms are designed by injecting random noise into the traditional algorithms. Such noise-injection procedures typically sacrifice statistical accuracy, and therefore, it is essential to understand what is the best accuracy an algorithm would output while maintaining a certain level of differential privacy requirement. 
This motivates us to study 
the trade-off between the privacy and the statistical accuracy. 

This paper is devoted to studying this trade-off in latent variable models by proposing a DP version of the expectation-maximization (EM) algorithm. The EM algorithm is a commonly used approach when dealing with latent variables and missing values \cite{ranjan2016robust, quost2016clustering, kadir2014high, ding2016algorithm}. The statistical properties, such as the local convergence and minimax optimality of standard EM algorithm has been recently studied in \cite{balakrishnan2017statistical,wang2015high,yi2015regularized,cai2019chime, zhao2020statistical}, while the development of DP EM algorithm, especially the theories of the optimal trade-off between privacy and accuracy, is still largely unexplored. In this paper, we propose novel DP EM algorithms in both  the classic low-dimensional setting and the contemporary high-dimensional setting, where the {parameter we are interested in is sparse and its dimension is greatly larger than the sample size}. We demonstrate the superiority of the proposed algorithm by applying it to specific statistical models. Under these specific statistical models, the convergence rates of the proposed algorithm are found to be minimax optimal  with privacy constraints.
The main contributions of this paper are summarized in the following.
\begin{itemize} 
    \item We design a novel DP EM algorithm in the high-dimensional setting based on a noisy iterative hard-thresholding algorithm in the M-step. Such a step effectively enforces the sparsity of the attained estimator while maintaining differential privacy, and {allows us to establish sharp rate of convergence of the algorithm}.
    To the best of our knowledge, this algorithm is the first DP EM algorithm in high-dimensions with statistical guarantees. 
    \item We then apply the proposed DP EM algorithm to three common models in the high-dimensional setting: Gaussian mixture model, mixture of regression model, and regression with missing covariates. Under mild regularity conditions, we show that our algorithm obtains the minimax optimal rate of convergence  with privacy constraints. 
    \item In the classical low-dimensional setting, a DP EM algorithm based on the Gaussian mechanism is designed.  The technical tools developed for the high-dimensional setting are then used to establish the optimality in this general low-dimensional setting. We show that both theoretically and numerically, the proposed algorithm outperforms several existing private EM algorithms in this classical setting.
\end{itemize}
\textbf{Related Work.} 
The expectation-maximization (EM) algorithm, proposed in \citep{dempster1977maximum}, is a common approach in handling latent variable models. There has been a long history on the study of the EM algorithm \citep{wu1983convergence, mclachlan2007algorithm}. Recently, a seminal work \citep{balakrishnan2017statistical} obtains a general framework to study the statistical guarantees of EM algorithms  in the classic low-dimensional setting. In subsequent works, the convergence rates of EM algorithm under various hidden variable models are studied, including the Gaussian mixture \citep{xu2016global, daskalakis2017ten, yan2017convergence, kwon2020algorithm} and mixture of linear regression \citep{yi2014alternating, kwon2019global, kwon2020minimax}. Another important direction is to design variants of EM algorithms in the high-dimensional regime. Such a goal is fulfilled through regularization \citep{cai2019chime, yi2015regularized, zhang2020estimation} and truncation \citep{wang2015high} in the M-step. Due to the increasing attention on data privacy nowadays, the design of private EM algorithms is in great demand but still largely lacking.

The differential privacy is arguably the most popular notion of privacy nowadays. After its invention, the basic properties has been  studied in \citep{dwork2010boosting,dwork2014algorithmic,dwork2017exposed,dwork2018privacy, mirshani2019formal}. The trade-off between statistical accuracy and privacy is one of the fundamental topics in differential privacy. In the low-dimensional setting, there are various works focusing on this trade-off, 
including mean estimation \citep{dwork2015robust, steinke2016between, bun2018fingerprinting, kamath2019privately, cai2019cost, kamath2020private}, confidence intervals of Gaussian mean  \citep{karwa2017finite} and binomial mean \citep{ awan2020differentially}, 
linear regression \citep{wang2018revisiting, cai2019cost}, generalized linear models \cite{song2020characterizing, cai2020cost, song2021evading}, 
 principal component analysis \citep{dwork2014analyze, chaudhuri2013near}, convex empirical risk minimization \citep{bassily2014private},  and robust M-estimators \citep{avella2020privacy,avella2021differentially}. 

However, in the high-dimensional setting where the dimension is much larger than the sample size, the trade-off between statistical accuracy and privacy is less studied. Most of the existing works focus on relatively standard statistical problems such as the sparse mean estimation and regression. For example, \citep{steinke2017tight} studies the optimal bounds for private top-$k$ selection
problems. \citep{cai2019cost} studies near-optimal algorithms for the sparse mean estimation. 
For the high-dimensional sparse linear regression, \citep{talwar2015nearly} obtains a DP LASSO algorithm which is near-optimal in terms of the excess risk; \citep{cai2019cost} proposes another DP LASSO algorithm with the optimal rate of convergence in estimation. In \citep{cai2020cost}, a near-minimax optimal DP algorithm for high-dimensional  generalized linear models is introduced. 

In the literature of differential privacy, most of the existing works for latent variable models focus on the low-dimensional cases, while the study in the high-dimensional regime is largely lacking. %
In the classical low-dimensional setting, \citep{park2017dp}  introduces a DP EM algorithm, but offers no statistical guarantees/accuracy analysis. 
\citep{nissim2007smooth} provides a result for the low-dimensional Gaussian mixture model based on the sample-aggregate framework and reaches a $O(\sqrt{d^{3}/{n}}\cdot{\log(1/\delta)}/{\epsilon})$  convergence rate in estimation for the mixture of spherical Gaussian distributions. 
\citep{kamath2020differentially} considers a more general Gaussian mixture models and studies the total variation distance. 
\citep{wang2020differentially} studies the DP EM algorithm in the classical low-dimensional setting and obtains the estimation error of order $O(\sqrt{d^{2}/{n}}\cdot{\log(1/\delta)}/{\epsilon})$. 
In the current paper, we are going to show this rate can be significantly improved to $O(\sqrt{d/n}+{d\cdot\sqrt{\log(1/\delta)}}/{n\epsilon})$, obtained by our proposed algorithm in the classic low-dimensional setting. 

\textbf{Organization of this paper.} This paper is organized as follows. In Section~\ref{sec:formulation}, we introduce the problem formulation as well as some preliminaries of the EM algorithm and differential privacy. In Section~\ref{sec:hdem}, we present the main results of this paper and establish the convergence rate of the proposed EM algorithm in high-dimensional settings. In Section~\ref{sec:hdemmodel}, we apply the results obtained in Section~\ref{sec:hdem} to three specific models: Gaussian mixture model, mixture of regression, and regression with missing covariates. We present the estimation error bounds for these three models respectively and show the optimality. In Section~\ref{sec:ldem}, we consider the DP EM algorithm in the classic low-dimensional setting. In Section~\ref{sec:simulation}, simulation studies of the proposed algorithm are conducted to support our theory. Section~\ref{sec:conclusion} summarizes the paper and discusses some possible future work.  In Appendix~\ref{sec:supp}, we provided some supplement materials. We prove the main results in Appendix~\ref{sec:proof}, and the proofs of other results and technical lemmas are in the Appendix~\ref{sec:appendix}. 

\textbf{Notations.} Throughout this paper, let $\bm v = (v_1, v_2,...v_d) ^\top \in \mathbb{R}^d$ be a vector. $\mathcal{S}$ denotes the set of indices and $\bm v_{\mathcal{S}}$ indicates the restriction of vector $\bm v$ on the set $\mathcal{S}$. Also, $\|\bm v\|_q$ denotes the $\ell_q$ norm for $1\le q\le \infty$ and $\|\bm v\|_0$ specifically denotes the number of non-zero coordinates of $\bm v$, and we also call it the sparsity level of $\bm v$. We denote $\odot$ to be the Hadamard product. Generally, we use $n$ to denote the number of samples, $d$ to denote the dimension of a vector and $s$ to denote the sparsity level of a vector. We also define a truncation function $\Pi_T : \mathbb{R}^d \rightarrow \mathbb{R}^d$ be a function denotes the projection into the $\ell_{\infty}$ ball of radius $T$ centered at the origin in $\mathbb{R}^d$. 
Moreover, we use $\nabla Q(\cdot;\cdot)$ to denote the gradient of the function $Q(\cdot;\cdot)$. If there is no further specification, this gradient is taken with respect to the first argument. 
For two sequences $\{a_n\}$ and $\{b_n\}$, we write $a_n = o(b_n)$ if $a_n/b_n \rightarrow 0$. We denote $a_n = O(b_n)$ if there exist a constant $c$ such that $a_n \le c b_n$ and $a_n = \Omega(b_n)$  if there exist a constant $c'$ such that $a_n \ge c' b_n$. We also denote $a_n \asymp b_n$ if $a_n= O(b_n)$ and $a_n = \Omega(b_n)$. In this paper, $c_0,c_1,m_0,m_1,C,C',K,K',...$ denote universal constants and their specific values may vary from place to place. 


\section{Problem Formulation}\label{sec:formulation}
 In this section, we present some preliminaries that are  important for the discussions in the rest of the paper. We are going to introduce the EM algorithm in Section~\ref{sec2-1}, and the formal definition and some critical properties of differential privacy in Section~\ref{sec2-2}. 
\subsection{The EM algorithm}\label{sec2-1}
The EM algorithm is a widely used algorithm to compute the maximum likelihood estimator when there are latent or unobserved variables in the model. We first introduce the standard EM algorithm. 
Let $\bm Y$ and $\bm Z$ be random variables taking values in the sample spaces $\mathcal{Y}$ and $\mathcal{Z}$. For each pair of data $(\bm Y,\bm Z)$,  we assume that only  $\bm Y$ is observed, while $\bm Z$ remains unobserved.  Suppose that the pair $(\bm Y, \bm Z)$ has a joint density function $f_{\bm \beta^*}(\bm y,\bm z)$, where $\bm \beta^*$ is the true parameter that we would like to estimate. Let $h_{\bm \beta^*}(\bm y)$ be the marginal density function of the observed variable $\bm Y$. Then, we can write $h_{\bm \beta^*}(\bm y)$ by integrating out $\bm z$
\begin{equation}\label{eq:model} 
h_{\bm \beta^*}(\bm y) = \int_{\mathcal{Z} }f_{\bm \beta^*}(\bm y, \bm z) d \bm z. 
\end{equation}
 
The goal of the EM algorithm is to obtain an estimator of $\bm \beta^*$ through maximizing the likelihood function. Specifically, suppose we have $n$ $i.i.d.$ observations of $\bm Y$: $ \bm y_1, \bm y_2,... \bm y_n\sim h_{\bbeta^*}(\bm y)$, we aim to use EM algorithm to maximize the log-likelihood $l_n(\bm \beta) = \sum_{i=1}^n \log h_{\bm \beta}(\bm y_i)$, and get an estimator of $\bbeta^*$. In many latent variable models, it is generally difficult to evaluate the log-likelihood $l_n(\bm \beta)$ directly, but relatively easy to compute the log-likelihood for the joint distribution $f_{\bm \beta}(\bm y, \bm z)$. Such situations are in need of EM algorithms. Specifically, for a given parameter $\bbeta$,  let $k_{\bm \beta}(\bm z| \bm y)$ be the conditional distribution of $\bm Z$ given the observed variable $\bm Y$, that is, $k_{\bm \beta}(\bm z| \bm y) = \frac{f_{\bm \beta}(\bm y, \bm z)}{h_{\bm \beta}( \bm y)}.$

The EM algorithm uses an iterative approach motivated by Jensen's inequality. Suppose that in the $t$-th iteration, we have obtained $\bbeta^t$ and would like to update it to $\bbeta^{t+1}$ with a larger log-likelihood. The log-likelihood evaluated at $\bm \beta^{t+1}$ can always be lower bounded, as shown in the following expression. 
\begin{align}
    \frac{1}{n}  l_n(\bm \beta^{t+1}) &= \frac{1}{n}\sum_{i=1}^n \log h_{\bm \beta^{t+1}}(\bm y_i)   \label{eq:em} \\ 
    \notag&\ge \underbrace{ \frac{1}{n}\sum_{i=1}^n\int_{\mathcal{Z}}
  k_{\bbeta^t}(\bm z_i| \bm y_i) \log f_{\bm \beta^{t+1}}(\bm y_i, \bm z_i)
  d \bm z_i}_{Q_n({\bm \beta^{t+1}} ; { \bbeta^t})} -  \frac{1}{n}\sum_{i=1}^n\int_{\mathcal{Z}} 
k_{ \bbeta^t}(\bm z_i |\bm y_i) \log k_{\bbeta^t}(\bm z_i
|\bm y_i) d \bm z_i,
\end{align}
with equality holds when $\bbeta^{t+1} = \bbeta^t$. The basic idea of EM algorithm is to successively maximize the lower bound with respect to $\bbeta^{t+1}$ in \eqref{eq:em}. In the E-step, we evaluate the lower bound in \eqref{eq:em} at the current parameter $\bbeta^t$. Since the second term in \eqref{eq:em} only depends on $\bm \beta^{t}$, which is fixed given the current $\bbeta^t$, we only need to evaluate $Q_n$  in the E-step. Then, in the M-Step, we calculate a new parameter $\bbeta^{t+1}$ which moves towards the direction that maximizes $Q_n$, so the lower bound in \eqref{eq:em} becomes larger when we update $\bbeta^t$ to $\bbeta^{t+1}$. We use the new parameter $\bbeta^{t+1}$ at the $(t+1)$-th iteration and continue the E-step and M-step iteratively until convergence.


In the $t$-th iteration, the M-step in the standard EM algorithm maximizes  $Q_n(\cdot ;\bm \beta^t)$ \citep{dempster1977maximum}, that is, $\bm \beta^{t+1} = \text{argmax}_{\bm \beta} Q_n(\bm \beta ;\bm \beta^t)$. However, sometimes it is computationally expensive to compute the maximizer directly. As an alternative, the gradient EM \citep{balakrishnan2017statistical} was proposed by taking one-step update of the gradient ascent $\bm \beta^{t+1} = \bm \beta^t + \eta \cdot \nabla Q_n(\bm \beta^t ; \bm \beta^t)$ in the M-step. When $Q_n(\bm \beta ; \bm \beta')$ is strongly convex with respect to $\bbeta$, this gradient EM approach is shown to reach the same statistical guarantee as that of the standard EM approach \citep{balakrishnan2017statistical}. 
%
In the high-dimensional setting, when the data dimension is much larger than the sample size and the true parameter is sparse, people have come up with different variants of EM algorithms. For example, in the M-step, the maximization approach is generalized to the regularized maximization \citep{yi2015regularized, cai2019chime, zhang2020estimation} and the gradient approach is generalized to the truncated gradient  \citep{wang2015high}.  
\subsection{Some basic properties of differential privacy}\label{sec2-2}

 In this section, we introduce the concepts and properties of differential privacy. These properties will play an important role in the design of the DP EM algorithm. First, the formal definition of differential privacy is given below.
\begin{definition}[Differential Privacy \citep{dwork2006calibrating}]
Let $\mathcal{X}$ be the sample space for an individual data, a randomized algorithm $M:\mathcal{X}^n\rightarrow\mathbb{R}$ is $(\epsilon, \delta)$-DP if and only if for every pair of adjacent data sets $ \bm X,  \bm X' \in \mathcal X^n$ and for any $S \subseteq  \R$, the inequality below holds:
\begin{align*}
	\mathbb{P}\left(M(\bm X) \in S\right) \leq e^\varepsilon \cdot \mathbb{P}\left(M( \bm X') \in S\right) + \delta,
\end{align*}
where we say that two data sets $\bm X = \{\bm x_i\}_{i=1}^n$ and $\bm X' = \{{{\bm x}_{i}'}\}_{i=1}^n$ are adjacent if and only if they differ by one individual datum.
\end{definition}
According to the definition, the two parameters $\epsilon,\delta$ control the privacy level. With smaller $\epsilon$ and $\delta$, the privacy constraint becomes more stringent. Intuitively speaking, this definition suggests that for a DP algorithm $M$, an adversary cannot distinguish if the original dataset is $\bm X$ or $\bm X'$ when $\bm X$ and $\bm X'$ is adjacent, implying that the information of each individual  is protected after releasing $M$. 

We then list several useful facts of designing DP algorithms. 
To create a DP algorithm, the arguably most common strategy is to add random noise to the output. 
Intuitively, the scale of the noise can not be too large, otherwise the accuracy of the output will be sacrificed. 
This scale is characterized by the sensitivity of the algorithm.
\begin{definition}
 For any algorithm $f : \mathcal{X}^n \rightarrow {\R}^d$ and two adjacent data sets $\bm X$ and $\bm X'$, the $\ell_p$-sensitivity of $f$ is defined as:
	$\Delta_{p}(f) = \sup_{\bm X,  \bm X'\in\mathcal{X}^n \text{ adjacent}}\|f(\bm X) - f(\bm X')\|_p.$
\end{definition}
For algorithms with finite $\ell_1$-sensitivity or $\ell_2$-sensitivity, we can achieve differential privacy by adding Laplace noises or Gaussian noises respectively. 
\begin{proposition}[The Laplace Mechanism \citep{dwork2006calibrating}]
Let $f: \mathcal X^n \to \R^d$ be a deterministic algorithm with $\Delta_1(f)< \infty$. For $\bm w \in \R^d$ with coordinates $w_1, w_2, \cdots, w_d$ be i.i.d samples drawn from Laplace$(\Delta_1(f)/\epsilon)$, $f(\bm X) +\bm  w$ is $(\epsilon, 0)$-DP. 
\end{proposition}
\begin{proposition}[The Gaussian Mechanism \citep{dwork2006calibrating}]
 Let $f: \mathcal X^n \to \R^d$ be a deterministic algorithm with $\Delta_2(f)< \infty$. For $ \bm w=(w_1, w_2, \cdots, w_d)$ with coordinates i.i.d  drawn from $ N(0, 2(\Delta_2(f)/\epsilon)^2\log(1.25/\delta))$, $f(\bm X) + \bm w$ is $(\epsilon, \delta)$-DP. 
\end{proposition}

These two mechanisms are computationally efficient, and are typically used to build more complicated algorithms. In the following, we introduce the post-processing and composition properties of differential privacy, which enable us to design complicated DP algorithms by combining simpler ones.
\begin{proposition}[Post-processing Property  \citep{dwork2006calibrating}]
Let $M$ be an $(\epsilon, \delta)$-DP algorithm and $g$ be an arbitrary function which takes $M(\bm X)$ as input, then $g(M(\bm X))$ is also $(\epsilon, \delta)$-DP.
\end{proposition}
\begin{proposition}[Composition property \citep{dwork2006calibrating}]\label{composition} For $i = 1, 2$, let $M_i$ be $(\varepsilon_i, \delta_i)$-DP algorithm, then $(M_1, M_2)$ is $(\epsilon_1 + \epsilon_2, \delta_1 + \delta_2)$-DP algorithm. 
\end{proposition}


In the following section, we will see that the above two properties are particularly useful in the design of the DP EM algorithm.

\section{High-dimensional EM Algorithm}\label{sec:hdem}
 In this section, we develop a novel DP EM algorithm for the (sparse) high-dimensional latent variable models. We are going to first introduce the detailed description of the proposed algorithm in Section~\ref{sec3-1}, and then present its theoretical properties in Section~\ref{sec3-2}. We will further apply our general method to three specific models in the next section.
\subsection{Methodology}\label{sec3-1} 

Suppose we have $i.i.d$ data sampled from the latent variable model~\eqref{eq:model} and aim to use the EM algorithm to find the maximum likelihood estimator in a DP manner. Since the EM algorithm is an iterative approach where the $t$-th iteration takes as input the $\bbeta^{t-1}$ from the M-step in the last iteration, it suffices to make each $\bbeta^{t}$ differentially private to ensure the privacy guarantee of the final output. 

Our algorithm relies on two key designs in the M-step. First, we use the gradient EM approach, and in the gradient update stage, we introduce a truncation step on the gradient to control the sensitivity of the gradient update, and thus we can appropriately determine the scale of noise we need to achieve differential privacy. Secondly, we propose to apply the noisy hard-thresholding algorithm (NoisyHT) \citep{dwork2018differentially} to pursue sparsity and achieve privacy at the same time. The NoisyHT algorithm is defined as follows, 

\begin{algorithm}[!htb]\label{algo:peeling}
\caption{Noisy Hard Thresholding (NoisyHT($\bm v, s, \lambda, \epsilon,\delta$)) \citep{dwork2018differentially}}
\begin{algorithmic}[1] 
\Require vector-valued function $ \bm v =  \bm v(\bm X) \in \R^d$ with data $\bm X$, sparsity $s$, privacy parameters $\varepsilon, \delta$, sensitivity $\lambda$. \\
\textbf{Initialization:} $S = \emptyset$.\\
\textbf{For} $i$ in $1$ \KwTo $s$:\\
\quad Generate $\bm w_i \in \R^d$ with $w_{i1}, w_{i2}, \cdots, w_{id} \stackrel{\text{i.i.d.}}{\sim} \text{Laplace}\left(\lambda \cdot \frac{2\sqrt{3s\log(1/\delta)}}{\varepsilon}\right)$\\
\quad Append $j^* = \text{argmax}_{j \in [d] \setminus S} (|v_j| + w_{ij})$ to $S$\\
\textbf{End For}\\
\quad Generate $\tilde {\bm w}$ with $\tilde w_{1}, \cdots, \tilde w_{d} \stackrel{\text{i.i.d.}}{\sim} \text{Laplace}\left(\lambda \cdot \frac{2\sqrt{3s\log(1/\delta)}}{\varepsilon}\right)$
\Ensure $P_S(\bm v+\tilde{\bm w})$ 

\end{algorithmic}

\end{algorithm}

In the last step, $P_S(\bm u)$ denotes the operator that makes $\bm u_{S^c}=0$ while preserving $\bm u_S$. A great advantage of this algorithm is that when the vector $\bm v= v(\bm X)$ has bounded $\ell_\infty$ sensitivity $\lambda$, the algorithm is guaranteed to be DP, as we can see in the lemma below.
\begin{lemma}[\citep{dwork2018differentially}]\label{lem:noisyht1}
If for every pair of adjacent data sets $\bm X,\bm X'\in\mathcal X^n$. we have $||v(\bm X)-v(\bm X')||_{\infty} < \lambda$, then the noisy hard-thresholding algorithm is $(\epsilon, \delta)$-DP. 
\end{lemma}
Another important property of the NoisyHT algorithm is its accuracy. Specifically, for the coordinates that are not chosen by the NoisyHT algorithm, the $\ell_2$ norm of these coordinates is upper bounded by that of the coordinates with the same size but chosen by the NoisyHT algorithm up to some error term. The formal statement is summarized in the following with the proof given in Appendix~\ref{plem:noisyht2}. 
\begin{lemma}\label{lem:noisyht2}
Let $S$ be the set chosen by the NoisyHT algorithm, $\bm v$ be the input vector and $\{\bm w_i\}_{i\in[s]}$ be defined as in the NoisyHT algorithm. For every set $R_1 \subset S$ and $R_2 \subset S^c$ such that $|R_1| = |R_2|$ and for every $c>0$, we have:
\[||\bm v_{R_2}||_2^2 \le (1+c)||\bm v_{R_1}||_2^2 +4 \cdot (1+1/c)\sum_{i\in[s]} ||\bm w_i||_{\infty}^2.\] 
\end{lemma}
After introducing the NoisyHT algorithm, we now proceed to the development of the DP EM algorithm. We update the estimator $\hat{\bm \beta}$ based on the NoisyHT algorithm in each M-step after truncation. 
Such a modified M-step guarantees that the output $\bbeta^t$ is sparse and differentially private in each iteration. We also note here that we use the sample-splitting in the algorithm, 
which makes $\bbeta^{t}$ independent with the batch of samples in $t$-th iteration and helps control the sensitivity of the gradient. The algorithm is summarized below:

\begin{algorithm}[H]\label{algo:hdem}
\caption{High-Dimensional DP EM algorithm}
\begin{algorithmic}[1] 
\Require Private parameters $(\epsilon,\delta)$, step size $\eta$, truncation level $T$, maximum number of iterations $N_0$, sparsity parameter $\hat{s}$.\\
\textbf{Initialization:} $\bm \beta^0$ with $||\bm \beta^0||_0 \le \hat{s}$.\\
\textbf{For} $t = 0,1,2,...N_0-1$:\\
\quad Compute $\bm \beta^{t+0.5} =\bm \beta^{t} + \eta \cdot f_T(\nabla Q_{n/N_0}(\bm \beta^t;\bm \beta^t)) $.\\
\quad Let $\bm \beta^{t+1} = \text{NoisyHT}(\bm \beta^{t+0.5},\hat{s},2\eta\cdot N_0\cdot T/n,\epsilon,\delta)$.\\
\textbf{End For}

\Ensure $\hat{\bbeta} = \bbeta^{N_0}$

\end{algorithmic}

\end{algorithm}

In the above algorithm, the truncation operator $f_T(\nabla Q_n(\bm \beta; \bm \beta))$ with truncation level $T$ is defined as $f_T(\nabla Q_n(\bm \beta; \bm \beta))=\frac{1}{n}\sum_{i=1}^n h_T(\nabla q_i(\bm \beta; \bm \beta))$, where $q_i(\bm \beta; \bm \beta)=\int_{\mathcal{Z}}
  k_{\bbeta}(\bm z_i| \bm y_i) \log f_{\bm \beta}(\bm y_i, \bm z_i)
  d \bm z_i$, 
and $h_T$ denotes some generic truncation function with $\ell_{\infty}$ norm upper bounded by $T$. 
Later in the application to different specific models, we will specify the form of $h_T$ for each model. The next lemma shows Algorithm~\ref{algo:hdem} is $(\epsilon,\delta)$-DP. 
\begin{lemma}\label{lem:noisyht3}
The output of high-dimensional DP EM algorithm (Algorithm~\ref{algo:hdem}) is $(\epsilon,\delta)$-DP.
\end{lemma}

The proof of Lemma~\ref{lem:noisyht3} is given in the Appendix~\ref{plem:noisyht3}. 
In theory, we shall take the truncation level $T$ and number of iterations $N_0$ to be the order of $\sqrt{\log n}$ and $\log n$, respectively, as we will show in the next section. For the sparsity level $\hat{s}$, we should choose it to have the same order of the true sparsity level $s^*$. While the true parameter $s^*$ is unknown in practice, $\hat{s}$ could be chosen through cross-validation. Given the  differential privacy guarantees, we are going to analyze the utility of this algorithm in the next section. 
\subsection{Theoretical gaurantees}\label{sec3-2}
In this section, we analyze the theoretical properties of the proposed high-dimensional DP EM algorithm. Before we lay out the main results, we first introduce four technical conditions. The first three conditions are standard in the literature of EM algorithms; for example, see \citep{balakrishnan2017statistical, wang2015high, zhu2017high, wang2020differentially}. The fourth condition is needed to control the sensitivity in the design of DP algorithms. We will verify these four conditions in the specific models in the next section.

\begin{condition}[Lipschitz-Gradient$(\gamma,\mathcal{B})$]\label{con: hdem1}
For the true parameter $\bm \beta^*$ and any $\bm \beta \in \mathcal{B}$, where $\mathcal{B}$ denotes a region in the parameter space. We have
\begin{equation*}
    ||\nabla Q(\bm \beta; \bm \beta^*)-\nabla Q(\bm \beta; \bm \beta)||_2 \le \gamma ||\bm \beta- \bm \beta^*||_2.
\end{equation*}
\end{condition}
This condition implies that if $\bm \beta$ is close to the true parameter $\bm \beta^*$, then the gradients $\nabla Q(\bm \beta; \bm \beta^*)$ and $\nabla Q(\bm \beta; \bm \beta)$ should also be close, which implies gradient stability. 
\begin{condition}[Concavity-Smoothness $(\mu,\nu,\mathcal{B})$]\label{con:hdem2}
For any $\bm \beta_1,\bm \beta_2 \in \mathcal{B}$, $Q(\cdot;\bm \beta^*)$ is $\mu$-smooth such that
\begin{equation*}
    Q(\bm \beta_1; \bm \beta^*) \ge  Q(\bm \beta_2; \bm \beta^*) + \langle \bm \beta_1- \bm \beta_2, \nabla Q(\bm \beta_2; \bm \beta^*) \rangle -\mu/2 \cdot ||\bm \beta_2- \bm \beta_1||_2^2,
\end{equation*}
and $\nu$-strongly concave such that
\begin{equation*}
     Q(\bm \beta_1; \bm \beta^*) \le  Q(\bm \beta_2; \bm \beta^*) + \langle \bm \beta_1- \bm \beta_2, \nabla Q(\bm \beta_2; \bm \beta^*) \rangle -\nu/2 \cdot ||\bm \beta_2- \bm \beta_1||_2^2.
\end{equation*}
\end{condition}

The Concavity-Smoothness condition indicates that when the second argument of $Q(\cdot;\cdot)$ is $\bm \beta^*$, $Q(\cdot;\cdot)$ is a well-behaved convex function that it can be upper bounded and lower bounded by two quadratic functions. 
This condition ensures the geometric decay of the optimization error in the statistical analysis. 


\begin{condition}[Statistical-Error$(\alpha,\tau,s,n,\mathcal{B})$]\label{con:hdem3}
For any fixed $\bm \beta \in \mathcal{B}$ and $||\bm \beta||_0 \le s$, we have with probability at least $1-\tau$,
\begin{equation*}
    ||\nabla Q(\bm \beta; \bm \beta)-\nabla Q_n(\bm \beta; \bm \beta)||_{\infty} \le \alpha.
\end{equation*}
\end{condition}
In this condition, the statistical error is quantified by the $\ell_\infty$ norm between the population quantity $\nabla Q(\bm \beta; \bm \beta)$ and its sample version $\nabla Q_n(\bm \beta; \bm \beta)$. Such a bound is different from the $\ell_2$ norm bound considered in the classic low-dimensional DP EM algorithms \citep{wang2020differentially}.  In the high-dimensional setting, although for each index, the statistical error is small, the $\ell_2$ norm can still be quite large. 
This fine-grained $\ell_\infty$ bound enables us to iteratively quantify the statistical accuracy when using the NoisyHT in the M-step.

\begin{condition}[Truncation-Error$(\xi,\phi,s,n,T,\mathcal{B})$]\label{con:hdem4}
For any $ \bm \beta \in \mathcal{B}$ and $\|\bm \beta\|_0 \le s$, there exists a non-incresaing function $\phi$, such that for the truncation level $T$, with probability $1-\phi(\xi)$, 
\[||\nabla Q_{n} (\bm \beta;\bm \beta)-f_T(\nabla Q_{n} (\bm \beta; \bm \beta))||_2 \le \xi. \]
\end{condition}
The Truncation-Error condition quantifies the error caused by the truncation step in Algorithm~\ref{algo:hdem}. Intuitively, when $T$ is large, the truncation error $\xi$ can be very small, while leading to larger sensitivity and larger injected noise to ensure privacy. We need to carefully choose $T$ to strike a balance between the statistical accuracy and privacy cost. Below we show the main result of this section, with the detailed proof given in Section~\ref{pthm:hdem}.
\begin{theorem}\label{thm:hdem}
Suppose the true parameter $\bm \beta^*$ is sparse with $\|\bm \beta^*\|_0 \le s^*$. We define $\mathcal{B} = \{\bm \beta : ||\bm \beta - \bm \beta^*||_2 \le R\}$ with $R = L \cdot \|\bm \beta^*\|_2$ for some $L \in (0,1)$. We assume the Concavity-Smoothness$(\mu,\nu,\mathcal{B})$ holds and the initialization $\bm \beta^0 $ satisfies $||\bm \beta^0 - \bm \beta^*||_2 \le R/2$. We further assume that the Lipschitz-Gradient$(\gamma,\mathcal{B})$ holds and define $\kappa = 1-2\cdot\frac{\nu-\gamma}{\nu+\mu} \in (0,1)$.   In Algorithm~\ref{algo:hdem}, we let the step size $\eta=2/(\mu +\nu)$, the number of iterations $N_0 \asymp \log n$, the sparsity level $\hat{s} \ge c_0 \cdot \max(\frac{16}{(1/\kappa -1)^2},\frac{4 \cdot (1+L)^2}{(1-L)^2} )\cdot s^*$ where $c_0$ is a constant greater than 1 and $\hat{s} = O(s^*)$. 
We assume the condition Truncation-Error($\xi, \phi, \hat{s}, n/N_0,T,\mathcal{B}$) holds with $T \asymp \sqrt{\log n}$ and $\phi(\xi)\cdot N_0 = o(1)$. Moreover, we assume the condition Statistical-Error$(\alpha,\tau/{N_0},\hat{s},n/{N_0},\mathcal{B})$ holds and assume that $\alpha = o(1) $ and there exists a constant $c_1>0$, and $(\sqrt{\hat{s}}+c_1/\sqrt{1-L}\cdot\sqrt{s^*})\cdot\eta\cdot\alpha \le \min((1-\sqrt{\kappa})^2\cdot R,  \frac{(1-L)^2}{2\cdot(1+L)}\cdot\|\beta^*\|_2)$, also $\frac{s^* \log d}{n} \cdot (\log n)^{\frac{3}{2}} = o(1)$. 
Then 
there exist constants $K, m_0, m_1$, it holds that
\begin{align}
    \notag\|\bm \beta^{t} -\bm \beta^*\|_2 \notag
    \notag&\le  \kappa^{t/2}\cdot  ||{\bm \beta}^{0}-\bm \beta^*||_2+ \frac{(\sqrt{\hat{s}}+c_1/\sqrt{1-L} \cdot \sqrt{s^*})\cdot \eta}{1-\sqrt{\kappa}}\cdot \alpha +   K \cdot \frac{s^* \cdot \log d \cdot \sqrt{\log(1/\delta)} \log^{3/2} n}{n \epsilon} \\
    &+\eta\cdot \xi/(1-\sqrt{\kappa}),  \label{eq:result}
\end{align} 
with probability $1-\tau-N_0 \cdot \phi(\xi)-m_0\cdot s^* \cdot \log n\cdot\exp(-m_1 \log d)$. Specifically, for the output in Algorithm~\ref{algo:hdem}, it holds that
\begin{align}
    \notag\|\bm \beta^{N_0} -\bm \beta^*\|_2 \notag
    \notag&\le \frac{(\sqrt{\hat{s}}+c_1/\sqrt{1-L} \cdot \sqrt{s^*})\cdot \eta}{1-\sqrt{\kappa}}\cdot \alpha +   K \cdot \frac{s^* \cdot \log d \cdot \sqrt{\log(1/\delta)} \log^{3/2} n}{n \epsilon} +\eta\cdot \xi/(1-\sqrt{\kappa}),  \label{eq:result2}
\end{align}
with probability $1-\tau-N_0 \cdot \phi(\xi)-m_0\cdot s^* \cdot \log n\cdot\exp(-m_1 \log d)$. 
\end{theorem}
To interpret this result, let us discuss the four terms in \eqref{eq:result}. The first term, $\kappa^{t/2}\cdot  ||{\bm \beta}^{0}-\bm \beta^*||_2 =O( \kappa^{t/2} \cdot \|\bbeta^*\|_2) $, is the optimization error. With $\kappa \in (0,1)$, this term shrinks to zero at a geometric rate when the iteration number $t$ is sufficiently large. 
The second term is of order $ \sqrt{s^*}\cdot\alpha$ when $\hat{s}$ is chosen as the same order of $s^*$, this is the statistical error caused by finite samples. 
We will further show that in some specific models, $\alpha$ is of the order $\sqrt{\log d\cdot \log n/n}$ and makes the second term to be $O(\sqrt{s^* \log d\cdot\log n/n})$.
The third term $ K \cdot \frac{s^* \log d \cdot \sqrt{\log(1/\delta)} \log^{3/2} n}{n \epsilon}$ can be seen as the cost of privacy, as this error is introduced by the additional requirement that the output needs to be $(\epsilon,\delta)$-DP. This term becomes larger when the privacy constraint becomes more stringent ($\epsilon,\delta$ become smaller). Typically, in practice, we choose $\delta=O(1/n^a)$ for some $a\ge 1$ and $\epsilon$ a small constant. This implies that when $d, s^*$ and $n$ satisfy $\frac{s^*  \log d \cdot (\log n)^2}{n} \cdot \frac{\log{1/\delta}}{\epsilon^2} = o(1)$, the cost of privacy will be negligible comparing to the statistical error. In this case, we can gain privacy for free in terms of convergence rate. 
 The fourth term is due to the truncation of the gradient. Under the Truncation-Error condition with an appropriately chosen truncation parameter,  this term reaches a convergence rate dominated by the statistical error up to logarithm factors.


\section{DP EM Algorithm in Specific Models}\label{sec:hdemmodel} 
In this section, we apply the results developed in Section~\ref{sec:hdem} to specific statistical models and establish concrete convergence rates. We will discuss three models, the Gaussian mixture model, mixture of regression model and regression with missing covariates, in Sections~\ref{sec4-1},~\ref{sec4-2} and~\ref{sec4-3} respectively. Further, for each statistical model, we establish the minimax lower bound of the convergence rate, and demonstrate that our algorithm obtains a near minimax optimal rate of convergence.
\subsection{Gaussian mixture model}\label{sec4-1}
 In this subsection, we first apply the results in  Section~\ref{sec:hdem} to the Gaussian mixture model. By verifying the conditions in Theorem~\ref{thm:hdem}, we establish the convergence rate of the DP estimation in the Gaussian mixture model. 

In a standard Gaussian mixture model, we assume:
\begin{equation}\label{model:gmm} 
\bm Y = Z\cdot \bm \beta^* + \bm e,
\end{equation} 
where $\bm Y$ is a $d$-dimensional  output and $\bm e \sim N(\bm 0, \sigma^2 \bm I_d)$. In this model, $\bm \beta^*$ and $-\bbeta^*$ are the $d$-dimensional vectors representing the population means of each class, and $Z$ is a class indicator variable with $\Pro(Z = 1) = 1/2$ and $\Pro(Z = -1) =1/2$. Note that $Z$ is a hidden variable and independent of $\bm e$.  In the high dimensional setting, we assume $\bm \beta^*$ to be sparse.  


Let $\bm y_1, \bm y_2 ...\bm y_n$ be $n$ $i.i.d$ samples from the Gaussian mixture model. Using the framework of the EM method introduced in  Section~\ref{sec:hdem}, we need to calculate
\[Q_n(\bm \beta' ;\bm \beta) = - \frac{1}{2n} \sum_{i=1}^n  w_{\bm \beta}(\bm y_i)
  ||\bm y_i - \bm \beta'||_2^2 + [1 - w_{\bm \beta}(\bm y_i)] \cdot  ||\bm y_i + \bm \beta'||_2^2 ,\]
where 
\[w_{\bm \beta}(\bm y) = \frac{1}{1+\exp(-\langle \bm \beta, \bm y \rangle /{\sigma^2} )}.\] 
Then, for the $M$-step in the $t$-th iteration given $\bm \beta^t$, the update rule is given by
\[ \bm \beta ^{t+1} = \bm \beta^ t + \eta \cdot \nabla Q_n(\bm \beta^t ; \bm \beta^t) \text{, where } \nabla Q_n(\bm \beta ;\bm \beta) = \frac{1}{n} \sum_{i=1}^n [2\cdot w_{\bm \beta}(\bm y_i)-1] \cdot \bm y_i -\bm \beta.\]
Given this expression, we now present the DP estimation in the high-dimensional Gaussian mixture model by applying Algorithm~\ref{algo:hdem}. The algorithm is presented in Algorithm~\ref{algo:hdgmm}.

\begin{algorithm}[!htb]\label{algo:hdgmm}
\caption{DP Algorithm for High-Dimensional  Gaussian Mixture Model}
\begin{algorithmic}[1] 
\Require Private parameters $(\epsilon,\delta)$, step size $\eta$, truncation level $T$, maximum number of iterations $N_0$, sparsity parameter $\hat{s}$. \\
\textbf{Initialization:} $\bm \beta^0$ with $||\bm \beta^0||_0 \le \hat{s}$.\\
\textbf{For} $t = 0,1,2,...N_0-1$:\\
\quad Compute $\bm \beta^{t+0.5} = \bm \beta^{t} + \eta\cdot N_0/n \cdot  \sum_{i=1}^{n/N_0} [(2 w_{\bm \beta^t}(\bm y_i)-1)\cdot\Pi_{T}(\bm y_i)-\bm \beta^t]$.\\
\quad Let $\bm \beta^{t+1} = \text{NoisyHT}(\bm \beta^{t+0.5},\hat{s},2\eta\cdot T\cdot N_0/n,\epsilon,\delta)$. \\
\textbf{End For}

\Ensure $\hat{\bm \beta} = \bm \beta^{N_0}$
\end{algorithmic}
\end{algorithm}

To derive the converge rate of $\hat\bbeta$, we need to first verify Conditions~\ref{con: hdem1}-\ref{con:hdem4}.  Conditions~\ref{con: hdem1}-\ref{con:hdem3} are standard in the literature of EM algorithms \citep{balakrishnan2017statistical, wang2015high}. 
We adapt them into our setting and state the results altogether below.

\begin{lemma}\label{lem:hdgmm}
Suppose we can always find a sufficiently large constant $\phi$ to be the lower bound of the signal-to-noise ratio, $||\bm \beta^*||_2/\sigma > \phi$. Then
\begin{itemize}
    \item There exists a constant $C>0$ such that Condition~\ref{con: hdem1}, Lipschitz-Gradient $(\gamma, \mathcal{B})$ and Condition~\ref{con:hdem2}, Concavity-Smoothness $(\mu,\nu,\mathcal{B})$ hold with the parameters
\[\gamma = \exp(-C \cdot \phi^2), \mu = \nu =1, \mathcal{B} = \{\bbeta:||\bm \beta- \bm \beta^*||_2 \le R\} \text{ with } R = 1/4 \cdot ||\bm \beta^*||_2.\]

\item Condition~\ref{con:hdem3}, Statistical-Error$(\alpha, \tau, \hat{s}, n, \mathcal{B})$ holds with a constant $C_1$ and
\[\alpha = C_1 \cdot (||\bm \beta^*||_{\infty}+\sigma)\cdot \sqrt{\frac{\log d + \log(2/\tau)}{n}}.\] 

\item Condition~\ref{con:hdem4}, Truncation-Error$(\xi, \phi, \hat{s}, n/N_0, T, \mathcal{B})$ holds with $T \asymp \sqrt{\log n}$ and with probability $1-m_0/\log d \cdot \log n$, there exists a constant $C_2$, such that
\[||\nabla Q_{n/N_0} (\bm \beta;\bm \beta)-f_T(\nabla Q_{n/N_0} (\bm \beta; \bm \beta))||_2 \le C_2 \cdot \sqrt{\frac{s^* \cdot \log d \cdot \log n}{n}}.\]

\end{itemize}

\end{lemma}

The detailed proof of Lemma~\ref{lem:hdgmm} is given in Appendix~\ref{plem:hdgmm}. Given these verified conditions, the following theorem establishes the results for the DP estimation in the high-dimensional Gaussian mixture model.
\begin{theorem}\label{thm:hdgmm}
We implement Algorithm~\ref{algo:hdgmm} to the observations generated from the Gaussian mixture model~\eqref{model:gmm}. Let $\mathcal{B}, R, \mu, \nu, \gamma$ defined the same way as in Lemma~\ref{lem:hdgmm}. We assume $||\bm \beta^*||_2/\sigma > \phi$ for a sufficiently large constant $\phi>0$. Let the initialization $\bbeta^0$ satisfy $||\bm \beta^0 - \bm \beta^*||_2 \le R/2$ and $\kappa =  \gamma$. 
Also, set the sparsity level $\hat{s} \ge c_0 \cdot \max(\frac{16}{(1/\kappa-1)^2}, 100/9) \cdot s^*$ with $c_0>1$ be a constant and $\hat{s} = O(s^*)$. The step size is chosen as $\eta = 1$. We choose the number of iterations $N_0 \asymp \log n$, and let truncation level $T \asymp \sqrt{\log n}$. 
   We further assume that $\frac{s^* \log d}{n} \cdot (\log n)^{\frac{3}{2}} = o(1)$.  
Then, the proposed Algorithm~\ref{algo:hdgmm} 
is $(\epsilon,\delta)$-DP. Also, we can show that 
there exist sufficient large constants $C,C_1$, it holds that
\begin{align}\label{gmmupp}
     ||\hat{\bm \beta}- \bm \beta^*||_2 &\le C \cdot \sqrt{\frac{s^* \cdot \log d \cdot \log n}{n}} + C_1 \cdot  \frac{s^* \log d \cdot \sqrt{\log(1/{\delta})}{(\log n)}^{\frac{3}{2}}}{n\epsilon}.
\end{align}
with probability $1-m_0\cdot s^* \cdot \log n\cdot\exp(-m_1 \log d) -m_2/\log d- m_3\cdot d^{- 1/2}$, where $m_0, m_1, m_2, m_3$ are constants.  
\end{theorem}
The proof of Theorem~\ref{thm:hdgmm} is given in Section~\ref{pthm:hdgmm}. 
Similar to the interpretation for the results in \eqref{eq:result}, the first and the second items are respectively the statistical error and the cost of privacy. 

In the following, we present the minimax lower bound for the estimation in the high-dimensional Gaussian mixture model with differential privacy constraints, indicating the convergence rate obtained above is near optimal. 
\begin{proposition}\label{thm:hdgmmlow}
Suppose $\bm Y = \{\bm y_1, \bm y_2,...,\bm y_n\}$ be the data set of $n$ samples observed from the Gaussian mixture model~\eqref{model:gmm} and let $M$ be any algorithm such that $M \in \mathcal{M}_{\epsilon,\delta}$, where $\mathcal{M}_{\epsilon,\delta}$ be the set of all $(\epsilon, \delta)$-DP algorithms for the estimation of the true parameter $\bm \beta^*$. Then there exists a constant $c$, if $s^* = o(d^{1-\omega})$ for some fixed $\omega > 0$, $0< \epsilon <1$ and $\delta < n^{-(1+\omega)}$ for some fixed $\omega > 0$, we have
\[\inf_{M \in \mathcal{M}_{\epsilon,\delta}} \sup_{\bm \beta^* \in \mathbb{R}^d, \|\bm \beta^*\|_0 \le s^*} \E\| {M(\bm Y)}-\bm \beta^*\|_2 \ge c \cdot (\sqrt{\frac{s^* \log d}{n}} + \frac{s^* \log d}{n \epsilon} ).\] 
\end{proposition}
The proof of Proposition~\ref{thm:hdgmmlow} is given in Section~\ref{pthm:hdgmmlow}. By comparing the results in Theorem~\ref{thm:hdgmm} and Proposition~\ref{thm:hdgmmlow}, our algorithm is shown to attain the minimax optimality up to logarithm factors in the high-dimensional Gaussian mixture models. 

\subsection{Mixture of regression model}\label{sec4-2}
\noindent We continue to demonstrate the proposed algorithm in the mixture of regression model, where we assume the following data generative process
\begin{equation}\label{model:mlr} Y = Z\cdot \bm X^\top \bm\beta^* + \bm e,
\end{equation}
where $ Y \in \mathbb{R}$ is the response, $Z$ is an indicator variable with $\Pro(Z = 1)=\Pro(Z = -1) =1/2$, $\bm X \sim N(0, \bm I_d)$, $\bm e \sim N(\bm 0, \sigma^2 \bm I_d)$, and $\bm \beta^*$ is a $d$-dimensional coefficients vector, we also require $\bm \beta^*$ to be sparse in the high-dimensional setting. Note that $Z,\bm e, \bm X$ are independent with each other. 

Let $(\bm x_1, y_1),(\bm x_2, y_2) ...(\bm x_n,y_n)$ be the $n$ $i.i.d.$  observed samples from the mixture of regression model. Then, to use the EM algorithm, we need to compute
\[Q_n(\bm \beta' ;\bm \beta) = - \frac{1}{2n} \sum_{i=1}^n  w_{\bm \beta}(\bm x_i,y_i)
  (y_i - \langle \bm x_i,\bbeta'\rangle)^2 + [1 - w_{\bm \bbeta}(\bm x_i, y_i)] \cdot  (y_i + \langle \bm x_i, \bm \beta'\rangle)^2, \]
where 
$w_{\bm \beta}(\bm x,y) = ({1+\exp(-y \cdot \langle \bm \beta, \bm x \rangle /{\sigma^2} )})^{-1}.$

According to the gradient EM update rule, for the $t$-th iteration $\bm \beta^t$, we update  $\hat{\bm \beta}$ by
\[ \bm \beta ^{t+1} = \bm \beta^ t + \eta \cdot \nabla Q_n(\bm \beta^t ;\bm \beta^t) \text{ ,where } \nabla Q_n(\bm \beta ;\bm \beta) = \frac{1}{n} \sum_{i=1}^n [2\cdot w_{\bm \beta}(\bm x_i,y_i)\cdot y_i\cdot \bm x_i - \bm x_i \cdot \bm x_i^\top \cdot \bm \beta^t].\]

Similar to the Gaussian Mixture model, to apply Algorithm~\ref{algo:hdem}, we need to specify the truncation operator $f_{T}(\nabla Q_n(\bm \beta^t;\bm \beta^t))$. Rather than using truncation on the whole gradient, we perform the truncation on $y_i$, $\bm x_i$ and $\bm x_i^\top\bm \beta$ respectively, which leads to a more refined analysis and improved rate in the statistical analysis. Specifically, we define
\[f_{T}(\nabla Q_n(\bm \beta;\bm \beta)) = \frac{1}{n} \sum_{i=1}^n[2 w_{\bm \beta}(\bm x_i, y_i)\cdot \Pi_{T}(y_i)\cdot\Pi_{T}(\bm x_i)-\Pi_{T}(\bm x_i)\cdot \Pi_{T}(\bm x_i^\top\bm \beta)].\]

Due to space constraints, we present the full algorithm for mixture of regression model in section~\ref{supp:mrm}. We also verify Conditions~\ref{con: hdem1}-\ref{con:hdem4} in the mixture of regression model, and summarize the results in Lemma~\ref{lem:hdmrm}. In the following, we show the theoretical guarantees for the high-dimensional DP EM algorithm on the mixture of regression model, with proof given in Section~\ref{pthm:hdmrm}.

\begin{theorem}\label{thm:hdmrm}
We implement the Algorithm~\ref{algo:hdmrm} to the sample generated from the mixture of regression model \eqref{model:mlr}. Let $\mathcal{B},R,\mu,\nu,\gamma$ defined as in Lemma~\ref{lem:hdmrm}. We assume $||\bm \beta^*||_2/\sigma > \phi$ for a sufficiently large $\phi>0$. Let the initialization $||\bm \beta^0 - \bm \beta||_2 \le R/2$ and $\kappa = \gamma$. 
Also, set the sparsity level $\hat{s} \ge c_0 \cdot \max(\frac{16}{(1/\kappa-1)^2}, \frac{4\cdot 33^2}{31^2}) \cdot s^*$ with $c_0>1$ be a constant and $\hat{s} = O(s^*)$. The step size is chosen as $\eta = 1$.  We choose  the number of iterations $N_0 \asymp \log n $, and let the truncation level $T \asymp \sqrt{\log n}$. We  assume that $\frac{s^* \log d}{n} \cdot (\log n)^{\frac{3}{2}} = o(1)$. 
Then, the proposed Algorithm~\ref{algo:hdmrm} is $(\epsilon,\delta)$-DP, 
there exist constants $C,C_1$, it holds that
\begin{align}\label{uppermrm}
     ||\hat{\bm \beta}- \bm \beta^*||_2 &\le C \cdot \sqrt{\frac{s^* \log d}{n}} \cdot \log n + C_1 \cdot  \frac{s^*\cdot\log d \cdot  \sqrt{ \log(1/{\delta})}{(\log n)}^{\frac{3}{2}}}{n\epsilon}. 
\end{align}
with probability $1-m_0\cdot s^* \log n\cdot\exp(-m_1 \log d) -m_2/\log d-m_3\cdot d^{-1/2}$, where $m_0, m_1, m_2, m_3$ are constants. 
\end{theorem}

Theorem~\ref{thm:hdmrm} achieves a similar rate consisting of statistical error and privacy cost as we have discussed above in the general private EM algorithm and the Gaussian mixture model. The proposition below shows the lower bound in the mixture of regression model.

\begin{proposition}\label{thm:hdmrmlow}
Suppose $( Y, \bm X) = \{( y_1, \bm x_1), ( y_2,\bm x_2),...( y_n, \bm x_n)\}$ be the data set of $n$ samples observed from the mixture of regression model \eqref{model:mlr}. Let $M$ and $\mathcal{M}_{\epsilon,\delta}$ defined as in Proposition~\ref{thm:hdgmmlow}. Then there exists a constant $c$, if $s^* = o(d^{1-\omega})$ for some fixed $\omega > 0$, $0< \epsilon <1$ and $\delta < n^{-(1+\omega)}$ for some fixed $\omega > 0$, we have
\[\inf_{M \in \mathcal{M}_{\epsilon,\delta}} \sup_{\bm \beta^* \in \mathbb{R}^d, \|\bm \beta^*\|_0 \le s^*} \E\|{M(Y, \bm X)}-\bm \beta^*\|_2 \ge c \cdot (\sqrt{\frac{s^* \log d}{n}} + \frac{s^* \log d}{n \epsilon} ).\]
\end{proposition}
The proof of Proposition~\ref{thm:hdmrmlow} is in Section~\ref{pthm:hdmrmlow}. Comparing the results from Theorem~\ref{thm:hdmrm} and Proposition~\ref{thm:hdmrmlow}, our algorithm attains the lower bound up to logarithm factors.
\subsection{Regression with missing covariates}\label{sec4-3}
The last model we discuss in this section is the regression with missing covariates. For the model setup, we assume the following data generative process
\[ Y = \bm X^\top \bm\beta^* +  e,\]
where $ Y \in \mathbb{R}$ is the response, $\bm X \sim N(0, \bm I_d)$, $ e \sim N( 0, \sigma^2 )$ and $e, \bm X$ are independent. 
 $\bm \beta^*$ is a $d$-dimensional coefficient vector, and we require $\bm \beta^*$ to be sparse in the high-dimensional setting with $\|\bm \beta^*\|_0 \le s^*$.  
  Let $(\bm x_1, y_1), ...,(\bm x_n,y_n)$ be $n$ $i.i.d.$ samples generated from the above model. For each $\bm x_i$, we assume the missing completely at random model such that each coordinate of $\bm x_i$ is missing independently with probability $p \in [0,1)$. Specifically, for each $\bm x_i$, we denote $\tilde{\bm x}_i$ be the observed covariates such that $\tilde{\bm x}_i=\bm z_i \odot \bm x_i $, where $\odot$ denotes the Hadamard product and $\bm z_i$ is a $d$-dimensional Bernoulli random vector with $z_{ij} = 1$ if $x_{ij}$ is observed and  $z_{ij} = 0$ if $x_{ij}$ is missing. 
  Then, by the EM algorithm, we compute 
\[Q_n(\bm \beta' ;\bm \beta) =  \frac{1}{n} \sum_{i=1}^n y_i \cdot (\bm \beta')^\top \cdot m_{\bm \beta}(\tilde {\bm x}_i, y_i) -\frac{1}{2} (\bm \beta')^\top \cdot K_{\bm \beta}(\tilde{\bm x}_i, y_i) \cdot \bm \beta'.\]
Here, $m_{\bm \beta}(\cdot,\cdot) \in \mathbb{R}^d$ and $K_{\bm \beta}(\cdot,\cdot) \in \mathbb{R}^{d \times d}$ are defined as
\[m_{\bm \beta}(\tilde{\bm x}_i,y_i) = \bm z_i \odot \bm x_i + \frac{y_i - \langle \bm \beta , \bm z_i \odot \bm x_i \rangle}{\sigma^2 + \|(\bm 1 - \bm z_i) \odot \bm \beta\|_2^2} \cdot (\bm 1 - \bm z_i) \odot \bm \beta,\] 
\begin{align*}
    K_{\bm \beta}(\tilde{\bm x}_i, y_i) = \text{diag}(\bm 1 - \bm z_i) + m_{\bm \beta}(\tilde{\bm x}_i,y_i)\cdot[ m_{\bm \beta}(\tilde{\bm x}_i,y_i)]^\top -[(\bm 1 - \bm z_i) \odot m_{\bm \beta}(\tilde{\bm x}_i,y_i)] \cdot[ (\bm 1 - \bm z_i) \odot m_{\bm \beta}(\tilde{\bm x}_i,y_i)]^\top.
\end{align*}
Then, according to the gradient EM update, for the $t$-th iteration $\bm \beta^t$, the update rule for the estimation of $\bm \beta$ is given below
\[ \bm \beta ^{t+1} = \bm \beta^ t + \eta \cdot \nabla Q_n(\bm \beta^t ;\bm \beta^t) \text{ ,where } \nabla Q_n(\bm \beta ;\bm \beta) = \frac{1}{n} \sum_{i=1}^n [y_i \cdot m_{\bm \beta}(\tilde{\bm x}_i ,y_i) - K_{\bm \beta}(\tilde{\bm x}_i, y_i) \cdot \bm \beta].\]

Similar as before, to apply Algorithm~\ref{algo:hdem}, we also need to specify the truncation operator $f_{T}(\nabla Q_n(\bm \beta^t;\bm \beta^t))$.  According to the definition of $m_{\bm \beta}(\tilde{\bm x}_i,y_i)$, when $\bm \beta$ is close to $\bm \beta^*$, we find that $m_{\bm \beta}(\tilde{\bm x}_i,y_i)$ is close to $\bm z_i \odot \bm x_i$, so we propose to truncate on $m_{\bm \beta}(\tilde{\bm x}_i,y_i)$ and $m_{\bm \beta}(\tilde{\bm x}_i,y_i)^\top \cdot \bm \beta$. 
 Specifically, let
\begin{align*}
    f_{T}(\nabla Q_n(\bm \beta;\bm \beta)) &= \frac{1}{n} \sum_{i=1}^n [\Pi_T (y_i) \cdot \Pi_{T} (m_{\bm \beta}(\tilde{\bm x}_i ,y_i)) - \text{diag}(\bm 1 - \bm z_i) \cdot \bm \beta \\&- \Pi_{T} (m_{\bm \beta}(\tilde{\bm x}_i,y_i))\cdot\Pi_T( m_{\bm \beta}(\tilde{\bm x}_i,y_i)^\top \cdot \bm \beta) \\
    &+ \Pi_{T}((\bm 1 - \bm z_i) \odot m_{\bm \beta}(\tilde{\bm x}_i,y_i)) \cdot \Pi_{T}(( (\bm 1 - \bm z_i) \odot m_{\bm \beta}(\tilde{\bm x}_i,y_i))^\top \cdot \bm \beta).
\end{align*}

We present the full DP algorithm for regression with missing covariates model in section~\ref{supp:mcm}, and verify Conditions~\ref{con: hdem1}-\ref{con:hdem4} in the regression with missing covariates model. The results are summarized as Lemma~\ref{lem:hdmcm}. In the following, we show the theoretical guarantees for the high-dimensional DP EM algorithm on the mixture of regression model, with the proof  given in Section~\ref{pthm:hdmcm}.

\begin{theorem}\label{thm:hdmcm}
We implement Algorithm~\ref{algo:hdmcm} to the samples generated from the regression with missing covariates model. Let $\mathcal{B},R, L, \mu,\nu,\gamma$ are defined as in Lemma~\ref{lem:hdmcm}, and assume the initialization $\bbeta^0$ satisfies $||\bm \beta^0 - \bm \beta^*||_2 \le R/2$ and $\kappa = \gamma$. 
Also, set the sparsity level $\hat{s} \ge c_0 \cdot \max(\frac{16}{(1/\kappa-1)^2}, \frac{4\cdot(1+L)^2}{(1-L)^2}) \cdot s^*$ with $c_0>1$ be a constant and $\hat{s} = O(s^*)$. We choose the step size $\eta = 1$, the number of iterations $N_0 \asymp \log n $, and the truncation level $T \asymp \sqrt{\log n}$. We further assume that $\frac{s^* \log d}{n} \cdot (\log n)^{\frac{3}{2}} = o(1)$. 
Then, the proposed Algorithm~\ref{algo:hdmcm} is $(\epsilon,\delta)$-DP, 
there exist constants $C,C_1, m_0, m_1, m_2, m_3 $, it holds that
\begin{align*}
     ||\hat{\bbeta}-\bbeta^*||_2 &\le C \cdot \sqrt{\frac{s^* \log d}{n}} \cdot \log n + C_1 \cdot  \frac{s^* \cdot \log d \cdot \sqrt{\log(1/{\delta})}{(\log n)}^{\frac{3}{2}}}{n\epsilon}. 
\end{align*}
with probability $1-m_0\cdot s^* \cdot \log n\cdot\exp(-m_1 \log d) -m_2/\log d - m_3\cdot d^{-1/2}$ where $m_0, m_1, m_2, m_3$ are constants. 
\end{theorem}
This results is similar as before, as 
the convergence rate in the DP estimation in the regression with missing covariates consists of the statistical error $O_P(\sqrt{\frac{s^* \log d}{n}} \cdot \log n)$ and the privacy cost $O_P( \frac{s^* \cdot \log d \cdot \sqrt{\log(1/{\delta})}{(\log n)}^{\frac{3}{2}}}{n\epsilon})$.  
Again, we show the minimax lower bound for the estimation in the regression with missing covariates with differential privacy constraints. 
\begin{proposition}\label{thm:hdmcmlow}
 Suppose $( Y, \bm X) = \{( y_1, \bm x_1), ( y_2,\bm x_2),...( y_n, \bm x_n)\}$ be the data set of $n$ samples observed from the regression with missing covariates discussed above and let $M$ and $\mathcal{M}_{\epsilon,\delta}$ defined as in Proposition~\ref{thm:hdgmmlow}. Then there exists a constant $c$, if $s^* = o(d^{1-\omega})$ for some fixed $\omega > 0$, $0< \epsilon <1$ and $\delta < n^{-(1+\omega)}$ for some fixed $\omega > 0$, we have
\[\inf_{M \in \mathcal{M}_{\epsilon,\delta}} \sup_{\bm \beta^* \in \mathbb{R}^d, \|\bm \beta^*\|_0 \le s^*} \E\|{M(Y, \bm X)}-\bm \beta^*\|_2 \ge c \cdot (\sqrt{\frac{s^* \log d}{n}} + \frac{s^* \log d}{n \epsilon} ).\]
\end{proposition}
As a result, for the high-dimensional regression of missing covariates model, comparing the upper and lower bounds in Theorem~\ref{thm:hdmcm} and Proposition~\ref{thm:hdmcmlow}, our algorithm attains the optimal rate of convergence up to logarithm factors. 

\section{Low-dimensional DP EM Algorithm}\label{sec:ldem} 
 In this section, we extend the technical tools we developed in Section~\ref{sec:hdem} to the classic low-dimensional setting, and 
propose the DP EM algorithm in this low-dimensional regime.  We will further apply our proposed algorithm to the Gaussian mixture model as an example, and show that the proposed algorithm obtains a near-optimal rate of convergence. 

For the low-dimensional case, instead of using the noisy hard-thresholding algorithm, here we use the Gaussian mechanism in the M-step for each iteration. Similar to the high-dimensional setting, we use the sample splitting in each iteration, and the truncation step in each M-step to ensure bounded sensitivity.  The algorithm is summarized below.
 
\begin{algorithm}[!htb]\label{algo:ldem}
\caption{Low-Dimensional DP EM algorithm}
\begin{algorithmic}[1] 
\Require Private parameters $(\epsilon,\delta)$, step size $\eta$, truncation level $T$, maximum number of iterations $N_0$. \\
\textbf{Initialization:} $\bm \beta^0$\\
\textbf{For} $t = 0,1,2,...N_0-1$:\\
\quad Compute $\bm \beta^{t+1} = \bm \beta^{t} + \eta f_T(\nabla Q_{n/N_0}(\bm \beta^t;\bm \beta^t)) +W_t $. $W_t$ is a random vector $(\xi_1,\xi_2,...\xi_d)^\top$, where $\xi_1,\xi_2,...\xi_d$ are i.i.d sample drawn from $N(0, \frac{2 \eta^2 d (2T)^2 {N_0}^2 \log(1.25/\delta)}{n^2 \epsilon^2})$\\
\textbf{End For}

\Ensure $\hat{\bm \beta} = \bm \beta^{N_0}$
 
\end{algorithmic}

\end{algorithm}
\begin{lemma}
The output $\hat\bbeta$ of the low dimensional DP EM algorithm (Algorithm~\ref{algo:ldem}) is $(\epsilon, \delta)$-DP.
\end{lemma}
 
Given the privacy guarantee, we then analyze the statistical accuracy of  this algorithm.  Before that, we need to modify the Condition~\ref{con:hdem3} and Condition~\ref{con:hdem4} to fit into the low-dimensional setting. 
\begin{condition}[Statistical-Error-2 $(\alpha,\tau,n,\mathcal{B})$]\label{con:ldem} 
For any fixed $\bbeta \in \mathcal{B}$, we have that with probability at least $1-\tau$,
\begin{equation*}
    ||\nabla Q(\bm \beta;\bm \beta)-\nabla Q_n(\bm \beta;\bm \beta)||_{2} \le \alpha.
\end{equation*}
\end{condition}

A difference between the high-dimensional and low-dimensional cases is that in the low-dimensional case, the dimension of $\bm \beta^*$, noted as $d$, can be much smaller than the sample size $n$, so rather than using the infinity norm, the statistical error can be directly measured in $\ell_2$ norm to reflect the accumulated difference between each index of the true $\bbeta^*$ and $\hat\bbeta$. 

\begin{condition}[Truncation-Error-2 $(\xi,\phi,n,T,\mathcal{B})$]\label{con:ldem2}
For any $ \bm \beta \in \mathcal{B}$, there exists a non-increasing function $\phi$, such that for the truncation level $T$, with probability $1-\phi(\xi)$, 
\[||\nabla Q_{n} (\bm \beta;\bm \beta)-f_T(\nabla Q_{n} (\bm \beta; \bm \beta))||_2 \le \xi. \]
\end{condition}

Below is the main theorem for the low-dimensional DP EM algorithm.
\begin{theorem}\label{thm:ldem}
 For the Algorithm~\ref{algo:ldem}, we define $\mathcal{B} = \{\bm \beta : ||\bm \beta - \bm \beta^*||_2 \le R\}$ and $||\bm \beta^0 - \bm \beta^*||_2 \le R/2$. We assume the Lipschitz-Gradient$(\gamma,\mathcal{B})$ and the Concavity-Smoothness$(\mu,\nu,\mathcal{B})$ hold. Define $\kappa = 1-\frac{2\nu-2\gamma}{\mu+\lambda} \in (0,1)$.
 For the parameters, we choose the step size $\eta=\frac{2}{\mu+\nu}$, the truncation level $T \asymp \sqrt{\log n}$, and the number of iterations $N_0 \asymp \log n$. We assume that $\alpha \le (\nu-\gamma)\cdot R/4$ and $\xi \le (\nu-\gamma)\cdot R/4$. For the sample size, there exists a constant $K$, and $n \ge K\cdot \frac{d(\log n)^{\frac{3}{2}} \sqrt{\log(1/\delta)}}{(1-\kappa)\cdot R \cdot \epsilon}$. Then, under the conditions Statistical-Error-2$(\alpha,\tau/{N_0},n/{N_0},\mathcal{B})$ and Truncation-Error-2$(\xi,\phi,n/N_0,T,\mathcal{B})$, there exists sufficient large constant $C$, such that it holds that with probability $1-c_0\log n\cdot \exp(-c_1 d) - c_2\phi(\xi)\cdot \log n - \tau$,
\begin{align}\label{eq:ldem}
     ||\bm\beta^t-\bm\beta^*||_2 &\le  \frac{\kappa^t}{2} R + C\cdot \frac{d\cdot (\log n)^{\frac{3}{2}} \sqrt{\log(1/\delta)}}{n\epsilon} + \frac{\eta\cdot(\xi+\alpha)}{1-\kappa}.
\end{align}
Specifically, for the output in Algorithm~\ref{algo:ldem}, it holds that
\begin{align*}
     ||\bm\beta^{N_0}-\bm\beta^*||_2 &\le C\cdot \frac{d\cdot (\log n)^{\frac{3}{2}} \sqrt{\log(1/\delta)}}{n\epsilon} + \frac{\eta\cdot(\xi+\alpha)}{1-\kappa}.
\end{align*}
\end{theorem}

The proof of Theorem~\ref{thm:ldem} is given in  Section~\ref{pthm:ldem}. There are three terms in the result (\ref{eq:ldem}), and their interpretations align with the high-dimensional cases. The first term is the optimization error, which converges to zero at a geometric rate. When the number of iterations $t$ is large, this term is small.
The second term is the cost of privacy caused by the Gaussian noise added in each iteration to achieve DP. When the privacy constraint becomes more stringent ($\epsilon,\delta$ become smaller), this term becomes larger. 
The third term reflects the truncation error and the statistical error. On one hand, by choosing an appropriate $T$, nearly every index is below the threshold $T$ in each iteration, so the truncation costs few accuracy loss. On the other hand, the statistical error is caused by the finite samples. With proper choice of $T$ and sufficiently large $n$, the third term can be quite small.


The result in Theorem~\ref{thm:ldem} is obtained for the general latent variables model. The convergence rates of $\alpha$ and $\xi$ are unspecified in this general case, which may vary according to different specific models. To show the theoretical guarantees of the proposed algorithm, as an example, we apply the algorithm to the Gaussian mixture model and leave the results of the other two specific models in the Appendix~\ref{pthm:ldmrm} and \ref{pthm:ldmcm}.

\begin{theorem}\label{thm:ldgmm}
For the Algorithm~\ref{algo:ldem} in the Gaussian mixture model, let the truncation of the gradient be the same as the truncation of gradient in Algorithm~\ref{algo:hdgmm}. Then, we define $\mathcal{B} = \{\bm \beta : ||\bm \beta - \bm \beta^*||_2 \le R\}$ and $||\bm \beta^0 - \bm \beta^*||_2 \le R/2$. Define $R,\mu,\nu,\gamma$ as in Lemma~\ref{lem:hdgmm} and $\kappa = \gamma$. For the choice of parameters, the step size is chosen as $\eta = 1$, the truncation level $T$ is chosen to be $T  \asymp \sqrt{\log n}$ and the number of iterations is chosen as $N_0 \asymp \log n$. For the sample size $n$, it is sufficiently large that there exists constants $K, K'$, such that $n \ge K\cdot \frac{d(\log n)^{\frac{3}{2}} \sqrt{\log(1/\delta)}}{(1-\kappa)\cdot R \cdot \epsilon}$ and $K' \cdot \sqrt{\frac{d}{n}} \cdot \log n \le (1-\gamma)\cdot R/4$. Then, 
there exists sufficient large constant $C$, such that it holds that with probability $1-c_0\log n\cdot \exp(-c_1 d) - c_2/\log n - c_3 \cdot n^{-1/2}$,
\begin{align*}
     ||\hat{\bm\beta}-\bm\beta^*||_2 &\le C\cdot \frac{d\cdot (\log n)^{\frac{3}{2}} \sqrt{\log(1/\delta)}}{n\epsilon} +\frac{\eta}{1-\kappa} \cdot \sqrt{\frac{d}{n}}\cdot \log n .\end{align*} 
\end{theorem}

We remark here that in the literature, \citep{wang2020differentially} also analyzed the Gaussian mixture model and obtained a $O(\sqrt{d^{2}/{n}}\cdot{\log(1/\delta)}/{\epsilon})$ rate of convergence. Our result in Theorem~\ref{thm:ldgmm} has faster rate of convergence than that.  In the following, we are going to show that such a rate cannot be improved further up to logarithm factors.

\begin{proposition}\label{thm:ldgmmlow}
Suppose $\bm Y = \{\bm y_1, \bm y_2,...\bm y_n\}$ be the data set of $n$ samples observed from the Gaussian mixture model and let $M$ be any algorithm such that $M \in \mathcal{M}_{\epsilon,\delta}$, where $\mathcal{M}_{\epsilon,\delta}$ be the set of all $(\epsilon, \delta)$-DP algorithms for the estimation of the true parameter $\bm \beta^*$ in the low-dimensional setting. Then there exists a constant $c$, if $0< \epsilon <1$ and $n^{-1}\exp(-n \epsilon)<\delta < n^{-(1+\omega)}$ for some fixed $\omega > 0$ with $d > c_0 \log(1/\delta)$ and $n> c_1 \cdot \sqrt{d \log(1/\delta)/\epsilon}$, we have
\[\inf_{M \in \mathcal{M}_{\epsilon,\delta}} \sup_{\bm \beta^* \in \mathbb{R}^d} \E\| {M(\bm Y)}-\bm \beta^*\|_2 \ge c \cdot (\sqrt{\frac{d}{n}} + \frac{d \sqrt{\log(1/\delta)}}{n \epsilon} ).\]
\end{proposition}
The detailed proofs of Theorem~\ref{thm:ldgmm} and Proposition~\ref{thm:ldgmmlow} are in Appendix~\ref{pthm:ldgmm}. By comparing the results in these two theorems, our proposed Algorithm~\ref{algo:ldem} reaches the minimax optimal rate of convergence up to logarithm factors. 



\section{Numerical Experiments}\label{sec:simulation} 
\noindent In this section, we investigate  the numerical performance of the proposed DP EM algorithms. Specifically, in the high-dimensional setting, for the illustration purpose, we investigate the Gaussian mixture model (Algorithm~\ref{algo:hdgmm}) in Section~\ref{sec6-1} on the simulated data sets in details. Due to space constraints, an additional simulation for Mixture of regression model is presented in Supplement materials~\ref{sec6-2}. 
Then, in the low-dimensional case, we compare our Algorithm~\ref{algo:ldem} with the algorithm in \citep{wang2020differentially} under the Gaussian mixture model in Section~\ref{sec6-3}. Further, in Section~\ref{sec6-4}, we demonstrate the numerical performance of the proposed algorithm on real datasets. 
\subsection{Simulation results for Gaussian mixture model} \label{sec6-1}
 
\begin{figure}[h]
 
\centering
\begin{minipage}[t]{0.32\textwidth}
\centering
\includegraphics[width=5.4cm]{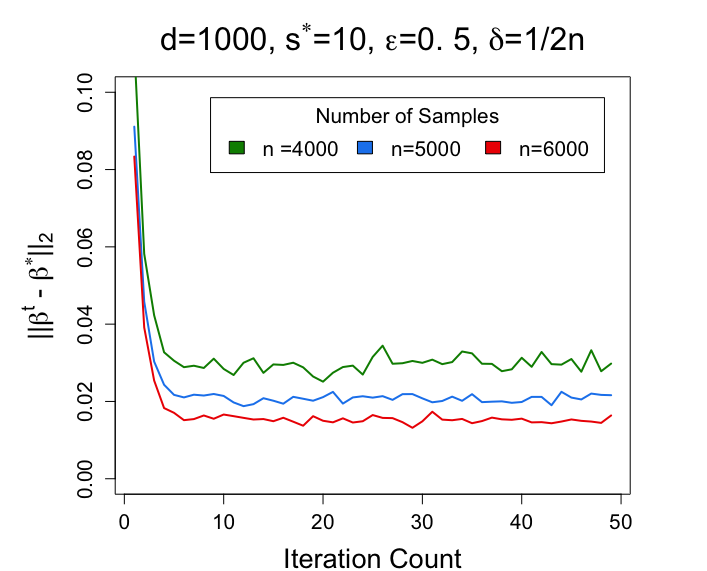}

\end{minipage}
\begin{minipage}[t]{0.32\textwidth}
\centering
\includegraphics[width=5.4cm]{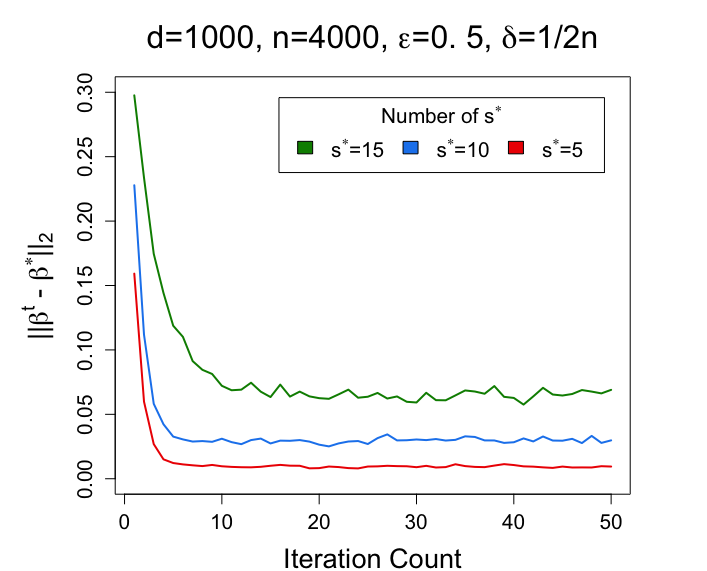}

\end{minipage}
\begin{minipage}[t]{0.32\textwidth}
\centering
\includegraphics[width=5.4cm]{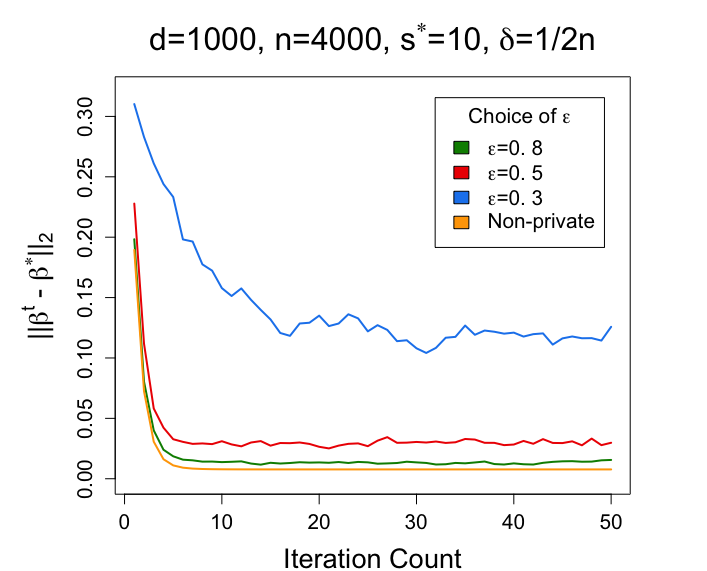}

\end{minipage}
\caption{The average estimation error under different settings in the high-dimensional Gaussian mixture model.} \label{Fig:1}
\end{figure}

\noindent For the DP EM algorithm in the  high-dimensional Gaussian mixture model, the simulated data set is constructed as follows. First, we set $\bm \beta^*$ to be a unit vector, where the first $s^*$ indices equal to $1/\sqrt{s^*}$ and the rest of the indices are  zero. For $i\in[n]$, we generate $\bm y_i = z_i \cdot \bm \beta^* + \bm e_i$, where $\Pro(z_i = 1) =\Pro(z_i =- 1) = 1/2 $ and  $\bm e_i \sim N(0, \sigma^2 I_d)$ with  $\sigma= 0.5$.
We consider the following three settings:
\begin{enumerate}
    \item Fix $d=1000, s^*=10, \epsilon =0.5, \delta = (2n)^{-1}$. Compare the results of Algorithm~\ref{algo:hdgmm} when $n=4000,5000,6000$, respectively. 
    \item Fix $n=4000, d=1000, \epsilon =0.5, \delta = (2n)^{-1}$. Compare the results of Algorithm~\ref{algo:hdgmm} when $s^* =5,10,15$, respectively. 
    \item Fix $n=4000, d=1000, s^*=10, \delta = (2n)^{-1}$. Compare the results of Algorithm~\ref{algo:hdgmm} when $\epsilon = 0.3, 0.5, 0.8$, respectively. 
\end{enumerate}
For each setting, we repeat the experiment for 50 times and report the average  of the estimation error $\|\bbeta^t - \bbeta^*\|_2$. For each experiment, we set the step size $\eta$ to be 0.5. The results are plotted in Figure~\ref{Fig:1}. 


From Figure~\ref{Fig:1}, we can clearly discover the relationship between different choices of $n, s^*, \epsilon$ and the performance of proposed  algorithm in the Gaussian mixture model.  The left figure in Figure~\ref{Fig:1} shows that  the estimation of $\bbeta^*$ becomes more accurate when $n$ becomes larger.  The middle figure in Figure~\ref{Fig:1} shows that when $s^*$ becomes smaller, the estimation error becomes smaller.  The right figure in Figure~\ref{Fig:1} shows that when $\epsilon$ becomes larger (the privacy constraints are more relaxed), the cost of privacy becomes smaller, and therefore the estimator achieves a smaller estimation error. With the $\epsilon$ becomes large enough, the estimator $\hat{\bbeta}$ becomes closer to the non-private setting. 

\subsection{Comparison with other algorithms}\label{sec6-3}
In the literature, \citep{wang2020differentially} also studied the DP EM algorithm in the classic low-dimensional latent variable models. In this section, we compare our proposed method with 1). the algorithm proposed in \citep{wang2020differentially}, and 2). the standard (non-private) gradient EM algorithm \citep{balakrishnan2017statistical}. 


The synthetic data is generated as follows. We first set the true parameter $\bbeta^*$ to be a unit vector with each element equal to $1/\sqrt{d}$. Then we simulate the Gaussian mixture model with $z_i$ satisfying $\Pro(z_i = 1) = \Pro(z_i =- 1) = 1/2 $ and  sample a multivariate Gaussian variable $\bm e_i \sim N(0, \sigma^2 I_d)$ with $\sigma= 0.5$. At last, we compute $\bm y_i = z_i \cdot \bm \beta^* + \bm e_i$. 

\begin{figure}[!htb]
\centering
\begin{minipage}[t]{0.40\textwidth}
\centering
\includegraphics[width=5.5cm]{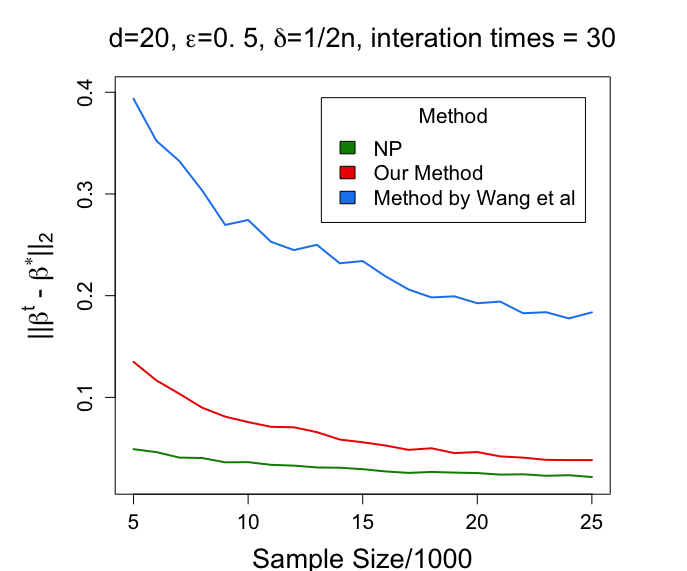}
\end{minipage}
\begin{minipage}[t]{0.40\textwidth}
\centering
\includegraphics[width=5.5cm]{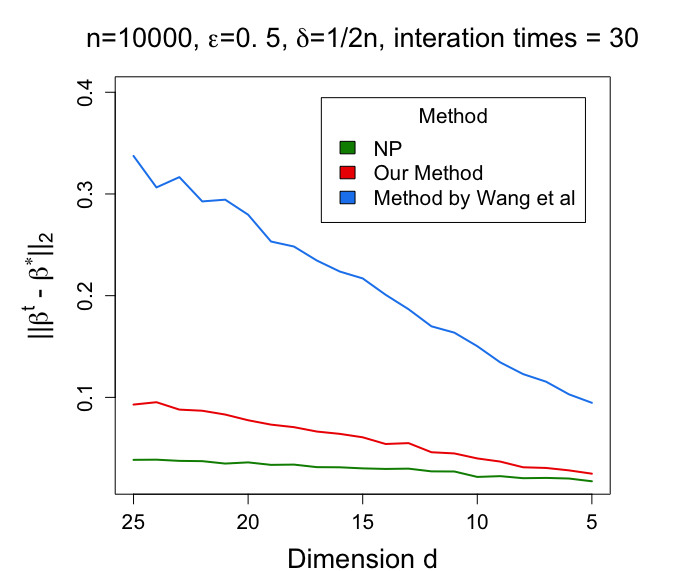}
\end{minipage}

\caption{The average estimation error under different settings in the classic low-dimensional Gaussian mixture model.}\label{fig:comparison}
\end{figure}

We consider the following two settings. In the first setting, we fix $d, \epsilon,\delta$ and vary $n$ from 5000 to 25000; in the second setting, we fix $n,\epsilon,\delta$  and vary $d$ from 5 to 25. For each setting, we report the average estimation error $\|\bbeta^* - \hat{\bbeta}\|_2$ among 50 repetitions.  The simulation results are summarized in Figure~\ref{fig:comparison}. The results indicate that although there is always a gap between our algorithm with the non-private EM algorithm (due to the cost of privacy), our algorithm has a much smaller error than that produced by that in \citep{wang2020differentially}.

\subsection{Real data analysis}\label{sec6-4}
In this section, we apply the proposed DP EM algorithm for high-dimensional Gaussian mixture to the Breast Cancer Wisconsin (Diagnostic) Data Set, which is available at the UCI Machine Learning Repository (\url{https://archive.ics.uci.edu/ml/datasets/Breast+Cancer+Wisconsin+(Diagnostic)}) \citep{Dua:2019}. This data set contains 569 instances and 30 attributes. Each instance describes the diagnostic result for an individual as `Benign' or `Malignant'. In the dataset, there are 212 instances labeled as `Malignant' while the rest 357 instances labeled as `Benign'. Such a medical diagnose data set often contains personal sensitive information and serves as a suitable data set to apply our privacy-preserving algorithms. 

In our experiment, all the attributes are normalized to have zero mean and unit variance. Moreover, to make the dataset symmetric, we first 
 randomly drop out 145 data points in the `Benign' class and compute the overall sample mean. 
Then for each data point, we subtract the overall sample mean from it.  
After preprocessing, the data is randomly split into two parts, where 70\% of the instances are used for training and 30\% of instances are used for testing.

In the training stage, we estimate the parameter $\bm \beta^*$ using the proposed algorithm for high-dimensional Gaussian mixture model (Algorithm~\ref{algo:hdgmm}). We run Algorithm~\ref{algo:hdgmm} for 50 iterations with step size $\eta = 0.5$. The initialization $\bm \beta^0$ is chosen as the unit vector where all indexes equal $1/\sqrt{30}$. We fix $\delta = 1/2n$, and choose various sparsity level $\hat{s}$ and privacy parameter $\epsilon$ as displayed in Table~\ref{fig:table}.  In the testing stage, we classify each testing data as `Benign' or `Malignant' according to the $\ell_2$ distance closeness between its attributes and $\hat{\bm \beta}$ with the distance between its attributes and $-\hat{\bm \beta}$. Then we compute the misclassification rate by comparing with the true diagnostic outcome. For each choice of parameters, we repeat the whole training and testing stages for 50 times, and report the average misclassification rate and the standard error. To compare with the non-private setting, for each choice of parameters, we also use the Algorithm described in \citep{wang2015high} as the baseline for the non-private setting (shown as $\epsilon = \infty$ in the table below). The results are summarized in the following Table~\ref{fig:table}.

\begin{figure}[H]
\centering
	\begin{tabular}{|c|c|c|c|}
			\hline
			& $\hat s = 5$     & $\hat s=10$       & $\hat s=15$      \\ \hline
			$\varepsilon = 0.2$  & 0.14(.07) & 0.12(.05) & 0.10(.04) \\ \hline
			$\varepsilon = 0.5$ & 0.08(.02) & 0.07(.02) & 0.07(.01) \\ \hline
			$\varepsilon = \infty$         & 0.07(.02) & 0.06(.02) & 0.06(.01) \\ \hline
	\end{tabular}
\caption{The average and standard error of misclassification rates of Algorithm \ref{algo:hdgmm} for the Breast Cancer Wisconsin Data Set.}\label{fig:table}
\end{figure}
The results suggest that when the privacy requirements become more stringent,  the classification accuracy drops in a mild way. 
Considering the significance to achieve privacy guarantees, such loss of accuracy could be acceptable.  

\section{Conclusion}\label{sec:conclusion} 
\noindent In this paper, we introduce a novel DP EM algorithm in both high-dimensional and low-dimensional settings. In the high-dimensional setting, we propose an algorithm based on noisy iterative hard thresholding and show this method is minimax rate-optimal up to logarithm factors in three specific models: Gaussian mixture, mixture of regression, and regression with missing covariates. In the low-dimensional setting, an algorithm based on Gaussian mechanism is also developed and shown to be near minimax optimal. 

\bibliographystyle{plainnat}

{\linespread{1.2}\selectfont\bibliography{reference}}
\newpage
\setcounter{page}{1}
\appendix
 
\section{Supplement materials}\label{sec:supp} 
\subsection{DP Algorithm and theories for Mixture of Regression Model in high-dimensional settings}\label{supp:mrm}

By applying Algorithm~\ref{algo:hdem}, the DP estimation algorithm in the high-dimensional mixture of regression is presented in the following in details: 

\begin{algorithm}[H]\label{algo:hdmrm}
\caption{DP Algorithm for High-Dimensional Mixture of Regression Model}
\begin{algorithmic}[1] 
\Require Private parameters $(\epsilon,\delta)$, step size $\eta$, truncation level $T$, maximum number of iterations $N_0$, sparsity parameter $\hat{s}$. \\
\textbf{Initialization:} $\bm \beta^0$ with $||\bm \beta^0||_0 \le \hat{s}$.\\
\textbf{For} $t = 0,1,2,...N_0-1$:\\
\quad Compute $\bm \beta^{t+0.5} = \bm \beta^{t} + \eta \cdot f_{T}(\nabla Q_{n/N_0}(\bm \beta^t; \bm \beta^t)) $.\\
\quad Let $\bm \beta^{t+1} = \text{NoisyHT}(\bm \beta^{t+0.5},\hat{s},4\eta\cdot T^2\cdot N_0/n,\epsilon,\delta)$.\\
\textbf{End For}

\Ensure $\hat{\bm \beta} = \bm \beta^{N_0}$
\end{algorithmic}
\end{algorithm}
For the truncation step $f_{T}(\nabla Q_n(\bm \beta^t;\bm \beta^t))$  on line 3 of Algorithm~\ref{algo:hdmrm}, rather than using truncation on the whole gradient, we perform the truncation on $y_i$, $\bm x_i$ and $\bm x_i^\top\bm \beta$ respectively, which leads to a more refined analysis and improved rate in the statistical analysis. Specifically, we define
\[f_{T}(\nabla Q_n(\bm \beta;\bm \beta)) = \frac{1}{n} \sum_{i=1}^n[2 w_{\bm \beta}(\bm x_i, y_i)\cdot \Pi_{T}(y_i)\cdot\Pi_{T}(\bm x_i)-\Pi_{T}(\bm x_i)\cdot \Pi_{T}(\bm x_i^\top\bm \beta)].\]

Then, we could verify Conditions~\ref{con: hdem1}-\ref{con:hdem4} in the mixture of regression model. 

\begin{lemma}\label{lem:hdmrm}
Suppose we can always find a sufficiently large constant $\phi$ to be the lower bound of the signal-to-noise ratio, $||\bm \beta^*||_2/\sigma > \phi$. Then
\begin{itemize}
    \item For Condition~\ref{con: hdem1}, Lipschitz-Gradient $(\gamma, \mathcal{B})$ condition and Condition~\ref{con:hdem2}, Concavity-Smoothness $(\mu,\nu,\mathcal{B})$ condition, both conditions hold with the parameters
\[\gamma \in (0,1/4), \mu = \nu =1, \mathcal{B} = \{\bm \beta:||\bm \beta-\bm \beta^*||_2 \le R\} \text{ with } R = 1/32 \cdot ||\bm \beta^*||_2.\]

\item For Condition~\ref{con:hdem3}, the condition Statistical-Error$(\alpha, \tau, \hat{s}, n, \mathcal{B})$ holds with a constant $C$ and
\[\alpha = C \cdot \eta \cdot \max(||\bm \beta^*||_{2}^2+\sigma^2,1,\sqrt{\hat{s}}\cdot||\bm \beta^*||_2)\cdot \sqrt{\frac{\log d + \log(4/\tau)}{n}}.\]

\item For Condition~\ref{con:hdem4}, the condition Truncation-Error $(\xi, \phi, \hat{s}, n/N_0, T, \mathcal{B})$ holds with $T \asymp \sqrt{\log n}$ and with probability $1-m_0/\log d \cdot \log n$, there exists a constant $C_1$, such that
\[||\nabla Q_{n/N_0} (\bm \beta;\bm \beta)-f_T(\nabla Q_{n/N_0} (\bm \beta; \bm \beta))||_2 \le C_1 \cdot \sqrt{\frac{s^* \cdot \log d }{n}}\cdot \log n.\]

\end{itemize}

\end{lemma}

The detailed proof of Lemma~\ref{lem:hdmrm} is given in Appendix~\ref{plem:hdmrm}. 

\subsection{DP Algorithm and theories for Regression with missing covariates Model in high-dimensional settings}\label{supp:mcm}

By applying Algorithm~\ref{algo:hdem}, we present in the following the DP estimation in the high-dimensional regression with missing covariate model.

\begin{algorithm}[H]\label{algo:hdmcm}
\caption{DP Algorithm for High-Dim Regression with Missing Covariates}
\begin{algorithmic}[1] 
\Require Private parameters $(\epsilon,\delta)$, step size $\eta$, truncation level $T$, maximum number of iterations $N_0$, sparsity parameter $\hat{s}$. \\
\textbf{Initialization:} $\bm \beta^0$ with $||\bm \beta^0||_0 \le \hat{s}$.\\
\textbf{For} $t = 0,1,2,...N_0-1$:\\
\quad Compute $\bm \beta^{t+0.5} = \bm \beta^{t} + \eta \cdot f_{T}(Q_{n/N_0}(\bm \beta^t ;\bm \beta^t) )$.\\
\quad Let $\bm \beta^{t+1} = \text{NoisyHT}(\bm \beta^{t+0.5},\hat{s},6\eta\cdot T^2\cdot N_0/n,\epsilon,\delta)$.\\
\textbf{End For}

\Ensure $\hat{\bm \beta} = \bm \beta^{N_0}$
\end{algorithmic}
\end{algorithm}
 For the term $f_{T}(\nabla Q_n(\bm \beta^t;\bm \beta^t))$ on line 3 of Algorithm 5, we design a truncation step specifically for this model. According to the definition of $m_{\bm \beta}(\tilde{\bm x}_i,y_i)$, when $\bm \beta$ is close to $\bm \beta^*$, we find that $m_{\bm \beta}(\tilde{\bm x}_i,y_i)$ is close to $\bm z_i \odot \bm x_i$, so we propose to truncate on $m_{\bm \beta}(\tilde{\bm x}_i,y_i)$ and $m_{\bm \beta}(\tilde{\bm x}_i,y_i)^\top \cdot \bm \beta$. 
 Specifically, let
\begin{align*}
    f_{T}(\nabla Q_n(\bm \beta;\bm \beta)) &= \frac{1}{n} \sum_{i=1}^n [\Pi_T (y_i) \cdot \Pi_{T} (m_{\bm \beta}(\tilde{\bm x}_i ,y_i)) - \text{diag}(\bm 1 - \bm z_i) \cdot \bm \beta \\&- \Pi_{T} (m_{\bm \beta}(\tilde{\bm x}_i,y_i))\cdot\Pi_T( m_{\bm \beta}(\tilde{\bm x}_i,y_i)^\top \cdot \bm \beta) \\
    &+ \Pi_{T}((\bm 1 - \bm z_i) \odot m_{\bm \beta}(\tilde{\bm x}_i,y_i)) \cdot \Pi_{T}(( (\bm 1 - \bm z_i) \odot m_{\bm \beta}(\tilde{\bm x}_i,y_i))^\top \cdot \bm \beta).
\end{align*}

Then, we could verify Conditions~\ref{con: hdem1}-\ref{con:hdem4} in the Regression with missing covariates model. 

\begin{lemma}\label{lem:hdmcm}
Suppose the signal-to-noise ratio $||\bm \beta^*||_2/\sigma \le r$ with $r >0$ be some constant. Also, for the probability $p$ that each coordinate of $\bm x_i$ is missing, we have $p < 1/(1+2b+2b^2)$ with $b = r^2 \cdot (1+L)^2$ and $L \in (0,1)$ be a constant. 
\begin{itemize}
    \item Condition~\ref{con: hdem1}, Lipschitz-Gradient $(\gamma, \mathcal{B})$ condition and Condition~\ref{con:hdem2}, Concavity-Smoothness $(\mu,\nu,\mathcal{B})$ condition hold with the parameters
\[\gamma  = \frac{b+ p\cdot(1+2b+2b^2)}{1+b} < 1, \mu = \nu =1, \mathcal{B} = \{\bm \beta:||\bm \beta-\bm \beta^*||_2 \le R\} \text{ with } R = L \cdot ||\bm \beta^*||_2.\]

\item For Condition~\ref{con:hdem3}, the condition Statistical-Error$(\alpha, \tau, \hat{s}, n, \mathcal{B})$ holds with a constant $C$ and
\[\alpha = C \cdot \eta \cdot [\sqrt{\hat{s}} \cdot \|\bm \beta^*\|_2^2\cdot(1+L)\cdot(1+L\cdot r)^2+\max(||\bm \beta^*||_{2}^2+\sigma^2,(1+L\cdot r)^2)]\cdot \sqrt{\frac{\log d + \log(12/\tau)}{n}}.\]

\item For Condition~\ref{con:hdem4}, the condition Truncation-Error$(\xi, \phi, \hat{s}, n/N_0, T, \mathcal{B})$ holds with $T \asymp \sqrt{\log n}$ and with probability $1-m_0/\log d \cdot \log n$, there exists a constant $C_1$, such that
\[||\nabla Q_{n/N_0} (\bm \beta;\bm \beta)-f_T(\nabla Q_{n/N_0} (\bm \beta; \bm \beta))||_2 \le C_1 \cdot \sqrt{\frac{s^* \cdot \log d }{n}}\cdot \log n.\]

\end{itemize}

\end{lemma}

The detailed proof of Lemma~\ref{lem:hdmcm} is in the Appendix~\ref{plem:hdmcm}.

\subsection{Simulation results for mixture of regression model}\label{sec6-2}
\begin{figure}[h]
\centering
\begin{minipage}[t]{0.32\textwidth}
\centering
\includegraphics[width=5.4cm]{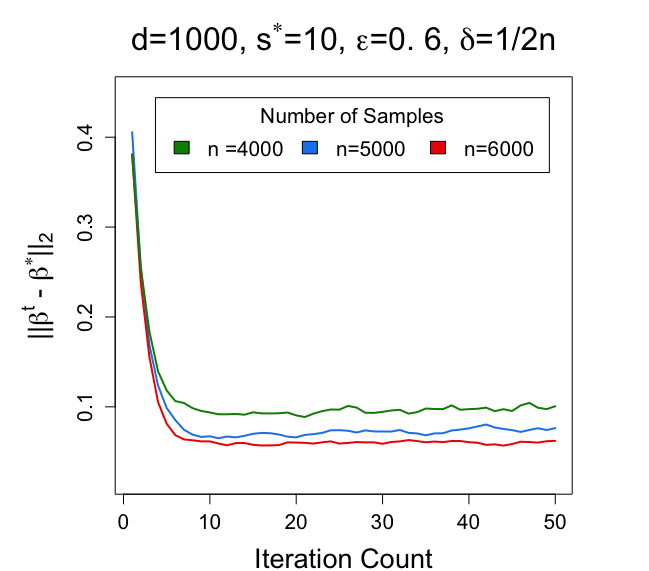}

\end{minipage}
\begin{minipage}[t]{0.32\textwidth}
\centering
\includegraphics[width=5.4cm]{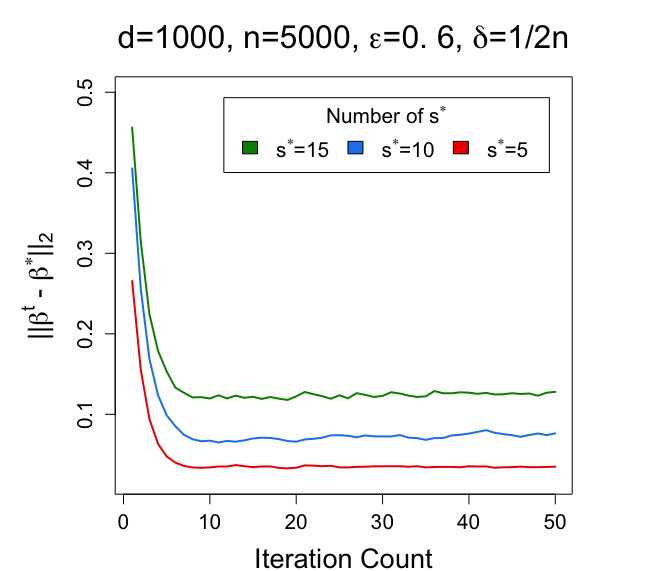}

\end{minipage}
\begin{minipage}[t]{0.32\textwidth}
\centering
\includegraphics[width=5.4cm]{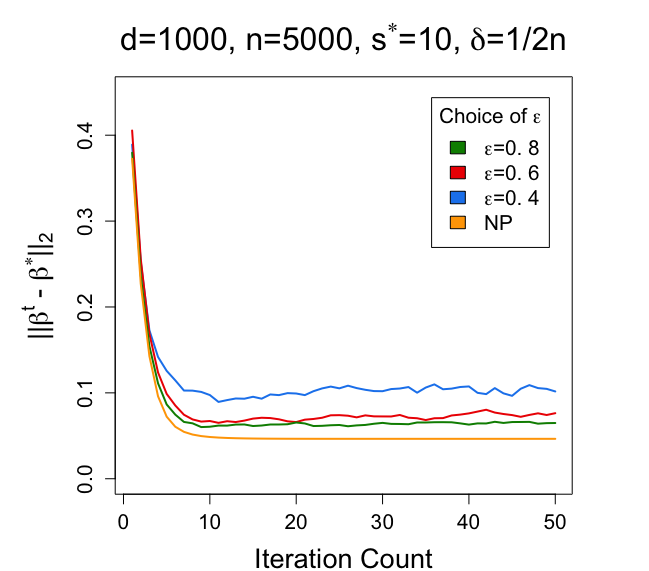}

\end{minipage}
\caption{The average estimation error under different settings under the high-dimensional mixture of regression model.}\label{Fig:2}
\end{figure}

\noindent For the DP EM algorithm in the high-dimensional mixture of regression model, the simulated dataset is constructed as follows. First, we set $\bm \beta^*$ to be a unit vector, where the first $s^*$ indices equal to $1/\sqrt{s^*}$ and the rest of the indices are  zero. For $i\in[n]$, we let $ y_i = z_i \cdot \bm x_i^\top  \bm \beta^* + e_i$, where $\bm x_i\sim N(0, \bm I_d)$, $\Pro(z_i = 1) =\Pro(z_i =- 1) = 1/2 $ and  $e_i \sim N(0, \sigma^2)$ with  $\sigma= 0.5$. We consider the following three experimental settings:
\begin{itemize}
    \item Fix $d=1000, s^*=10, \epsilon =0.6, \delta = (2n)^{-1}$. Compare the results of Algorithm~\ref{algo:hdmrm} when $n=4000,5000,6000$, respectively. 
    \item Fix $n=5000, d=1000, \epsilon =0.6, \delta = (2n)^{-1}$. Compare the results of Algorithm~\ref{algo:hdmrm} when $s^* =5,10,15$, respectively. 
    \item Fix $n=5000, d=1000, s^*=10, \delta = (2n)^{-1}$. Compare the results of Algorithm~\ref{algo:hdmrm} when $\epsilon = 0.4, 0.6, 0.8$, respectively. 
\end{itemize}
For each setting, we repeat the experiment for 50 times and report the average error $\|\bbeta^t - \bbeta^*\|_2$. For each experiment, the choice of initialized $\bbeta^0$ 
should be close to the true $\bbeta^*$ 
and the step size $\eta$ is set to be 0.5. The results are shown in Figure~\ref{Fig:2}. 

From the results in Figure~\ref{Fig:2}, we also discover the relationship between different choices of $n, s^*, \epsilon$ and the performance of the proposed DP EM algorithm on the mixture of regression model. Similar to the results under the Gaussian mixture model, with larger $n$, smaller $s^*$ and larger $\epsilon$, the estimator $\hat{\bbeta}$ has a smaller error in estimating $\bbeta^*$.

\subsection{Results for specific models in Low-dimensional settings}\label{sec:appendix2}
In this subsection, we will show results for DP algorithm for the mixture of regression model and regression with missing covariates in low-dimensional settings. we will list the algorithm and the theorem below. Since the proofs of these two specific models are highly similar to the proof of the high dimensional setting Theorem~\ref{thm:hdmrm}, Proposition~\ref{thm:hdmrmlow}, Theorem~\ref{thm:hdmcm}, Proposition~\ref{thm:hdmcmlow} and also the low dimensional setting for Gaussian mixture Model Theorem~\ref{thm:ldgmm} and Proposition~\ref{thm:ldgmmlow}, we omit the proofs here.

\subsubsection{Algorithm and theories for Mixture of Regression Model in low-dimensional settings }\label{pthm:ldmrm}
The algorithm is listed as below: 

\begin{algorithm}[H]\label{algo:ldmrm}
\caption{Low Dimensional Private EM algorithm on Mixture of Regression Model}
\begin{algorithmic}[1] 
\Require Private parameters $(\epsilon,\delta)$, step size $\eta$, truncation level $T$, maximum number of iterations $N_0$. \\
\textbf{Initialization:} $\bm \beta^0$\\
\textbf{For} $t = 0,1,2,...N_0-1$:\\
\quad Compute $\bm \beta^{t+1} = \bm \beta^{t} + \eta f_T(\nabla Q_{n/N_0}(\bm \beta^t;\bm \beta^t)) +W_t $. $W_t$ is a random vector $(\xi_1,\xi_2,...\xi_d)^\top$, where $\xi_1,\xi_2,...\xi_d$ are i.i.d sample drawn from $N(0, \frac{2 \eta^2 d (4T^2)^2 {N_0}^2 \log(1.25/\delta)}{n^2 \epsilon^2})$\\
\textbf{End For}

\Ensure $\hat{\bm \beta} = \bm \beta^{N_0}$

\end{algorithmic}

\end{algorithm}
where the truncation on the gradient is the same as the truncation in Algorithm~\ref{algo:hdmrm}
\begin{theorem}\label{thm:ldmrm}
For the Algorithm~\ref{algo:ldmrm} in the Mixture of Regression Model, we define $\mathcal{B} = \{\bm \beta : ||\bm \beta - \bm \beta^*||_2 \le R\}$ and $||\bm \beta^0 - \bm \beta^*||_2 \le R/2$. Define $R,\mu,\nu,\gamma$ as in Lemma~\ref{lem:hdmrm}  and $\kappa = \gamma$. For the choice of parameters, the step size is chosen as $\eta = 1$, the truncation level is chosen as $T \asymp \sqrt{\log n}$ and the number of iterations is chosen as $N_0 \asymp \log n$. For the sample size $n$, it is sufficiently large that there exists constants $K, K'$, such that $n \ge K\cdot \frac{d(\log n)^{\frac{3}{2}} \sqrt{\log(1/\delta)}}{(1-\kappa)\cdot R \cdot \epsilon}$ and $K' \cdot \sqrt{\frac{d}{n}} \cdot (\log n)^{3/2} \le (1-\gamma)\cdot R/4$. Then, 
there exists sufficient large constant $C$, it holds that with probability $1-c_0\log n\cdot \exp(-c_1 d) - c_2/ \log n - c_3 n^{-1/2}$:
\begin{align*}
     ||\hat{\bm\beta}-\bm\beta^*||_2 &\le  C\cdot \frac{d\cdot (\log n)^{\frac{3}{2}} \sqrt{\log(1/\delta)}}{n\epsilon} +\frac{\eta}{1-\kappa} \cdot \sqrt{\frac{d}{n}}\cdot (\log n)^{3/2}.
\end{align*} 
\end{theorem}

\begin{theorem}\label{thm:ldmrmlow}
 Suppose $( Y, \bm X) = \{( y_1, \bm x_1), ( y_2,\bm x_2),...( y_n, \bm x_n)\}$ be the data set of $n$ samples observed from the Mixture of Regression Model discussed above and let $M$ be any corresponding $(\epsilon, \delta)$-differentially private algorithm for the estimation of the true parameter $\bm \beta^*$. Then there exists a constant $c$, if $0< \epsilon <1$ and $\delta < n^{-(1+\omega)}$ for some fixed $\omega > 0$, we have:
\[\inf_{M \in \mathcal{M}_{\epsilon,\delta}} \sup_{\bm \beta^* \in \mathbb{R}^d} \mathbb{E}\|{M(Y, \bm X)}-\bm \beta^*\|_2 \ge c \cdot (\sqrt{\frac{d}{n}} + \frac{d}{n \epsilon} ).\]
\end{theorem}

\subsubsection{Algorithm and theories for Regression with Missing Covariates in low dimension cases}\label{pthm:ldmcm}
The algorithm is listed as below: 

\begin{algorithm}[H]\label{algo:ldmcm}
\caption{Low Dimensional Private EM algorithm on Regression with missing covariates}
\begin{algorithmic}[1] 
\Require Private parameters $(\epsilon,\delta)$, step size $\eta$, truncation level $T$, maximum number of iterations $N_0$. \\
\textbf{Initialization:} $\bm \beta^0$\\
\textbf{For} $t = 0,1,2,...N_0-1$:\\
\quad Compute $\bm \beta^{t+1} = \bm \beta^{t} + \eta f_T(\nabla Q_{n/N_0}(\bm \beta^t;\bm \beta^t)) +W_t $. $W_t$ is a random vector $(\xi_1,\xi_2,...\xi_d)^\top$, where $\xi_1,\xi_2,...\xi_d$ are i.i.d sample drawn from $N(0, \frac{2 \eta^2 d (6T^2)^2 {N_0}^2 \log(1.25/\delta)}{n^2 \epsilon^2})$\\
\textbf{End For}

\Ensure $\hat{\bm \beta} = \bm \beta^{N_0}$

\end{algorithmic}

\end{algorithm}
where the truncation on the gradient is the same as the truncation in Algorithm~\ref{algo:hdmcm}
\begin{theorem}\label{thm:ldmcm}
For Algorithm~\ref{algo:ldmcm}, we define $\mathcal{B} = \{\bm \beta : ||\bm \beta - \bm \beta^*||_2 \le R\}$ and $||\bm \beta^0 - \bm \beta^*||_2 \le R/2$. Define $R,\mu,\nu,\gamma$ as in Lemma~\ref{lem:hdmcm} and $\kappa = \gamma$. For the choice of parameters, the step size is chosen as $\eta = 1$, the truncation level is chosen to be $T \asymp \sqrt{\log n}$ and the number of iterations is chosen as $N_0 \asymp \log n$. For the sample size $n$, it is sufficiently large that there exists constants $K, K'$, such that $n \ge K\cdot \frac{d(\log n)^{\frac{3}{2}} \sqrt{\log(1/\delta)}}{(1-\kappa)\cdot R \cdot \epsilon}$ and $K' \cdot \sqrt{\frac{d}{n}} \cdot (\log n)^{3/2} \le (1-\gamma)\cdot R/4$. Then, 
there exists sufficient large constant $C$, it holds that with probability $1-c_0\log n\cdot \exp(-c_1 d) - c_2/ \log n - c_3 n^{-1/2}$:
\begin{align*}
     ||\hat{\bm\beta}-\bm\beta^*||_2 &\le  C\cdot \frac{d\cdot (\log n)^{\frac{3}{2}} \sqrt{\log(1/\delta)}}{n\epsilon} +\frac{\eta}{1-\kappa} \cdot \sqrt{\frac{d}{n}}\cdot (\log n)^{3/2}.
\end{align*} 
\end{theorem}

\begin{theorem}\label{thm:ldmcmlow}
 Suppose $( Y, \bm X) = \{( y_1, \bm x_1), ( y_2,\bm x_2),...( y_n, \bm x_n)\}$ be the data set of $n$ samples observed from the Regression with Missing covariates discussed above and let $M$ be any corresponding $(\epsilon, \delta)$-differentially private algorithm for the estimation of the true parameter $\bm \beta^*$. Then there exists a constant $c$, if $0< \epsilon <1$ and $\delta < n^{-(1+\omega)}$ for some fixed $\omega > 0$, we have:
\[\inf_{M \in \mathcal{M}_{\epsilon,\delta}} \sup_{\bm \beta^* \in \mathbb{R}^d} \mathbb{E}\|{M(Y, \bm X)}-\bm \beta^*\|_2 \ge c \cdot (\sqrt{\frac{d}{n}} + \frac{d}{n \epsilon} ).\]
\end{theorem}

Therefore, both the low-dimensioanl EM algorithm on Mixture of Regression Model and Regression with missing covariates attain a near-optimal rate of convergence up to a log factor.


\section{Proofs of main results}\label{sec:proof}
\subsection{Proof of Theorem~\ref{thm:hdem}}\label{pthm:hdem}
Assume that the step size $\eta$ is chosen, by Lemma~\ref{lem:noisyht3}, we can claim the privacy is guaranteed. Then we can start the proof. For simplicity, we denote $n_0=n/N_0$. During the $t$-th iteration, we can write the iterated two steps in the following way:
\begin{equation} 
    \bm \beta^{t+0.5} = \bm \beta^{t} + \eta \cdot f_T(\nabla Q_{n_0}(\bm \beta^t;\bm \beta^t)), \quad
    \bm \beta^{t+1} = \text{trunc}(\bm \beta^{t+0.5},\mathcal{S}^{t+0.5}) + \bm W_{Lap}^{t}.
\end{equation}
where $\mathcal{S}^{t+0.5}$ is the set of indices selected by the private peeling algorithm, and $ \bm W_{Lap}^{t}$ is the vector of Laplace noises with support on $\mathcal{S}^{t+0.5}$. Furthermore, let us introduce the following notations, we define:
\begin{equation}
    \bar{\bm \beta}^{t+0.5} = \bm \beta^{t} + \eta \cdot \nabla Q(\bm \beta^t;\bm \beta^t), \quad
    \bar{\bm \beta}^{t+1} = \text{trunc}(\bar{\bm \beta}^{t+0.5},\mathcal{S}^{t+0.5}).
\end{equation}
Then, before continue the proof, we will first introduce two lemmas:
\begin{lemma}\label{lem:phdem1} 
 Suppose we have
\begin{equation} \label{peq1}
    \|\bar{\bm \beta}^{t+0.5} -\bm \beta^*\|_2 \le L \|\bm \beta^*\|_2,
\end{equation}
for some $L \in (0,1)$. Also, for the sparsity level, we have that $\hat{s} \ge \frac{4 \cdot (1+L)^2}{(1-L)^2} \cdot s^*$
We assume that $\alpha = o(1)$ and also $\sqrt{\hat{s}} \cdot \alpha \le \frac{(1-L)^2}{2\eta(1+L)} \cdot ||\bm \beta^*||_2$. We further assume $\frac{s^* \log d}{n} \cdot (\log n)^{\frac{3}{2}} = o(1)$. Then, it holds that, there exists a constant $K_1 > 0$:
\begin{equation}
     \|\bar{\bm \beta}^{t+1}-\bm \beta^*\|_2\leq (1+4\sqrt{{s^*}/\hat{s}})^{\frac{1}{2}}\|\bar{\bm \beta}^{t+0.5}-\bm \beta^*\|_2+\frac{K_1\sqrt{s^*}}{\sqrt{1-L}}{\cdot\eta\cdot\alpha}.
\end{equation}
 with probability $1-\tau-N_0\phi(\xi)-m_0\cdot s^* \cdot \log n\cdot\exp(-m_1 \log d)$, for all $t$ in $0,1,2,...,N_0-1$.
\end{lemma}

\begin{lemma}\label{lem:phdem2} 
For $\nu$, $\mu$ and $\gamma$ defined in the Theorem~\ref{thm:hdem}, the following inequality holds:
\begin{equation}
    ||\bar{\bm \beta}^{t+0.5}-\bm \beta^*||_2 \le (1-2\cdot\frac{\nu-\gamma}{\nu+\mu})\cdot  ||{\bm \beta}^{t}-\bm \beta^*||_2.
\end{equation}
\end{lemma}
The detailed proof of Lemma~\ref{lem:phdem1} is in Appendix~\ref{plem:phdem1} and the proof for Lemma~\ref{lem:phdem2} is in  Lemma 5.2 from \citep{wang2015high}. Then, by the two lemmas discussed above:
\begin{align}{\label{peq22}}
    &\|\bm \beta^{t+1} -\bm \beta^*\|_2 \notag\\
    &\le \|\text{trunc}(\bm \beta^{t+0.5},\mathcal{S}^{t+0.5})-\bm \beta^*\|_2 + \| \bm W_{Lap}^{t}\|_2 \notag\\
    &\le  \|\text{trunc}(\bm \beta^{t+0.5},\mathcal{S}^{t+0.5})-\text{trunc}(\bar{\bm \beta}^{t+0.5},\mathcal{S}^{t+0.5})\|_2 + \|\text{trunc}(\bar{\bm \beta}^{t+0.5},\mathcal{S}^{t+0.5})-\bm \beta^*\|_2 + \| \bm W_{Lap}^{t}\|_2 \notag\\
    &= \underbrace{\|\text{trunc}(\bm \beta^{t+0.5},S^{t+0.5})- \text{trunc}(\bar{\bm \beta}^{t+0.5},\mathcal{S}^{t+0.5})\|_2}_{(1)} + \underbrace{\| \bar{\bm \beta}^{t+1}-\bm \beta^*\|_2}_{(2)} + \| \bm W_{Lap}^{t}\|_2.
\end{align}
First, we notice that for the term (1) in (\ref{peq22}):
\begin{align}{\label{peq23}}
    &\|\text{trunc}(\bm \beta^{t+0.5},\mathcal{S}^{t+0.5})- \text{trunc}(\bar{\bm \beta}^{t+0.5},\mathcal{S}^{t+0.5})\|_2 \notag\\
    &= \|(\eta f_T(\nabla Q_{n_0}(\bm \beta^t;\bm \beta^t)) - \eta \nabla Q(\bm \beta^t;\bm \beta^t))_{\mathcal{S}^{t+0.5})}\|_2 \notag\\
    &\le \|(\eta f_T(\nabla Q_{n_0}(\bm \beta^t;\bm \beta^t)) - \eta \nabla Q_{n_0}(\bm \beta^t;\bm \beta^t))_{\mathcal{S}^{t+0.5})}\|_2 +\|(\eta \nabla Q_{n_0}(\bm \beta^t;\bm \beta^t) - \eta \nabla Q(\bm \beta^t;\bm \beta^t))_{\mathcal{S}^{t+0.5})}\|_2 \notag \\
    &\le \eta\|f_T(\nabla Q_{n_0}(\bm \beta^t;\bm \beta^t)) - \nabla Q_{n_0}(\bm \beta^t;\bm \beta^t)\|_2 + \eta \sqrt{\hat{s}}\cdot \| \nabla Q_{n_0}(\bm \beta^t;\bm \beta^t) - \nabla Q(\bm \beta^t;\bm \beta^t)\|_\infty\notag\\
    &\le  \eta\|f_T(\nabla Q_{n_0}(\bm \beta^t;\bm \beta^t)) - \nabla Q_{n_0}(\bm \beta^t;\bm \beta^t)\|_2 + \eta \sqrt{\hat{s}}\cdot \alpha.
\end{align}
Second, for the term (2) in (\ref{peq22}), by Lemma~\ref{lem:phdem1} and Lemma~\ref{lem:phdem2}, we have:
\begin{align}{\label{peq24}}
   \|\bar{\bm \beta}^{t+1}-\bm \beta^*\|_2 &\le (1+4\sqrt{{s^*}/\hat{s}})^{\frac{1}{2}}\|\bar{\bm \beta}^{t+0.5}-\bm \beta^*\|_2+\frac{K_1\sqrt{s^*}}{\sqrt{1-L}}{\cdot\eta\cdot\alpha} \notag \\
   &\le (1+4\sqrt{{s^*}/\hat{s}})^{\frac{1}{2}}\cdot (1-2\cdot\frac{\nu-\gamma}{\nu+\mu})\cdot  ||{\bm \beta}^{t}-\bm \beta^*||_2+\frac{K_1\sqrt{s^*}}{\sqrt{1-L}}{\cdot\eta\cdot\alpha}.
\end{align}
Third, for the term  $\| \bm W_{Lap}^{t}\|_2$, notice that for any $i = 1,2,...d$, $ \{W_{Lap}^{t}\}_{i} \sim Laplace(\lambda_0)$,  by the concentration of Laplace Distribution, for every $C > 1$, we have:
\begin{equation}
    \Pro(\| W_{Lap}^{t}\|_2^2 > \hat{s} C^2 \lambda_0^2) \le \hat{s} e^{-C}.
\end{equation}
here, in this case, $\lambda_0 = C'  \sqrt{\log n}\sqrt{\hat{s} \log(1/\delta)}/(n_0 \cdot \epsilon)$, let $C \asymp \log d$, we have there exists a constant $C_1, m_2, m_3$, such that with probability $1-m_2\cdot s^* \cdot \exp(-m_3 \log d)$, we have:
\begin{equation}
    \| \bm W_{Lap}^{t}\|_2^2 \le C_1 \cdot \frac{(s^* \log d)^2 \log(1/\delta) \log n}{n_0^2 \epsilon^2}.
\end{equation}
By a union bound and the choice of $N_0$ to be $N_0 \asymp \log n$, we find that with probability $1-m_2 \cdot s^* \cdot\log n\cdot \exp(-m_3 \log d)$, we have:
\begin{equation}
    \max_{t} \|\bm  W_{Lap}^{t}\|_2^2 \le C_1 \cdot \frac{(s^* \log d)^2 \log(1/\delta) \log n}{n_0^2 \epsilon^2}.
\end{equation}
According to our assumption, we find that $ \| \bm W_{Lap}^{t}\|_2$ is o(1). Thus, by combining the results from (\ref{peq22})(\ref{peq23})(\ref{peq24}), we find that: 
\begin{align}
    \|\bm \beta^{t+1} -\bm \beta^*\|_2 \notag
    &\le \eta \sqrt{\hat{s}}\cdot \alpha + (1+4\sqrt{{s^*}/\hat{s}})^{\frac{1}{2}}\cdot (1-2\cdot\frac{\nu-\gamma}{\nu+\mu})\cdot  ||{\bm \beta}^{t}-\bm \beta^*||_2+\frac{K_1\sqrt{s^*}}{\sqrt{1-L}}{\cdot\eta\cdot\alpha}\\
    &+\eta\|f_T(\nabla Q_{n_0}(\bm \beta^t;\bm \beta^t)) - \nabla Q_{n_0}(\bm \beta^t;\bm \beta^t)\|_2  +  \| \bm W_{Lap}^{t}\|_2.
\end{align}
Denote $\kappa = 1-2\cdot\frac{\nu-\gamma}{\nu+\mu}$. Then, if $\beta^t \in \mathcal{B}$, by our assumption in the theorem that $\hat{s} \ge 16 \cdot (1/\kappa -1)^{-2} \cdot s^*$, we have $(1+4\cdot\sqrt{s^*/\hat{s}})^{1/2} \le 1/\sqrt{\kappa}$. thus we have $(1+4\sqrt{{s^*}/\hat{s}})^{\frac{1}{2}}\cdot (1-2\cdot\frac{\nu-\gamma}{\nu+\mu}) \le \sqrt{\kappa}$.

On the other hand, we find that by the assumption we also have $(\sqrt{\hat{s}} + \frac{K_1 \sqrt{s^*}}{\sqrt{1-L}})\cdot \eta \cdot \alpha \le (1-\sqrt{\kappa})^2 \cdot R$. Further more, by the assumptions, we also have that$\| W_{Lap}^{t}\|_2 $ are $o(1)$ for any $t$, and with probability $1-N_0\cdot \phi(\xi)$,  $\max_t \|f_T(\nabla Q_{n_0}(\bm \beta^t;\bm \beta^t)) - \nabla Q_{n_0}(\bm \beta^t;\bm \beta^t)\|_2 <\xi$ where $\xi = o(1)$. 
\begin{align}
    \|\bm \beta^{t+1} -\bm \beta^*\|_2 \le (1-\sqrt{\kappa})^2 R + \sqrt{\kappa} R \le R,
\end{align}
which guarantees that if $\bm \beta^t \in \mathcal{B}$, we also have $\bm \beta^{t+1} \in \mathcal{B}$. Iterate this, we can get the connection between $\|\bm \beta^{t} -\bm \beta^*\|_2$ and $\|\bm \beta^{0} -\bm \beta^*\|_2$:
\begin{align}
    \|\bm \beta^{t} -\bm \beta^*\|_2 \notag
    &\le \frac{(\sqrt{\hat{s}}+K_1/\sqrt{1-L} \cdot \sqrt{s^*})\cdot \eta}{1-\sqrt{\kappa}}\cdot \alpha + \kappa^{t/2}\cdot  ||{\bm \beta}^{t}-\bm \beta^*||_2\\
    &+\eta\cdot\xi/(1-\sqrt{\kappa})+  K_2 \cdot \frac{s^* \log d \cdot \sqrt{\log(1/\delta)} (\log n)^{3/2}}{n \epsilon},
\end{align}
which finishes the proof for the theorem. \hfill$\square$
\subsection{Proof of Theorem~\ref{thm:hdgmm}}\label{pthm:hdgmm}
\noindent The proof of theorem~\ref{thm:hdgmm} consists of three parts. The first part is the privacy guarantees. The second part is to verify the conditions for $\alpha$ is satisfied. The third part is to find the convergence rate of $||\nabla Q_{n/N_0} (\bm \beta;\bm \beta) - f_T(\nabla Q_{n/N_0} (\bm \beta;\bm \beta))||_2$, for any $\bm \beta$ on the iteration path. Let us begin the proof with the first part. \\\\
For the privacy guarantee, notice that for two adjacent data sets, let $y_i$ and ${y_i}'$ be the difference between these two data sets. 
\begin{align}
    &||f_T(\nabla Q_{n/N_0} (\bm \beta;\bm \beta)) -f_T(\nabla {Q_{n/N_0}}' (\bm \beta;\bm \beta))||_{\infty} \notag\\
    &= ||\frac{N_0}{n}\cdot[2 w_{\bm \beta}(y_i)-1]\cdot\Pi_{T}(y_i) - \frac{N_0}{n}\cdot[2 w_{\bm \beta}({y_i}')-1]\cdot\Pi_{T}({y_i}')||_{\infty} \notag\\
    &\le \frac{N_0}{n}\cdot [||\Pi_{T}(y_i)||_{\infty} +||\Pi_{T}({y_i}')||_{\infty} ] \notag\\
    &\le \frac{2T\cdot N_0}{n}.
\end{align}
Then, by Lemma~\ref{lem:noisyht3}, we can claim that the private Gaussian Mixture Model is $(\epsilon,\delta)$-differentially private. 

Also, for the conditions of $\alpha$ is Theorem~\ref{thm:hdem}, we find that $(\sqrt{\hat{s}}+c_1/\sqrt{1-L}\cdot\sqrt{s^*})\cdot\eta\cdot\alpha$ is roughly $O(\sqrt{s^* \cdot \log d \cdot \log n/n})$ and also by the assumption that $\frac{s^* \log d}{n} \cdot (\log n)^{\frac{5}{2}} =o(1)$, we can claim that $(\sqrt{\hat{s}}+c_1/\sqrt{1-L}\cdot\sqrt{s^*})\cdot\eta\cdot\alpha$ is actually $o(1)$, thus the condition $(\sqrt{\hat{s}}+c_1/\sqrt{1-L}\cdot\hat{s^*})\cdot\eta\cdot\alpha \le \min((1-\sqrt{\kappa})^2\cdot R,  \frac{(1-L)^2}{2\cdot(1+L)}\cdot\|\bm \beta^*\|_2)$ can be satisfied. Therefore, we can find a constant $C$, such that:
\begin{equation}
   \frac{ (\sqrt{\hat{s}}+c_1/\sqrt{1-L}\cdot\sqrt{s^*})\cdot\eta}{1-\sqrt{\kappa}}\cdot\alpha \le C \cdot \sqrt{\frac{s^* \cdot \log d \cdot \log n}{n}}.
\end{equation}
Then, to finish the proof, it suffices to show that for each $\bm \beta^t$, $t = 0,1,2,...N_0-1$, we can claim $||\nabla Q_{n/N_0} (\bm \beta^t;\bm \beta^t) - f_T(\nabla Q_{n/N_0} (\bm \beta^t;\bm \beta^t))||_2$ is $O(\sqrt{\frac{s^* \cdot \log d \cdot \log n}{n}})$. Then we can follow the proof of Theorem~\ref{thm:hdem} to finish the proof of Theorem~\ref{thm:hdgmm}.

By the third proposition in Lemma~\ref{lem:hdgmm}, we have already observed that for any $t = 0,1,2,\cdots,N_0-1$, if we choose $\xi = O(\sqrt{\frac{s^* \cdot \log d \cdot \log n}{n}})$. We can find that for a constant $C'$:
\[P(||\nabla Q_{n/N_0} (\bm \beta^t;\bm \beta^t) - f_T(\nabla Q_{n/N_0} (\bm \beta^t;\bm \beta^t))||_2 >\xi) \le C' \cdot \frac{1}{\log d \cdot \log n}.\]
Furthermore, for all $t = 0,1,...N_0-1$, since $N_0$ is chosen to be $O(\log n)$, 
we could apply an union bound, and claim that for all $t$, with probability $1-C'\log d$, $||\nabla Q_{n/N_0} (\bm \beta^t;\bm \beta^t) - f_T(\nabla Q_{n/N_0} (\bm \beta^t;\bm \beta^t))||_2 = O(\sqrt{\frac{s^* \cdot \log d \cdot \log n }{n}})$, which finishes the proof of Theorem~\ref{thm:hdgmm}. \hfill$\square$

\subsection{Proof of Proposition~\ref{thm:hdgmmlow}}\label{pthm:hdgmmlow}
Suppose $\bm Y = \{\bm y_1, \bm y_2,...\bm y_n\}$ be the data set of $n$ samples observed from the Gaussian Mixture Model and let $M$ be any corresponding $(\epsilon, \delta)$-differentially private algorithm. Suppose we have another model where there are no hidden variables $Z$ where $\bm Y = \beta^* + \bm e$. Suppose $\bm Y_1 = \{\bm y_1', \bm y_2',...\bm y_n'\}$ be the data set of $n$ samples observed from the latter model and let $M_1$ be any corresponding $(\epsilon, \delta)$-differentially private algorithm. Then the estimation of true $\bm \beta^*$ can be seen as a mean-estimation problem. By Lemma 3.2 and Theorem 3.3 from \citep{cai2019cost}, we have that:
\[\inf_{M_1 \in \mathcal{M}_{\epsilon,\delta}} \sup_{\bm \beta^* \in \mathbb{R}^d, \|\bm \beta^*\|_0 \le s^*} \E\|{M_1(\bm Y_1)}-\bm \beta^*\|_2 \ge c \cdot (\sqrt{\frac{s^* \log d}{n}} + \frac{s^* \log d}{n \epsilon} )\]
Since the no hidden variables model can be seen as a special case where all hidden variables $Z$ equals 1, so we have:
\[\inf_{M_0 \in \mathcal{M}_{\epsilon,\delta}} \sup_{\bm \beta^* \in \mathbb{R}^d, \|\bm \beta^*\|_0 \le s^*} \E\|{M(\bm Y)}-\bm \beta^*\|_2 \ge \inf_{M_1 \in \mathcal{M}_{\epsilon,\delta}} \sup_{\bm \beta^* \in \mathbb{R}^d, \|\bm \beta^*\|_0 \le s^*} \E\|{M_1(\bm Y_1)}-\bm \beta^*\|_2 \]
By combining the two inequalities together, we finish the proof. \hfill$\square$

\subsection{Proof of Theorem~\ref{thm:hdmrm}}\label{pthm:hdmrm}
Similar to the proof for Theorem~\ref{thm:hdgmm}, the proof consists of three parts. The first part is the privacy guarantees. The second part is to verify the conditions of $\alpha$. The third part is to find the convergence of $\max_{t}||\nabla Q_{n/N_0} (\bm \beta^t;\bm \beta^t) - f_T(\nabla Q_{n/N_0} (\bm \beta^t;\bm \beta^t))||_2$. Here, the verification of $\alpha$ is similar to the proof of Theorem~\ref{thm:hdgmm}, so we omit the proof here.

For the privacy guarantee, notice that for two adjacent data sets, let $(\bm x_i, y_i)$ and $(\bm x_i', y_i')$ be the difference between these two data sets. 
\begin{align}
    &||f_T(\nabla Q_{n/N_0} (\bm \beta;\bm \beta)) -f_T(\nabla {Q_{n/N_0}'} (\bm \beta;\bm \beta))||_{\infty} \notag\\
    &= ||\frac{N_0}{n}\cdot[2 w_{\bm \beta}(\bm x_i, y_i)\cdot \Pi_{T} (y_i) \cdot \Pi_{T} (\bm x_i) - \Pi_{T}(\bm x_i)\cdot \Pi_{T}(\bm x_i^\top \bm \beta)] \notag\\
    &- \frac{N_0}{n}\cdot[2 w_{\bm \beta}(\bm x_i', y_i')\cdot \Pi_{T} (y_i') \cdot \Pi_{T} (\bm x_i') - \Pi_{T}(\bm x_i')\cdot \Pi_{T}(\bm {x_i'}^\top \bm \beta)] ||_{\infty} \notag\\
    &\le \frac{N_0}{n}\| \Pi_{T} (y_i) \cdot \Pi_{T} (\bm x_i) -  \Pi_{T} (y_i') \cdot \Pi_{T} (\bm x_i')\|_{\infty} + \frac{N_0}{n}\|\Pi_{T}(\bm x_i)\cdot \Pi_{T}(\bm x_i^\top \bm \beta)-\Pi_{T}(\bm x_i')\cdot \Pi_{T}(\bm {x_i'}^\top \bm \beta)\|_{\infty} \notag\\
    &\le \frac{4T^2\cdot N_0}{n}.
\end{align}
Next, let us find the convergence rate for the truncation error. By Lemma~\ref{lem:hdmrm}, for the truncation error, we have obtained that with probability 
$1-  m_0/(\log n \cdot \log d)$, therefore, for any $t = 0,1,2,...,N_0-1$, we have: 
\begin{equation}
    \Pro (||\nabla Q_{n_0} (\bm \beta^t;\bm \beta^t) - f_T(\nabla Q_{n_0} (\bm \beta^t;\bm \beta^t))||_2 > C\cdot\sqrt{\frac{s^* \log d}{n}}\cdot \log n) \le C' \cdot 1/{\log d \cdot \log n}, 
\end{equation}
By a union bound on $t$, and if we take the times of iteration $N_0 \asymp \log n$, we have: 
\begin{equation}
    \Pro (\max_{t}||\nabla Q_{n_0} (\bm \beta^t;\bm \beta^t) - f_T(\nabla Q_{n_0} (\bm \beta^t;\bm \beta^t))||_2 > C\cdot\sqrt{\frac{s^* \log d}{n}}\cdot \log n) \le C' \cdot 1/{\log d}.
\end{equation}
Thus, we can get the result that for all $t$, with probability $1-C'/{\log d}$, $||\nabla Q_{n/N_0} (\bm\beta^t;\bm\beta^t) - f_T(\nabla Q_{n/N_0} (\bm\beta^t;\bm\beta^t))||_2 = O(\sqrt{\frac{s^* \log d}{n}}\cdot \log n) $, thus follows the proof of Theorem~\ref{thm:hdem}, we finish the proof of Theorem~\ref{thm:hdmrm}. \hfill$\square$
\subsection{Proof of Proposition~\ref{thm:hdmrmlow}}\label{pthm:hdmrmlow}
Suppose we have the traditional linear regression model where there are no hidden variables $Z$ where $Y = \bm X^\top \beta^* + \bm e$. Suppose $ (Y_1, \bm X_1) =  \{( y_1', \bm x_1'), ( y_2',\bm x_2'),...( y_n', \bm x_n')\}$ be the data set of $n$ samples observed from the linear regression model and let $M_1$ be any corresponding $(\epsilon, \delta)$-differentially private algorithm. Then by Lemma 4.3 and Theorem 4.3 in \citep{cai2019cost}, we have that:
\[\inf_{M_1 \in \mathcal{M}_{\epsilon,\delta}} \sup_{\bm \beta^* \in \mathbb{R}^d, \|\bm \beta^*\|_0 \le s^*} \E\|{M_1( Y_1, \bm X_1)}-\bm \beta^*\|_2 \ge c \cdot (\sqrt{\frac{s^* \log d}{n}} + \frac{s^* \log d}{n \epsilon} ).\]
Since the no hidden variables model can be seen as a special case where all hidden variables $Z$ equals 1, so we have:
\[\inf_{M \in \mathcal{M}_{\epsilon,\delta}} \sup_{\bm \beta^* \in \mathbb{R}^d, \|\bm \beta^*\|_0 \le s^*} \E\|{M( Y, \bm X)}-\bm \beta^*\|_2 \ge \inf_{M_1 \in \mathcal{M}_{\epsilon,\delta}} \sup_{\bm \beta^* \in \mathbb{R}^d, \|\bm \beta^*\|_0 \le s^*} \E\|{M_1(Y_1, \bm X_1)}-\bm \beta^*\|_2. \]
By combining the two inequalities together, we finish the proof. \hfill$\square$

\subsection{Proof of Theorem~\ref{thm:hdmcm}}\label{pthm:hdmcm}
In the proof of Theorem~\ref{thm:hdmcm}, we also need to verify two properties. First is the privacy guarantees. The second part is to find the convergence of $\max_{t}||\nabla Q_{n/N_0} (\bm \beta^t;\bm \beta^t) - f_T(\nabla Q_{n/N_0} (\bm \beta^t;\bm \beta^t))||_2$.

For the privacy guarantee, notice that for two adjacent data sets, let $(\bm x_i, y_i)$ and $(\bm x_i', y_i')$ be the difference between these two data sets. For the ease of notation, we denote $n_{\bm \beta}(\tilde{\bm x}_i, y_i, z_i)) = (1-\bm z_i) \odot m_{\bm \beta}(\tilde{\bm x}_i, y_i))$ and $u_i = (1-\bm z_i) \odot m_{\bm \beta}(\tilde{\bm x}_i, y_i)^\top\cdot \bm \beta $. Then:
\begin{align}
    &||f_T(\nabla Q_{n/N_0} (\bm \beta;\bm \beta)) -f_T(\nabla {Q_{n/N_0}'} (\bm \beta;\bm \beta))||_{\infty} \notag\\
    &\le \frac{N_0}{n} ||\Pi_{T} (y_i) \cdot \Pi_{T}(m_{\beta}(\tilde{\bm x}_i, y_i)) - \Pi_{T} (y_i') \cdot \Pi_{T}(m_{\beta}(\tilde{\bm x}_i', y_i'))||_{\infty} \notag\\
    &+\frac{N_0}{n}|| \Pi_{T}(m_{\bm \beta}(\tilde{\bm x}_i, y_i)) \cdot  \Pi_{T}(m_{\bm \beta}(\tilde{\bm x}_i, y_i)^\top\cdot\bm \beta) -\Pi_{T}(m_{\bm \beta}(\tilde{\bm x}_i', y_i')) \cdot  \Pi_{T}(m_{\bm \beta}(\tilde{\bm x}_i', y_i')^\top\cdot\bm \beta)||_{\infty} \notag\\
    &+ \frac{N_0}{n}|| \Pi_{T}(n_{\bm \beta}(\tilde{\bm x}_i, y_i, z_i))\cdot  \Pi_{T}(u_i) -\Pi_{T}(n_{\bm \beta}(\tilde{\bm x}_i', y_i',z_i')) \cdot  \Pi_{T}(u_i')||_{\infty} \notag \\
    &\le \frac{6T^2\cdot N_0}{n}
\end{align}
From Lemma~\ref{lem:hdmcm}, we have that under the truncation condition, with probability 
$1-  m_0/(\log n \cdot \log d)$, for any $t = 0,1,2,...,N_0-1$, we have: 
\begin{equation}
    P (||\nabla Q_{n/N_0} (\bm \beta^t;\bm \beta^t) - f_T(\nabla Q_{n/N_0} (\bm \beta^t;\bm \beta^t))||_2 > C\cdot\sqrt{\frac{s^* \log d}{n}}\cdot \log n) \le C' \cdot 1/{\log d \cdot \log n}, 
\end{equation}
Again, by a union bound on $t$, and if we take the times of iteration $N_0 \asymp \log n$, we have: 
\begin{equation}
    P (\max_{t}||\nabla Q_{n/N_0} (\bm \beta^t;\bm \beta^t) - f_T(\nabla Q_{n/N_0} (\bm \beta^t;\bm \beta^t))||_2 > C\cdot\sqrt{\frac{s^* \log d}{n}}\cdot \log n) \le C' \cdot 1/{\log d}
\end{equation}
Thus, we can get the result that for all $t$, with probability $1-C'/{\log d}$, $\max_{t}||\nabla Q_{n/N_0} (\bm \beta^t;\bm \beta^t) - f_T(\nabla Q_{n/N_0} (\bm \beta^t;\bm \beta^t))||_2 =O(\sqrt{\frac{s^* \log d}{n}}\cdot \log n)$, thus by Theorem~\ref{thm:hdem}, we finish the proof of Theorem~\ref{thm:hdmcm}. \hfill$\square$
\subsection{Proof of Theorem~\ref{thm:ldem}}\label{pthm:ldem}
Before the proof, let us first introduce a lemma: 
\begin{lemma}\label{lem:pldem1}
If $0 \leq \gamma < \nu \leq \mu$ holds,
suppose that Condition~\ref{con: hdem1}, \ref{con:hdem2} holds with parameter $\gamma, (\mu, \nu)$, respectively. Then if step size $\eta= \frac{2}{\mu +\nu}$, we have: 
\begin{align*}
||\bm \beta +\eta \cdot \nabla{Q(\bm \beta;\bm \beta)} - \bm \beta^*||_2 \le (1-\frac{2\nu -2\gamma}{\mu+\nu})||\bm \beta - \bm\beta^*||_2.
\end{align*}
\end{lemma}
The detailed proof for \ref{lem:pldem1} is in Theorem 3 from \citep{balakrishnan2017statistical}. Then for $t = 0,1,2,3,...N_0-1$, we have:
\begin{align}\label{peq25}
    ||\bm \beta^{t+1}- \bm \beta^*||_2
    &= || \bm \beta^{t} + \eta f_T(\nabla Q_{n/N_0}(\bm \beta^t;\bm \beta^t)) + \bm W_t - \bm \beta^*||_2 \notag\\
    &\le || \bm \beta^{t} + \eta \nabla Q(\bm \beta^t;\bm \beta^t) - \bm \beta^* ||_2 + \eta ||f_T(\nabla Q_{n/N_0}(\bm \beta^t;\bm \beta^t))-\nabla Q(\bm \beta^t;\bm \beta^t)||_2+||\bm W_t||_2 \notag\\
    &\le \kappa||\bm \beta^t-\bm \beta^*||_2 + ||\bm W_t||_2 +\eta ||f_T(\nabla Q_{n/N_0}(\bm \beta^t;\bm \beta^t))- \nabla Q_{n/N_0}(\bm \beta^t;\bm \beta^t)||_2\notag\\&+\eta ||\nabla Q_{n/N_0}(\bm \beta^t;\bm \beta^t)-\nabla Q(\bm \beta^t;\bm \beta^t)||_2 \notag\\
    &\le \kappa||\bm \beta^t-\bm \beta^*||_2 + ||\bm W_t||_2 +\eta ||f_T(\nabla Q_{n/N_0}(\bm \beta^t;\bm \beta^t))- \nabla Q_{n/N_0}(\bm \beta^t;\bm \beta^t)||_2+\eta\cdot\alpha.
\end{align}
In the above equality, if by choosing $T \asymp \sqrt{\log n}$, from the assumption, with probability $1-N_0\cdot\phi(\xi)$, for any $t$, $||f_T(\nabla Q_{n/N_0}(\bm \beta^t;\bm \beta^t))- \nabla Q_{n/N_0}(\bm \beta^t;\bm \beta^t)||_2 \le \xi$. Also, by Condition~\ref{con:ldem}, for a single $t$, with probability $1-\tau/N_0$, $||\nabla Q_{n/N_0}(\beta^t;\beta^t)-\nabla Q(\beta^t;\beta^t)||_2 \le \alpha$, then by a union bound, for all $t = 0,1,2,...{N_0}-1$, we have with probability $1-\tau$, $\max_{t}||\nabla Q_{n/N_0}(\beta^t;\beta^t)-\nabla Q(\beta^t;\beta^t)||_2 \le \alpha$. Then, if $\beta^t \in \mathcal{B}$, which means that $||\beta^t -\beta^*||_2 \le R$. Then if $\alpha$ satisfies that $\alpha \le \frac{\nu-\gamma}{4}\cdot R$, then we can have $\eta \cdot (\alpha+\xi) \le \frac{1-\kappa}{2}\cdot R$.\\\\
On the other hand, and if we choose $T \asymp \sqrt{\log n }$ and the times of rotation $N_0$ to be $N_0 \asymp \log n$. Notice that for each $W_{ti}, i = 1,2,3,...d$, $W_{ti} \sim N(0, \sigma^2)$. $\sigma^2 = \frac{2 \eta^2 d (2T)^2 {N_0}^2 \log(1.25/\delta)}{n^2 \epsilon^2}$. Then we find that actually $||\bm W_t||_2^2/\sigma^2$ follows a chi-square distribution $\chi^2(d)$. By the concentration of chi-square distribution, there exists constants $c_0, c_1, c_2$, such that:
\begin{align}
    \Pro(||\bm W_t||_2^2 \ge \sigma^2(1+c_1)d) \le c_0 \exp(-\min\{c_1^2 d, c_1 d\}/8) \le c_0\exp(-c_2 d).
\end{align}
Then with probability $1-c_0\exp(-c_2 d)$, we can derive that, their exists a constant $c_3$ such that:
\[||\bm W_t||_2 \le c_3\cdot \frac{d\sqrt{\log^3 n} \sqrt{\log(1/\delta)}}{n\epsilon}.\]
By a union bound, we find that with probability $1-c_0 \cdot \log n \cdot \exp(-c_2 d)$, we can derive that, their exists a constant $c_4$ such that:
\[\max_{t}||\bm W_t||_2 \le c_4\cdot \frac{d\sqrt{\log^3 n} \sqrt{\log(1/\delta)}}{n\epsilon}.\]
Then, we properly choose $n$ such that $\max_{t}||\bm W_t||_2 \le \frac{1-\kappa}{2}\cdot R$. Thus, when $||\bm \beta^t -\bm \beta^*||_2 \le R$, we can also guarantee that $||\bm \beta^{t+1} - \bm \beta^*||_2 \le R$, thus we can iterate the conclusions in (\ref{peq25}), and we can get: 
\begin{align*}
    ||\bm \beta^{t}-\bm \beta^*||_2 &\le \kappa^t ||\bm \beta^0 -\bm \beta^*||_2 + \sum_{i=0}^{t-1} \kappa^{t-1-i}||\bm W_i||_2 + \frac{\eta}{1-\kappa} [ \xi+\alpha] \\&\le \frac{\kappa^t}{2} R + C\cdot \frac{d\sqrt{\log^3 n} \sqrt{\log(1/\delta)}}{n\epsilon} + \frac{\eta}{1-\kappa} [ \xi+\alpha].
\end{align*}
$\hfill\square$

\subsection{Proof of Theorem~\ref{thm:ldgmm} and Proposition~\ref{thm:ldgmmlow}}\label{pthm:ldgmm}
For the proof of Theorem~\ref{thm:ldgmm}, we are more focusing on two parts. The first part is the convergence of $\alpha$, which has already been shown in Corollary 9 from \citep{balakrishnan2017statistical}, which shows that when $T \asymp \sqrt{\log n}$, $\alpha = O(\sqrt{\frac{d}{n}}\cdot \log n)$. So we will later show the convergence rate of the truncation error. 

Follow the proof in Lemma~\ref{lem:hdgmm}, we could observe that for the vector $\nabla Q_{n/N_0} (\bm \beta^t;\bm \beta^t) - f_T(\nabla Q_{n/N_0} (\bm \beta^t;\bm \beta^t))$, $\E ||\nabla Q_{n/N_0} (\bm \beta^t;\bm \beta^t) - f_T(\nabla Q_{n/N_0} (\bm \beta^t;\bm \beta^t))||_2^2 = O(\frac{d}{n})$ in the low dimensional settings. Thus if we choose $\xi = O(\sqrt{\frac{d \cdot \log n}{n}})$. We can find that for a constant $C'$:
\[\Pro(||\nabla Q_{n/N_0} (\bm \beta^t;\bm \beta^t) - f_T(\nabla Q_{n/N_0} (\bm \beta^t;\bm \beta^t))||_2 >\xi) \le C' \cdot \frac{1}{ \log n}\]

By the definition of $\nabla Q_{n/N_0} (\bm \beta^t;\bm \beta^t)$ in the Gaussian Mixture Model, we found that the truncation error does not rely on the choice of $\bm \beta^t$, thus for all $t$, by probability $1-C'\log n$, $||\nabla Q_{n/N_0} (\bm \beta^t;\bm \beta^t) - f_T(\nabla Q_{n/N_0} (\bm \beta^t;\bm \beta^t))||_2 = O(\sqrt{\frac{d \cdot \log n }{n}})$, which finishes the proof of Theorem~\ref{thm:ldgmm}.

Then, for the proof of Proposition~\ref{thm:ldgmmlow}, we could follow the same proof of Proposition~\ref{thm:hdgmmlow}. But in the proof, rather than using  Lemma 3.2 and Theorem 3.3 from \citep{cai2019cost}, we should use Lemma 3.1 and Theorem 3.1 from \citep{cai2019cost}. 
$\hfill\square$

 
\section{Proof of lemmas}\label{sec:appendix} 
\subsection{Proof of Lemma~\ref{lem:noisyht2}}\label{plem:noisyht2}
Let $\psi: R_2 \to R_1$ be a bijection. By the selection criterion of Algorithm \ref{algo:peeling}, for each $j \in R_2$ we have $|v_j| +  w_{ij} \leq |v_{\psi(j)}| +  w_{i\psi(j)}$, where $i$ is the index of the iteration in which $\psi(j)$ is appended to $S$. It follows that, for every $c > 0$, 
	\begin{align*}
		 v_j^2 &\leq \left(| v_{\psi(j)}| + w_{i\psi(j)} - w_{ij} \right)^2 \\
		&\leq (1+c)  v_{\psi(j)}^2 + (1 + 1/c)(w_{i\psi(j)} - w_{ij})^2 \leq (1+c)v_{\psi(j)}^2 + 4(1+1/c)\| \bm w_i\|_\infty^2
	\end{align*}
	Summing over $j$ then leads to
	\begin{align*}
		\|\bm v_{R_2}\|_2^2 \leq (1 + c)\|\bm v_{R_1}\|_2^2 + 4(1 + 1/c)\sum_{i \in [s]} \|\bm w_i\|^2_\infty,
	\end{align*}
which finishes the proof of Lemma~\ref{lem:noisyht2}. \hfill$\square$
\subsection{Proof of Lemma~\ref{lem:noisyht3}}\label{plem:noisyht3}
To prove this result, we first prove that for each iteration, we can gain a $(\epsilon,\delta)$-differentially private algorithm. Suppose the data points we have gathered is $y_1,y_2,...y_n$, and if one individual of the data point, say, $y_n$ is replaced by $\Tilde{y}_n$, then define $\nabla Q_{y_n}(\cdot ;\cdot )$ be the gradient taken with respect to $y_n$, we can show that: 
\begin{align}
{||f_T(\nabla Q_{n/N_0}(\bm \beta^t;\bm \beta^t)) - f_T(\nabla {Q_{n/N_0}'(\bm \beta^t;\bm \beta^t))||}}_{\infty} &= \frac{N_0}{n} ||h_T(\nabla Q_{y_n}(\bm \beta^t; \bm \beta^t)) - h_T(\nabla Q_{\Tilde{y}_n}(\bm \beta^t; \bm \beta^t))||_{\infty} \notag \\&\le \frac{2T\cdot N_0}{n}, 
\end{align}
where the last inequality follows from the definition of $h_T$. Then by Lemma~\ref{lem:noisyht1} we can get the result that for each iteration, we can obtain a $(\epsilon,\delta)$-differentially private algorithm. Next, we will show that for an iterative algorithm with data-splitting, the whole algorithm is also a $(\epsilon,\delta)$-differentially private algorithm. Let us start from the simple case, a two step iterative algorithm. 

Let $D$ denote the data set and $D'$ be the adjacent data set of $D$. Assume the data set are split into two data sets where $D = D_1 \cup D_2$ and $D_1 \cap D_2 = \varnothing$. Then let $M_1(D_1)$ be a $(\epsilon,\delta)$-differentially private algorithm with output $v$. $M_2(v, D_2)$ also be a $(\epsilon,\delta)$-differentially private algorithm with any given $v$. Then we define $M(D) = M_2(M_1(D_1), D_2)$, we claim that $M(D)$ is also $(\epsilon,\delta)$-differentially private. 
To prove this claim, we should use the definition of differential privacy. Since $D$ and $D'$ are two adjacent data sets and differ by only one individual data. Thus, $D' = D_1 \cup D_2'$ or $D' = D_1' \cup D_2$. We will discuss these two cases one by one.
For the first case, $D' = D_1' \cup D_2$. from the definition, we have:
\begin{align*}
    \mathbb{P}(M(D') \in S) &= \mathbb{P}(M_2(M_1(D_1'), D_2) \in S)\\
    &= \mathbb{E}_{M_2} \Pro(M_2(M_1(D_1'), D_2) \in S | M_2)\\
    &= \E_{M_2} \Pro(M_1(D_1')\in S(M_2) | M_2)\\
    &\le \E_{M_2} e^{\epsilon} \Pro(M_1(D_1)\in S(M_2) | M_2) +\delta\\
    &\le e^{\epsilon} \Pro(M_2(M_1(D_1), D_2) \in S) +\delta\\
    &= e^{\epsilon}\Pro(M(D) \in S)+\delta
\end{align*}
From the definition, we could claim that $M$ is $(\epsilon,\delta)$-differentially private. Then, for the second case, we have:
\begin{align*}
    \Pro(M(D') \in S) &= \Pro(M_2(M_1(D_1), D_2') \in S)\\
    &= \E_{M_1} P(M_2(M_1(D_1), D_2') \in S | M_1)\\
    &\le \E_{M_1} e^{\epsilon}\Pro(M_2(M_1(D_1), D_2) \in S | M_1) + \delta\\
    &= e^{\epsilon}\Pro(M_2(M_1(D_1), D_2) \in S) + \delta\\
    &= e^{\epsilon}\Pro(M(D) \in S) +\delta
\end{align*}
For the second case, we also claim that $M$ is $(\epsilon,\delta)$-differentially private. Then, when the data set is split into $k = 2,3,4,...$ subsets, we could use the induction to show that for an iterative algorithm with data-splitting and each iteration being $(\epsilon,\delta)$-differentially private, the combined algorithm is also a $(\epsilon,\delta)$-differentially private algorithm, which finish the proof of this lemma.  {\hfill $\square$}
\subsection{Proof of Lemma~\ref{lem:hdgmm}}\label{plem:hdgmm}
The proof consists of three parts, for the first proposition of verification of Condition~\ref{con: hdem1} and Condition~\ref{con:hdem2}, the proof is slight different with the proof for Corollary 1 in \citep{balakrishnan2017statistical}. The difference is that for the Lipschitz-Gradient condition, instead the Lipschitz-Gradient-1$(\gamma_1,\mathcal{B})$ condition in \citep{balakrishnan2017statistical}, we are using Lipschitz-Gradient$(\gamma,\mathcal{B})$. However, because the population level $\nabla Q(\cdot ;\cdot)$ satisfies:
\[\nabla Q(\bm \beta' ;\bm \beta) = 2 \E[w_{\bm \beta}(Y)\cdot Y] - \bm \beta'.\]
So obviously, 
\[\nabla Q(\bm \beta ;\bm \beta^*) - \nabla Q(\bm \beta ;\bm \beta) =  2 \E[(w_{\bm \beta^*}-w_{\bm \beta})(Y)\cdot Y] . \]
Also, let $M(\bm \beta) = \text{argmax}_{\bm \beta'} Q(\bm \beta';\bm \beta)$, then we find that,
\[\nabla Q(M(\bm \beta) ;\bm \beta^*) - \nabla Q(M(\bm \beta) ;\bm \beta) =  2 \E[(w_{\bm \beta^*}-w_{\bm \beta})(Y)\cdot Y] \]
so in the case of Gaussian Mixture Model, $\gamma_1 = \gamma$, follow the proof of Corollary 1 in \citep{balakrishnan2017statistical}, the Lipschitz-Gradient$(\gamma,\mathcal{B})$ holds when taking $\gamma = \exp(-L \cdot \phi^2)$.
For the second proposition of Condition~\ref{con:hdem3}, see detailed proof of Lemma 3.6 in \citep{wang2015high}.
For the third proposition of Condition~\ref{con:hdem4}, by the proof in (\ref{peq23}) for Theorem~\ref{thm:hdem}, we find that we only need to bound $||\nabla Q_{n/N_0} (\bm \beta^t;\bm \beta^t) - f_T(\nabla Q_{n/N_0} (\bm \beta^t;\bm \beta^t))||_2$ with the support on $\mathcal{S}^{t+0.5}$. Where $\mathcal{S}^{t+0.5}$ is the set of indexes chosen by the NoisyHT algorithm during the $t$-th iteration. Thus, for any $\xi > 0$, since:
\begin{align*}
 ||\nabla Q_{n/N_0} (\bm \beta^t;\bm \beta^t) - f_T(\nabla Q_{n/N_0} (\bm \beta^t;\bm \beta^t))||_2 &\le \frac{1}{n}||\sum_{i=1}^n [2 w_{\bm \beta^t}(y_i)-1](\Pi_{T}(\bm y_i)-\bm y_i)||_2\\
 &\le  \frac{1}{n}\sum_{i=1}^n|| \Pi_{T}(\bm y_i)- \bm y_i||_2
\end{align*}
Thus, we have:
\begin{align}
     \Pro(||(\nabla Q_{n/N_0} (\beta^t;\beta^t) - f_T(\nabla Q_{n/N_0} (\beta^t;\beta^t)))_{\mathcal{S}^{t+0.5}}|| >\xi) 
    &\le \Pro(\frac{N_0}{n}\sum_{i=1}^{n/N_0}||( \Pi_{T}(\bm y_i)-\bm y_i)_{\mathcal{S}^{t+0.5}}||_2 > \xi)\notag\\
    &\le \frac{\E[\frac{N_0}{n}\sum_{i=1}^{n/N_0}||( \Pi_{T}(\bm y_i)-\bm y_i)_{\mathcal{S}^{t+0.5}}||_2]^2}{\xi^2}\notag\\
    &\le \frac{\frac{N_0}{n}\sum_{i=1}^{n/N_0} \E ||( \Pi_{T}(\bm y_i)-\bm y_i)_{\mathcal{S}^{t+0.5}}||_2^2}{\xi^2}\notag\\
    &\le \frac{ \E ||( \Pi_{T}(\bm Y)-\bm Y)_{\mathcal{S}^{t+0.5}}||_2^2}{\xi^2},
\end{align}
where $\bm Y$ is the response of the Gaussian Mixture model on the population level. Then for any $j = 1,2,...d \in \mathcal{S}^{t+0.5}$, we can obtain that:
\begin{equation}
    \E ||( \Pi_{T}(\bm Y)-\bm Y)_{\mathcal{S}^{t+0.5}}||_2^2 = \sum_{j=1}^{d} \E(\Pi_{T}(Y_j)-Y_j)^2 \cdot \textbf{1}_{j \in \mathcal{S}^{t+0.5}}.
\end{equation}
Then since for any $j \in \mathcal{S}^{t+0.5}$, $\Pro(Y_j \sim N(\beta_j,\sigma^2)) =\frac{1}{2}$ and $\Pro(Y_j \sim N(-\beta_j,\sigma^2)) =\frac{1}{2}$. Denote the density function of $N(\beta_j,\sigma^2)$ as $f_{j1}$, the density function of $N(-\beta_j,\sigma^2)$ as $f_{j2}$, the density function of $N(0,\sigma^2)$ as $f_{z}$, we can get:
\begin{align}\label{peq26}
    &\E(\Pi_{T}(Y_j)-Y_j)^2 \notag\\ &=\frac{1}{2}\E[(\Pi_{T}(Y_j)-Y_j)^2 |Y_j \sim N(\beta_j,\sigma^2) ] + \frac{1}{2}\E[(\Pi_{T}(Y_j)-Y_j)^2 |Y_j \sim N(-\beta_j,\sigma^2) ] \notag\\
    &= \frac{1}{2} [\int_{T}^{\infty}(y-T)^2f_{j1} dy + \int_{-\infty}^{-T}(y+T)^2f_{j1} dy+ \int_{T}^{\infty}(y-T)^2f_{j2} dy+ \int_{-\infty}^{-T}(y+T)^2f_{j2} dy ]\notag\\
    &= \frac{1}{2} [\int_{T+\beta_j}^{\infty}(z-\beta_j-T)^2f_{z} dz + \int_{-\infty}^{-T+\beta_j}(z-\beta_j+T)^2f_{z} dz+ \int_{T-\beta_j}^{\infty}(z+\beta_j-T)^2f_{z} dz \notag\\
    &+ \int_{-\infty}^{-T-\beta_j}(z+\beta_j+T)^2f_{z} dz ]\notag\\
    &= \int_{T+\beta_j}^{\infty}(z-\beta_j-T)^2f_{z} dz + \int_{T-\beta_j}^{\infty}(z+\beta_j-T)^2f_{z} dz.
\end{align}
For the first term in the above result (\ref{peq26}), by Fubini theorem:
\begin{align}\label{peq27}
    \int_{T+\beta_j}^{\infty}(z-\beta_j-T)^2f_{z} dz 
    &= \int_{T+\beta_j}^{\infty}\int_{0}^{z-\beta_j-T} \frac{1}{2} t dt f_{z} dz \notag\\
    &=  \frac{1}{2}\int_{0}^{\infty} \int_{T+\beta_j +t}^{\infty} f_z dz\cdot t dt\notag\\
    &=  \frac{1}{2}\int_{0}^{\infty} \Pro(Z \ge T+\beta_j+t) t dt.
\end{align}
By the tail bound of Gaussian distributions, we can have:
\begin{equation}\label{peq28}
\Pro(Z \ge T+\beta_j+t) \le  \exp(-\frac{(T+\beta_j+t)^2}{2\sigma^2})    
\end{equation}
Insert the tail bound from (\ref{peq28}) into (\ref{peq27}), we can obtain:
\begin{align}
    \int_{T+\beta_j}^{\infty}(z-\beta_j-T)^2f_{z} dz 
    &\le \frac{1}{2} \int_{T+\beta_j}^{\infty} \exp(-\frac{t^2}{2\sigma^2})(t-T-\beta_j) dt \notag\\
    &\le  \int_{T+\beta_j}^{\infty} \exp(-\frac{t^2}{2\sigma^2})dt^2 \notag\\
    &\le 2\sigma^2 \cdot \exp(-\frac{(T+\beta_j)^2}{2\sigma^2}).
\end{align}
If we choose $T = c \cdot \sigma \sqrt{\log n}$ and analyze the second term in similar approach in (\ref{peq26}), we can find that $\E(\Pi_{T}(Y_j)-Y_j)^2 = O(\frac{1}{n})$. Then, according to (\ref{peq26}), we can claim that for sufficiently large $n$, $\E ||( \Pi_{T}(\bm Y)-\bm Y)_{\mathcal{S}^{t+0.5}}||_2^2 = O(\frac{s^*}{n})$. So if we choose $\xi = O(\sqrt{\frac{s^* \cdot \log d \cdot \log n}{n}})$. We can find that for a constant $C'$:
\[\Pro(||\nabla Q_{n/N_0} (\bm \beta^t;\bm \beta^t) - f_T(\nabla Q_{n/N_0} (\bm \beta^t;\bm \beta^t))||_2 >\xi) \le C' \cdot \frac{1}{\log d \cdot \log n},\]
which finished the proof of the third proposition in Lemma~\ref{lem:hdgmm}. Thus, we complete the proof of Lemma~\ref{lem:hdgmm}.\hfill$\square$
\subsection{Proof of Lemma~\ref{lem:hdmrm}}\label{plem:hdmrm}
For the first proposition in Lemma~\ref{lem:hdmrm}, see detailed proof in Corollary 3 from \citep{balakrishnan2017statistical}, for the second proposition in Lemma~\ref{lem:hdmrm}, see detailed proof in Lemma 3.9 from \citep{wang2015high}. In this section, our major focus is the third proposition, which verifies Condition~\ref{con:hdem4} for the Truncation-error. Before we start the proof of third proposition, let us first introduce two lemmas, which are significant in the following analysis.
\begin{lemma}\label{lem:phdmrm1}
Let $X$ be a sub-gaussian random variable on $\mathbb{R}$ with mean zero and variance $\sigma^2$. Then, for the choice of $T$, if $T = O(\sigma \sqrt{\log n})$, we have $\E(\Pi_{T}(x)-x)^2 = O(\frac{1}{n})$. Further, we also have  $\E(\Pi_{T}(x)-x)^4 = O(\frac{\log n}{n})$.
\end{lemma}
\begin{lemma}\label{lem:phdmrm2}
Let $\bm X \in \mathbb{R}^d$ be a sub-gaussian random vector and $Y \in \mathbb{R}$ be a sub-gaussian random variable. Also, let $x_1,x_2,...x_{n_0}$ be $n_0$ realizations of $\bm X$ and $y_1,y_2,...y_{n_0}$ be $n_0$ realizations of $Y$. Suppose we have an index set $\mathcal{S}$ with $|\mathcal{S}| = s$. Let $T \asymp \sqrt{\log n}$, then it holds that, there exists constants $C, m_0$:
\begin{equation}\label{peq34}
\frac{1}{n_0}\sum_{i=1}^{n_0}||(y_i \cdot \bm x_i - \Pi_{T}(y_i)\Pi_{T}(\bm x_i))_{\mathcal{S}}||_2 \le C \cdot \sqrt{\frac{s\log d}{n}}\cdot \log n   
\end{equation}
with probability greater than $1-m_0/(\log n \cdot \log d)$.
\end{lemma}
The proof of Lemma~\ref{lem:phdmrm1} and Lemma~\ref{lem:phdmrm2} can be found in the Appendix~\ref{plem:phdmrm1} and \ref{plem:phdmrm2}. Then we could start the proof of Lemma~\ref{lem:hdmrm}. Similar to Gaussian Mixture Model, we also assume that for any $t$,  $||\nabla Q_{n/N_0} (\bm \beta^t;\bm \beta^t) - f_T(\nabla Q_{n/N_0} (\bm \beta^t;\bm \beta^t))||_2$ has support $\mathcal{S}^{t+0.5}$. Then let $n_0 = n/N_0$ for the ease of notation, we can break $||\nabla Q_{n_0} (\bm \beta^t;\bm \beta^t) - f_T(\nabla Q_{n_0} (\bm \beta^t;\bm \beta^t))||_2$ into two parts:
\begin{align}\label{peq29}
    &||\nabla Q_{n_0} (\bm \beta^t;\bm \beta^t) - f_T(\nabla Q_{n_0} (\bm \beta^t;\bm \beta^t))||_2 \notag\\
    &= ||(\nabla Q_{n_0} (\bm \beta^t;\bm \beta^t) - f_T(\nabla Q_{n_0} (\bm \beta^t;\bm \beta^t)))_{\mathcal{S}^{t+0.5}}||_2 \notag\\
    &\le \underbrace{\frac{1}{n_0}\sum_{i=1}^{n_0} ||(y_i \cdot \bm x_i - \Pi_{T}(y_i)\Pi_{T}(\bm x_i))_{\mathcal{S}^{t+0.5}}||_2}_{(\ref{peq29}.1)}+ \underbrace{\frac{1}{n_0}\sum_{i=1}^{n_0}||(\bm x_i \cdot \bm x_i^\top\bm\beta^t - \Pi_{T}(\bm x_i^\top \bm\beta^t)\Pi_{T}(\bm x_i))_{\mathcal{S}^{t+0.5}}||_2.}_{(\ref{peq29}.2)}
\end{align}
Since $\bm x_i \sim N(\bm 0, \bm I_d)$, $y_i \sim N(0, \beta^\top \beta+\sigma^2)$ and $\bm x_i^\top \beta^t \sim N(0, {\beta^t}^\top \beta^t)$ are all gaussian random variables. So by the conclusion from Lemma~\ref{lem:phdmrm2}, we could conclude that both the term $(\ref{peq29}.1)$ and the term $(\ref{peq29}.2)$ are all $O(\sqrt{\frac{s^*\log d}{n}} \cdot \log n)$ with probability 
$1-  m_0/(\log n \cdot \log d)$, therefore, for any $t = 0,1,2,...,N_0-1$, we have: 
\begin{equation}
    \Pro (||\nabla Q_{n_0} (\bm \beta^t;\bm \beta^t) - f_T(\nabla Q_{n_0} (\bm \beta^t;\bm \beta^t))||_2 > C\cdot\sqrt{\frac{s^* \log d}{n}}\cdot \log n) \le C' \cdot 1/{\log d \cdot \log n}, 
\end{equation}
which finishes the proof of the third proposition in Lemma~\ref{lem:hdmrm}.\hfill$\square$
\subsection{Proof of Lemma~\ref{lem:hdmcm}}\label{plem:hdmcm}
For the first proposition, The detailed proof is in Corollary 6 from \citep{balakrishnan2017statistical}. For the second proposition Lemma~\ref{lem:hdmcm}, the proof is in Lemma 3.12 from \citep{wang2015high}. Then, we will focus on the proof of the third proposition.

Then, similar to the previous approach, we assume that for any $t$,  $||\nabla Q_{n/N_0} (\bm \beta^t;\bm \beta^t) - f_T(\nabla Q_{n/N_0} (\bm \beta^t;\bm \beta^t))||_2$ has support $\mathcal{S}^{t+0.5}$. Denote $n_0 = n/N_0$, then we can break it into three parts. For the ease of notation, we denote $n_{\bm \beta}(\tilde{\bm x}_i, y_i, z_i)) = (1-\bm z_i) \odot m_{\bm \beta}(\tilde{\bm x}_i, y_i))$ and $u_i = (1-\bm z_i) \odot m_{\bm \beta}(\tilde{\bm x}_i, y_i)^\top\cdot \bm \beta $.
\begin{align}
    &||\nabla Q_{n_0} (\bm \beta^t;\bm \beta^t) - f_T(\nabla Q_{n_0} (\bm \beta^t;\bm \beta^t))||_2 \notag\\
    &= ||(\nabla Q_{n_0} (\bm \beta^t;\bm \beta^t) - f_T(\nabla Q_{n_0} (\bm \beta^t;\bm \beta^t)))_{\mathcal{S}^{t+0.5}}||_2 \notag\\
    &\le \frac{1}{n_0}\sum_{i=1}^{n_0} ||(\Pi_{T} (y_i) \cdot \Pi_{T}(m_{\beta}(\tilde{\bm x}_i, y_i)) - y_i\cdot m_{\beta}(\tilde{\bm x}_i, y_i))_{\mathcal{S}^{t+0.5}}||_{2} \label{peq37}\\ 
    &+ \frac{1}{n_0}\sum_{i=1}^{n_0}|| (\Pi_{T}(n_{\beta}(\tilde{\bm x}_i, y_i, z_i))\cdot  \Pi_{T}(u_i) -n_{\beta}(\tilde{\bm x}_i, y_i,z_i) \cdot  u_i)_{\mathcal{S}^{t+0.5}}||_{2} \label{peq38}\\
    &+\frac{1}{n_0}\sum_{i=1}^{n_0}|| (\Pi_{T}(m_{\beta}(\tilde{\bm x}_i, y_i)) \cdot  \Pi_{T}(m_{\beta}(\tilde{\bm x}_i, y_i)^\top\cdot\beta) - m_{\beta}(\tilde{\bm x}_i, y_i) \cdot  (m_{\beta}(\tilde{\bm x}_i, y_i)^\top\cdot\beta))_{\mathcal{S}^{t+0.5}}||_{2} \label{peq39}
\end{align}
Then, to analysis (\ref{peq37}), (\ref{peq38}) and (\ref{peq39}), we should first introduce a lemma below.
\begin{lemma}\label{lem:phdmcm1}
Under the conditions of Theorem~\ref{thm:hdmcm}, the random vector $m_{\beta}(\tilde{\bm x}_i, y_i)$ is sub-gaussian with a constant parameter.
\end{lemma}
The detailed proof of Lemma~\ref{lem:phdmcm1} is in the proof for Lemma 10 in \citep{balakrishnan2017statistical}. Then by the definition of sub-gaussian random vector, we can also claim that $m_{\beta}(\tilde{\bm x}_i, y_i)^\top\cdot\beta$ is sub-gaussian.

Further, for the term $n_{\beta}(\tilde{\bm x}_i, y_i, z_i) = (1-\bm z_i) \odot m_{\beta}(\tilde{\bm x}_i, y_i)$, we also claim that, for any unit vector $v \in \mathbb{R}^d$, $n_{\beta}(\tilde{\bm x}_i, y_i, z_i))^\top \cdot v= m_{\beta}(\tilde{\bm x}_i, y_i))^\top \cdot [(1-\bm z_i) \odot v]$, so $n_{\beta}(\tilde{\bm x}_i, y_i, z_i))$ is also a sub-gaussian vector. Similarly, we also have $u_i$ be sub-gaussian random variables. So by the conclusion from Lemma~\ref{lem:phdmrm2}, we could conclude that both the term $(\ref{peq37})$, the term $(\ref{peq38})$ and the term $(\ref{peq39})$ are all $O(\sqrt{\frac{s^* \log d}{n}} \cdot \log n)$ with probability 
$1-  m_0/(\log n \cdot \log d)$, therefore, for any $t = 0,1,2,...,N_0-1$, we have: 
\begin{equation}
    \Pro (||\nabla Q_{n/N_0} (\bm \beta^t;\bm \beta^t) - f_T(\nabla Q_{n/N_0} (\bm \beta^t;\bm \beta^t))||_2 > C\cdot\sqrt{\frac{s^* \log d}{n}}\cdot \log n) \le C' \cdot 1/{\log d \cdot \log n}, 
\end{equation}
which completes the proof of Lemma~\ref{lem:hdmcm}. \hfill$\square$
\subsection{Proof of Lemma~\ref{lem:phdem1}}\label{plem:phdem1}
By  assumption (\ref{peq1}), we have 
\begin{equation}
    (1-L)\|\bm \beta^*\|_2\leq \|\bar{\bm \beta}^{t+0.5}\|_2\leq (1+L)\|\bm \beta^*\|_2.
\end{equation}
Then, for the simplicity of the proof, we can denote that  
\begin{equation*}
    \bar{\bm \theta}=\frac{\bar{\bm \beta}^{t+0.5}}{\|\bar{\bm \beta}^{t+0.5}\|_2}, \bm \theta=\frac{\bm \beta^{t+0.5}}{\|\bar{\bm \beta}^{t+0.5}\|_2} \text{, and } \bm \theta^*= \frac{\bm \beta^*}{\|\bm \beta^*\|_2}.
\end{equation*}
We can find that both $\bar{\bm \theta}$ and $\bm \theta^*$ are unit vectors. Further, we denote the sets $\mathcal{I}_1, \mathcal{I}_2$ and $\mathcal{I}_3$ as the follows
\begin{equation*}
    \mathcal{I}_1=\mathcal{S}^*\backslash {\mathcal{S}}^{t+0.5}, \mathcal{I}_2=\mathcal{S}^*\bigcap {\mathcal{S}}^{t+0.5} \text{, and }   \mathcal{I}_3={\mathcal{S}}^{t+0.5} \backslash \mathcal{S}^*,
\end{equation*}
where $\mathcal{S}^*=\text{supp}(\bm \beta^*)$, the support of $\bm \beta^*$. And ${\mathcal{S}}^{t+0.5}$ be the set of indexes chosen by the NoisyHT algorithm during the $t$-th iteration. 
Let $s_i=|\mathcal{I}_i|$ for $i=1, 2, 3$, respectively. Then, we define $\Delta=\langle \bar{\bm \theta}, \bm \theta^*\rangle$, so the results below holds:
\begin{equation}\label{peq2}
    \Delta=\langle \bar{\bm \theta}, \bm \theta^*\rangle= \sum_{j\in \mathcal{S}^*}\bar{\bm \theta}_j\theta_j^*=\sum_{j\in \mathcal{I}_1}\bar{\bm \theta}_j\theta_j^*+\sum_{j\in \mathcal{I}_2}\bar{\bm \theta}_j\theta_j^*\leq \|\bar{\bm \theta}_{\mathcal{I}_1}\|_2\|\bm \theta^*_{\mathcal{I}_1}\|_2+\|\bar{\bm \theta}_{\mathcal{I}_2}\|_2\|\bm \theta^*_{\mathcal{I}_2}\|_2.
\end{equation}
By Cauchy-Schwartz inequality, we  have 
\begin{align}\label{peq3}
    \Delta^2&\leq (\|\bar{\bm \theta}_{\mathcal{I}_1}\|_2\|\bm \theta^*_{\mathcal{I}_1}\|_2+\|\bar{\bm \theta}_{\mathcal{I}_2}\|_2\|\bm \theta^*_{\mathcal{I}_2}\|_2)^2 \nonumber 
    \leq (\|\bar{\bm \theta}_{\mathcal{I}_1}\|_2^2+\|\bar{\bm \theta}_{\mathcal{I}_2}\|_2^2)(\|\bm \theta^*_{\mathcal{I}_1}\|_2^2+\|\bm \theta^*_{\mathcal{I}_2}\|_2^2)\nonumber\\
    &= (1-\|\bar{\bm \theta}_{\mathcal{I}_3}\|_2^2)(1-\|\bm \theta^*_{\mathcal{I}_3}\|_2^2)\leq 1-\|\bar{\bm \theta}_{\mathcal{I}_3}\|_2^2.
\end{align}
Then, let $R$ be a set of indexes for the smallest $s_1$ indexes in $\mathcal{I}_3$, this is possible, since by the assumption, we have $\hat{s} > s^*$, thus $|\mathcal{I}_1| < |\mathcal{I}_3|$. Then, by Lemma~\ref{lem:noisyht2}, define $\bm W = \sum_{i \in [\hat{s}]} ||\bm w_i||_{\infty}^2$, we have for any $c_0 > 0$:
\begin{equation}
||\bm \beta_{\mathcal{I}_1}^{t+0.5}||_2^2 \le (1+c_0) ||\bm \beta_{R}^{t+0.5}||_2^2+4(1+\frac{1}{c_0})\bm W  .
\end{equation}
Then by the fact that $\frac{||\bm \beta_{\mathcal{I}_3}^{t+0.5}||_2}{\sqrt{s_3}} \ge \frac{||\bm \beta_{R}^{t+0.5}||_2}{\sqrt{s_1}}$ according to the choice of $R$, and also a standard inequality $a^2 +b^2 \le (a+b)^2$ when $a,b>0$.
Thus, we have:
\begin{equation}
    \frac{||\bm \beta_{\mathcal{I}_1}^{t+0.5}||_2}{\sqrt{s_1}} \le \sqrt{1+c_0}\cdot \frac{||\bm \beta_{\mathcal{I}_3}^{t+0.5}||_2}{\sqrt{s_3}}+2\sqrt{(1+\frac{1}{c_0})\bm W  }/{\sqrt{s_1}}.
\end{equation}
Therefore,
\begin{equation}\label{peq4}
    \frac{||\bm \theta_{\mathcal{I}_1}^{t+0.5}||_2}{\sqrt{1+c_0}\cdot\sqrt{s_1}} \le  \frac{||\bm \theta_{\mathcal{I}_3}^{t+0.5}||_2}{\sqrt{s_3}}+2\sqrt{(1+\frac{1}{c_0})\bm W  }/{(\sqrt{1+c_0}\cdot\sqrt{s_1}\cdot||\bar{\bm \beta}^{t+0.5}||_2)}.
\end{equation}
Then, we denote $\bm W_1 = 2\sqrt{(1+\frac{1}{c_0})\bm W  }/{(\sqrt{1+c_0}\cdot\sqrt{s_1}\cdot||\bar{\bm \beta}^{t+0.5}||_2)}$. Further, we define that 
\begin{equation*}
    \epsilon_0 = 2\eta \alpha /||\bar{\bm \beta}^{t+0.5}||_2 \text{  and  } \epsilon_1 = \eta \cdot ||f_T(\nabla Q_{n_0}(\bm \beta^t ;\bm \beta^t)) - \nabla Q_{n_0}(\bm \beta^t ;\bm \beta^t)||_2/||\bar{\bm \beta}^{t+0.5}||_2. 
\end{equation*}
Thus we have, with probability $1-\tau-N_0 \cdot \phi(\xi)$:
\begin{align}
    \frac{\|\bm \theta_{\mathcal{I}_1}-\bar{\bm \theta}_{\mathcal{I}_1}\|_2 }{\sqrt{s_1}} 
    &\le \frac{\|\bm \beta_{\mathcal{I}_1}^{t+0.5}-\bar{\bm \beta}_{\mathcal{I}_1}^{t+0.5}\|_2 }{\sqrt{s_1}\cdot||\bar{\bm \beta}^{t+0.5}||_2} \notag\\ 
    &\le \frac{\|[\eta\nabla Q(\bm \beta^t;\bm \beta^t) -\eta f_T(\nabla Q_{n_0} (\bm \beta^t;\bm \beta^t))] _{\mathcal{I}_1}\|_2}{\sqrt{s_1}\cdot||\bar{\bm \beta}^{t+0.5}||_2} \notag\\
    &\le \frac{\eta\|[\nabla Q(\bm \beta^t;\bm \beta^t) - \nabla Q_{n_0} (\bm \beta^t;\bm \beta^t)] _{\mathcal{I}_1}\|_2}{\sqrt{s_1}\cdot||\bar{\bm \beta}^{t+0.5}||_2} +\frac{\eta\|[\nabla Q_{n_0}(\bm \beta^t;\bm \beta^t) - f_T(\nabla Q_{n_0} (\bm \beta^t;\bm \beta^t))] _{\mathcal{I}_1}\|_2}{\sqrt{s_1}\cdot||\bar{\bm \beta}^{t+0.5}||_2} \notag\\
    &\le \frac{\eta\sqrt{s_1}\|\nabla Q(\bm \beta^t;\bm \beta^t) - \nabla Q_{n_0} (\bm \beta^t;\bm \beta^t)\|_\infty}{\sqrt{s_1}\cdot||\bar{\bm \beta}^{t+0.5}||_2} +\frac{\eta\|[\nabla Q_{n_0}(\bm \beta^t;\bm \beta^t) - f_T(\nabla Q_{n_0} (\bm \beta^t;\bm \beta^t))] _{\mathcal{I}_1}\|_2}{\sqrt{s_1}\cdot||\bar{\bm \beta}^{t+0.5}||_2} \notag\\
    &\le \epsilon_0/2 + \epsilon_1/\sqrt{s_1}. \notag
\end{align}
Similarly, we also have that:
\begin{equation*}
     \frac{\|\bm \theta_{\mathcal{I}_3}-\bar{\bm \theta}_{\mathcal{I}_3}\|_2 }{\sqrt{s_3}} \le \epsilon_0/2 + \epsilon_1/\sqrt{s_3}.
\end{equation*}
Define $\tilde{\epsilon} = \bm W_1 + (1+\frac{1}{\sqrt{1+c_0}})\epsilon_0/2 + (\frac{1}{\sqrt{s_3}}+\frac{1}{\sqrt{s_1}\cdot\sqrt{1+c_0}})\epsilon_1$, which implies that 
\begin{align}\label{peq5}
\frac{\|\bar{\bm \theta}_{\mathcal{I}_3}\|_2}{\sqrt{s_3}}\geq \frac{\|\bm \theta_{\mathcal{I}_3}\|_2}{\sqrt{s_3}}-\frac{\|\bm \theta_{\mathcal{I}_3}-\bar{\bm \theta}_{\mathcal{I}_3}\|_2 }{\sqrt{s_3}}
& \geq \frac{\|\bm \theta_{\mathcal{I}_1}\|_2}{\sqrt{1+c_0}\cdot\sqrt{s_1}}-\bm W_1-\frac{\|\bm \theta_{\mathcal{I}_3}-\bar{\bm \theta}_{\mathcal{I}_3}\|_2 }{\sqrt{s_3}}\nonumber \\
&\geq \frac{\|\bar{\bm \theta}_{\mathcal{I}_1}\|_2}{\sqrt{1+c_0}\cdot\sqrt{s_1}}-\bm W_1-\frac{\|\bm \theta_{\mathcal{I}_3}-\bar{\bm \theta}_{\mathcal{I}_3}\|_2 }{\sqrt{s_3}}- \frac{\|\bm \theta_{\mathcal{I}_1}-\bar{\bm \theta}_{\mathcal{I}_1}\|_2 }{\sqrt{1+c_0}\cdot\sqrt{s_1}} \notag\\
&\geq \frac{\|\bar{\bm \theta}_{\mathcal{I}_1}\|_2}{\sqrt{1+c_0}\cdot\sqrt{s_1}}-\tilde{\epsilon},
\end{align}
where the second inequality is by plugging (\ref{peq4}) into the inequality.
Plugging (\ref{peq5}) into (\ref{peq3}), we have 
\begin{equation}\label{peq6}
    \Delta^2\leq 1-\|\bar{\bm \theta}_{\mathcal{I}_3}\|_2^2 \leq 1-(\sqrt{\frac{s_3}{s_1 \cdot (1+c_0)}}\|\bar{\bm \theta}_{\mathcal{I}_1}\|_2-\sqrt{s_3}\tilde{\epsilon})^2.
\end{equation}
After solving $\|\bar{\theta}_{\mathcal{I}_1}\|_2$ in (\ref{peq6}), we can obtain the inequality below:  
\begin{equation}\label{peq7}
    \|\bar{\bm \theta}_{\mathcal{I}_1}\|_2\leq \sqrt{\frac{s_1 \cdot (1+c_0)}{s_3}}\sqrt{1-\Delta^2}+\sqrt{s_1 \cdot (1+c_0)}\cdot \tilde{\epsilon}\leq \sqrt{\frac{s^*}{\hat{s}}}\cdot\sqrt{1-\Delta^2}+\sqrt{1+c_0}\cdot\sqrt{s_1}\cdot\tilde{\epsilon}.
\end{equation}
The final inequality is due to the inequality $\frac{s_1}{s_3}\leq \frac{s_1+s_2}{s_3+s_2}=\frac{s^*}{\hat{s}}$. By a properly chosen $c_0$ satisfies that $c_0 \le \min((s^* \cdot s_3)/(\hat{s}\cdot s_1)-1, (2\sqrt{s^*/s_1}-1)^2-1)$, we have  $(1+c_0) \cdot \frac{s_1}{s_3}\leq \frac{s^*}{\hat{s}}$. In the following, we will prove that the right hand side of (\ref{peq7}) is upper bounded by $\Delta$. \\\\
By the assumptions from this lemma, we first observe that $\epsilon_0 = o(1)$ and $\epsilon_1 =o(1)$. We can also find that for $\bm W$, by Lemma A.1 from  \citep{cai2019cost}, we can claim that, there exists constants $c_1, m_0, m_1$, such that, with probability $1-m_0 \cdot s^* \cdot \exp(-m_1\log d)$:
\begin{equation*}
    \bm W \le c_1\cdot \frac{(s^*\log d)^2\log(1/\delta)\log^3 n}{n^2\epsilon^2}.
\end{equation*}
Thus, if we let $\bm W'$ be the maximum of $\bm W$ for all the $N_0$ iterations, then by a union bound, if we let $N_0 = O(\log n)$ be the total number of iterations, then with probability $1-m_0\cdot  s^* \log n \cdot \exp(-m_1\log d)$:
\begin{equation*}
    \bm W' \le c_1\cdot \frac{(s^*\log d)^2\log(1/\delta)\log^3 n}{n^2\epsilon^2} .
\end{equation*}
Therefore, 
\begin{equation}
    \sqrt{s_1} \cdot \bm W_1 \le {c_1'}\cdot \frac{s^*\log d \cdot \log(1/\delta) \log^{3/2} n}{n \epsilon}.
\end{equation}
by the assumptions of this Lemma, we have $\sqrt{s_1}\cdot \bm W_1 =o(1)$, thus for $\tilde{\epsilon}$, we find that:
\begin{align}\label{peq8}
    \sqrt{s_1} \tilde{\epsilon} &\le \sqrt{s_1}\bm W_1 +\sqrt{s_1}  (1+\frac{1}{\sqrt{1+c_0}})\epsilon_0/2 + (\frac{\sqrt{s_1} }{\sqrt{s_3}}+\frac{\sqrt{s_1} }{\sqrt{s_1}\cdot\sqrt{1+c_0}})\epsilon_1 \notag\\
    &\le \sqrt{s_1}\bm W_1 +\sqrt{s_1}  (1+\frac{1}{\sqrt{1+c_0}})\epsilon_0/2 + (1+\frac{1}{\sqrt{1+c_0}})\epsilon_1
\end{align}
The first term and the third term of (\ref{peq8}) are all o(1), plug this result into (\ref{peq7}), we have that,
\begin{equation*}
     \|\bar{\bm \theta}_{\mathcal{I}_1}\|_2\leq  \sqrt{\frac{s^*}{\hat{s}}}\cdot\sqrt{1-\Delta^2}+(\sqrt{1+c_0}+1)\cdot\sqrt{s_1}\epsilon_0/2.
\end{equation*}
By a properly chosen $c_0$, we can attain that $(\sqrt{1+c_0}+1)\cdot\sqrt{s_1} \le 2\sqrt{s^*}$, so,
\begin{equation}\label{peq9}
     \|\bar{\bm \theta}_{\mathcal{I}_1}\|_2\leq \sqrt{\frac{s^*}{\hat{s}}}\cdot\sqrt{1-\Delta^2}+\sqrt{s^*}\epsilon_0.
\end{equation}
In the following steps, we will prove that the right hand side of (\ref{peq8}) is upper bounded by $\Delta$. Such bound holds if we have: 
\begin{align}\label{peq10}
    \Delta\geq \frac{\sqrt{s^*}\epsilon_0+[s^*\tilde{\epsilon}^2-(s^*/\hat{s}+1)(s^*{\epsilon_0^2-s^*/\hat{s})]^{1/2}}}{s^*/\hat{s}+1}=\frac{\sqrt{s^*}{\epsilon_0}
    +[-(s^*\tilde{\epsilon})^2/\hat{s}+(s^*/\hat{s}+1)s^*/\hat{s}]^{1/2}
    }{s^*/\hat{s}+1}.
\end{align}
To prove (\ref{peq10}), we first note that $\sqrt{s^*}{\epsilon_0}\leq \Delta$, which holds because: 
\begin{equation}\label{peq11}
    \sqrt{s^*}{\epsilon_0}\leq \sqrt{\hat{s}}{\epsilon_0}=\frac{2\eta\alpha\sqrt{\hat{s}}}{\|\bar{\bm \beta}^{t+0.5}\|_2}\leq \frac{1-L}{1+L}\leq \Delta,
\end{equation}
where the second inequality is due to the assumptions in the lemma and the final inequality is due to the fact that:
\begin{equation}\label{peq12}
    \|\bar{\bm \beta}^{t+0.5}\|_2^2+\|\bm \beta^*\|_2^2-2\langle \bar{\bm \beta}^{t+0.5}, \bm \beta^* \rangle = \|\bar{\bm \beta}^{t+0.5}-\bm \beta^*\|_2^2 \le L^2\|\bm \beta^*\|_2^2.
\end{equation}
and 
\begin{equation}\label{peq13}
    \Delta=\langle \bar{\bm \theta}, \bm \theta^*\rangle =\frac{\langle \bar{\bm \beta}^{t+0.5}, \bm \beta^*\rangle }{\|\bar{\bm \beta}^{t+0.5}\|_2 \|\bm \beta^*\|_2}\geq \frac{\|\bar{\bm \beta}^{t+0.5}\|_2^2+\|\bm \beta^*\|_2^2-\kappa^2\|\bm \beta^*\|_2^2}{2\|\bar{\bm \beta}^{t+0.5}\|_2\|\bm \beta^*\|_2}\geq \frac{(1-L)^2+1-L^2}{2(1+L)}=\frac{1-L}{1+L}.
\end{equation}
By combining (\ref{peq12}) and (\ref{peq13}), we can finish the proof of (\ref{peq11}). Then,  we can verify that (\ref{peq10}) holds. By (\ref{peq11}), we have 
\begin{equation}
\sqrt{\hat{s}}\cdot{\epsilon_0}\leq \frac{1-L}{1+L}<1<\sqrt{\frac{s^*+\hat{s}}{\hat{s}}},
\end{equation}
The above inequality implies that ${\epsilon_0}\leq \frac{\sqrt{s^*+\hat{s}}}{\hat{s}}$. Then, for the right hand side of (\ref{peq10}), we observe that: 
\begin{align}
    \frac{\sqrt{s^*}{\epsilon_0}
    +[-(s^*{\epsilon_0})^2/\hat{s}+(s^*/\hat{s}+1)s^*/\hat{s}]^{1/2}
    }{s^*/\hat{s}+1}&\leq \frac{\sqrt{s^*}{\epsilon_0}
    +[(s^*/\hat{s}+1)s^*/s]^{1/2}
    }{s^*/\hat{s}+1}\notag\\
    &\leq 2\sqrt{\frac{s^*}{s^*+\hat{s}}} \notag\\
    &\leq 2\sqrt{\frac{1}{1+4(1+L)^2/(1-L)^2}} \notag\\
    &\leq \frac{1-L}{1+L}\leq \Delta.
\end{align}
Thus, we can claim that 
\begin{equation}
    \|\bar{\bm \theta}_{\mathcal{I}_1}\|_2\leq \Delta.
\end{equation}
Furthermore, from (\ref{peq3}), we have: 
\begin{equation*}
     \Delta\leq \|\bar{\bm \theta}_{\mathcal{I}_1}\|_2\|\bm \theta^*_{\mathcal{I}_1}\|_2+\|\bar{\bm \theta}_{\mathcal{I}_2}\|_2\|\bm \theta^*_{\mathcal{I}_2}\|_2 \leq \|\bar{\bm \theta}_{\mathcal{I}_1}\|_2\|\bm \theta^*_{\mathcal{I}_1}\|_2+\sqrt{(1-\|\bar{\bm \theta}_{\mathcal{I}_1}\|_2^2)}\sqrt{(1-\|\bm \theta^*_{\mathcal{I}_1}\|^2_2)},
\end{equation*}
noticing that $\|\bar{\bm \theta}_{\mathcal{I}_1}\|_2\|\bm \theta^*_{\mathcal{I}_1}\|_2 \le \Delta$, thus, 
\begin{equation*}
   (\Delta- \|\bar{\bm \theta}_{\mathcal{I}_1}\|_2\|\bm \theta^*_{\mathcal{I}_1}\|_2)^2 \leq (1-\|\bar{\bm \theta}_{\mathcal{I}_1}\|_2^2)(1-\|\bm \theta^*_{\mathcal{I}_1}\|^2_2).
\end{equation*}
By solving $\|\bm \theta^*_{\mathcal{I}_1}\|_2$ from the above inequality, and by the fact that $\Delta \le 1 $, we have: 
\begin{align}\label{peq14}
    \|\bm \theta^*_{\mathcal{I}_1}\|_2
    &\leq \|\bar{\bm \theta}_{\mathcal{I}_1}\|_2\Delta+\sqrt{1-\|\bar{\bm \theta}_{\mathcal{I}_1}\|_2^2}\sqrt{1-\Delta^2}\leq \|\bar{\bm \theta}_{\mathcal{I}_1}\|_2+\sqrt{1-\Delta^2} \notag \\ 
    &\leq (\sqrt{\frac{s^*}{\hat{s}}}+1)\sqrt{1-\Delta^2}+\sqrt{s^*}{\epsilon_0},
\end{align}
where the third inequality is due to (\ref{peq9}). Therefore, by (\ref{peq9}) and (\ref{peq14}), we can combine the result as follows: 
\begin{equation}\label{peq15}
     \|\bm \theta^*_{\mathcal{I}_1}\|_2  \|\bar{\bm \theta}_{\mathcal{I}_1}\|_2\leq [(\sqrt{\frac{s^*}{\hat{s}}}+1)\sqrt{1-\Delta^2}+\sqrt{s^*}{\epsilon_0}]\cdot [\sqrt{\frac{s^*}{\hat{s}}}\sqrt{1-\Delta^2}+\sqrt{s^*}{\epsilon_0}].
\end{equation}
Note by the definition of $\bar{\bm \theta}$, we can find that: 
\begin{equation}
    \bar{\bm \beta}^{t+1}=\text{trunc}(\bar{\bm \beta}^{t+0.5}, {\mathcal{S}}^{t+0.5})=\text{trunc}(\bar{\bm \theta}, {\mathcal{S}}^{t+0.5})\cdot \| \bar{\bm \beta}^{t+0.5}\|_2.
\end{equation}
Therefore, we have:
\begin{equation}\label{peq16}
    \langle \frac{\bar{\bm \beta}^{t+1}}{\|\bar{\bm \beta}^{t+0.5}\|_2}, \frac{\bm \beta^*}{\|\bm \beta^*\|_2} \rangle= \langle \text{trunc}(\bar{\bm \theta}, {\mathcal{S}}^{t+0.5}), \bm \theta^* \rangle =\langle \bar{\bm \theta}_{\mathcal{I}_2}, \bm \theta^*_{\mathcal{I}_2}\rangle \geq \langle \bar{\bm \theta}, \bm \theta^*\rangle-\|\bar{\bm \theta}_{\mathcal{I}_1}\|_2\|\bm \theta^*_{\mathcal{I}_1}\|_2.
\end{equation}
Define $\chi=\|\bar{\bm \beta}^{t+0.5}\|_2\|\bm \beta^*\|_2$. Combining the results from (\ref{peq16}) and (\ref{peq15}), we observe that:
\begin{align}
   & \langle \bar{\bm \beta}^{t+1}, \bm \beta^*\rangle \nonumber \\
   &\geq \langle \bar{\bm \beta}^{t+0.5}, \bm \beta^*\rangle- [(\sqrt{s^*/
   \hat{s}}+1)\cdot \sqrt{\chi(1-\Delta^2)}+\sqrt{s^*}\cdot\sqrt{\chi}{\epsilon_0}]\cdot [\sqrt{{s^*}/\hat{s}}\cdot\sqrt{\chi(1-\Delta^2)}+\sqrt{s^*}\cdot\sqrt{\chi}{\epsilon_0}]\nonumber\\
   &= \langle \bar{\bm \beta}^{t+0.5}, \bm \beta^*\rangle-(\sqrt{{s^*}/\hat{s}}+{s^*}/\hat{s})\cdot\chi(1-\Delta^2)-(1+2\sqrt{{s^*}/\hat{s}})\cdot\sqrt{\chi(1-\Delta^2)}\sqrt{s^*}\sqrt{\chi}\tilde{\epsilon}-(\sqrt{s^*}\sqrt{\chi}{\epsilon_0})^2 \label{peq17}
\end{align}
Then, note the definition of $\Delta$, we can bound the term $\sqrt{\chi(1-\Delta^2)}$ by:
\begin{align}\label{peq18}
    \sqrt{\chi(1-\Delta^2)}&\leq \sqrt{2\chi(1-\Delta)}\leq \sqrt{2\|\bar{\bm \beta}^{t+0.5}\|_2\|\bm \beta^*\|_2-2\langle \bar{\bm \beta}^{t+0.5},\bm \beta^*\rangle } \notag\\
    &\le \sqrt{\|\bar{\bm \beta}^{t+0.5}\|_2^2+\|\bm \beta^*\|_2^2-2\langle \bar{\bm \beta}^{t+0.5},\bm \beta^*\rangle } \notag \\
    &\leq \|\bar{\bm \beta}^{t+0.5}-\bm \beta^*\|_2. 
\end{align}
Then, for the term $\sqrt{\chi}{\epsilon_0}$, we have
\begin{align}\label{peq19}
    \sqrt{\chi}{\epsilon_0}=\sqrt{\|\bar{\bm \beta}^{t+0.5}\|_2\|\bm \beta^*\|_2}\frac{2\eta\alpha}{\|\bar{\bm \beta}^{t+0.5}\|_2}\leq \frac{2\eta\alpha}{\sqrt{1-L}}.
\end{align}
Then, we can plug (\ref{peq18}) and (\ref{peq19}) into (\ref{peq17}), we can get the following results:  
\begin{align}
    \langle \bar{\bm \beta}^{t+1}, \bm \beta^*\rangle &\geq \langle \bar{\bm \beta}^{t+0.5}, \bm \beta^*\rangle-(\sqrt{{s^*}/\hat{s}}+{s^*}/\hat{s})\|\bar{\bm \beta}^{t+0.5}-\bm \beta^*\|_2^2\notag \\&-
    (1+2\sqrt{{s^*}/\hat{s}})\|\bar{\bm \beta}^{t+0.5}
    -\bm \beta^*\|_2\cdot\frac{\sqrt{s^*}}{\sqrt{1-L}}\cdot2\eta\alpha-\frac{4\eta^2s^*}{1-L}{\alpha}^2.\label{peq20}
\end{align}
Then, notice that $\bar{\bm \beta}^{t+1}$ is obtained by truncating $\bar{\bm \beta}^{t+0.5}$, so we have $\|\bar{\bm \beta}^{t+1}\|_2^2+\|\bm \beta^*\|_2^2\leq \|\bar{\bm \beta}^{t+0.5}\|_2^2+\|\bm \beta^*\|_2^2$, plug this fact into (\ref{peq20}), we get the following results: 
\begin{align}
    \|\bar{\bm \beta}^{t+1}-\bm \beta^*\|_2^2&\leq (1+\sqrt{{s^*}/\hat{s}}+\frac{s^*}{\hat{s}})\|\bar{\bm \beta}^{t+0.5}-\bm \beta^*\|_2^2 +\frac{4\eta^2s^*}{1-L}{\alpha}^2+ (1+2\sqrt{{s^*}/\hat{s}})\|\bar{\bm \beta}^{t+0.5}
    -\bm \beta^*\|_2\frac{\sqrt{s^*}}{\sqrt{1-L}}{2\eta\alpha}
   \nonumber \\
    &\leq (1+2\sqrt{{s^*}/\hat{s}}+2{s^*}/\hat{s})[\|\bar{\bm \beta}^{t+0.5}-\bm \beta^*\|_2+\frac{\sqrt{s^*}}{\sqrt{1-L}}{\eta\alpha}]^2 
    +\frac{4\eta^2s^*}{1-L}{\alpha}^2. \label{peq21}
\end{align}
By taking square root on both sides of (\ref{peq21}) and by the fact that for $a,b>0$, $\sqrt{a^2+b^2} \le a+b$, we can find a constant $K_1$, such that: 
\begin{equation}
     \|\bar{\bm \beta}^{t+1}-\bm \beta^*\|_2\leq (1+4\sqrt{{s^*}/\hat{s}})^{\frac{1}{2}}\|\bar{\bm \beta}^{t+0.5}-\bm \beta^*\|_2+\frac{K_1\sqrt{s^*}}{\sqrt{1-L}}{\eta\alpha}.
\end{equation}
Then, the lemma is proved. \hfill$\square$
\subsection{Proof of Lemma~\ref{lem:phdmrm1}}\label{plem:phdmrm1}
\noindent Since $X$ is a sub-gaussian random variable with mean zero and variance $\sigma^2$, then by the definition of sub-gaussian random variables, we have: 
\begin{equation}
    \Pro(X > t) \le \exp(-\frac{t^2}{2\sigma^2}) \text{  ,  } \Pro(X < -t) \le \exp(-\frac{t^2}{2\sigma^2}).
\end{equation}
Thus, with this tail bound, we can calculate $\E(\Pi_{T}(X)-X)^2)$ directly, suppose the density function of $X$ is $f_x$, then, according to the definition of $\Pi_T$, we have:
\begin{align}\label{peq32}
    \E[(\Pi_{T}(X)-X)]^2
    &= \int_{T}^{\infty} (x-T)^2 f_{x} dx + \int_{-\infty}^{-T} (T+x)^2 f_{x} dx. 
\end{align}
Then, we will analyze these two terms in (\ref{peq32}) separately. 
For the first term, we have:
\begin{align}
    \int_{T}^{\infty} (x-T)^2 f_{x} dx
    &=  \int_{T}^{\infty} \int_{T}^{x} 2(t-T) dt f_{x} dx \notag\\
    &=  \int_{T}^{\infty} 2(t-T)P(x \ge t) dt \notag\\
    &\le 2\cdot \int_{T}^{\infty} (t-T) \exp(-\frac{t^2}{2\sigma^2}) dt \notag\\
    &\le 2\cdot \exp(-\frac{T^2}{2\sigma^2}).
\end{align}
Then, by choosing $T = c\cdot \sigma\sqrt{2\log n}$, we have $ \int_{T}^{\infty} (x-T)^2 f_{x} dx = O(\frac{1}{n})$. By the similar approach, we can also attain that $ \int_{-\infty}^{-T} (x+T)^2 f_{x} dx = O(\frac{1}{n})$. Thus, we claim that $\E(\Pi_{T}(x)-x)^2 = O(\frac{1}{n})$.

For the second part proof of this lemma, we can also calculate that:
\begin{align}\label{peq33}
    \E[(\Pi_{T}(X)-X)]^4
    &= \int_{T}^{\infty} (x-T)^4 f_{x} dx + \int_{-\infty}^{-T} (T+x)^4 f_{x} dx.
\end{align}
Same as the first part, we will also analyze these two terms in (\ref{peq33}) separately. 
For the first term, we have:
\begin{align}
    \int_{T}^{\infty} (x-T)^4 f_{x} dx
    &=  \int_{T}^{\infty} \int_{T}^{x} 4(t-T)^3 dt f_{x} dx \notag\\
    &= \int_{T}^{\infty} \int_t^{\infty} f_x dx 4(t-T)^3 dt \notag\\
    &\le 4  \int_{T}^{\infty} (t-T)^3 \exp(-\frac{t^2}{2\sigma^2}) dt \notag\\
    &\le 4  \int_{T}^{\infty} t^3 \exp(-\frac{t^2}{2\sigma^2}) dt \notag\\
    &= c_0 \cdot t^2 \exp(-\frac{t^2}{2\sigma^2})\Big{|}_T^{\infty} - c_1 \int_{T}^{\infty} t \exp(-\frac{t^2}{2\sigma^2}) dt  \notag\\
    &= c_0 \cdot T^2\exp(-\frac{T^2}{2\sigma^2}) + c_1 \cdot \exp(-\frac{T^2}{2\sigma^2}).
\end{align}
Then, by choosing $T = c\cdot \sigma\sqrt{2\log n}$, we have $ \int_{T}^{\infty} (x-T)^2 f_{x} dx = O(\frac{\log n }{n})$. By the similar approach, we can also attain that $ \int_{-\infty}^{-T} (x+T)^4 f_{x} dx = O(\frac{\log n}{n})$. Thus, we claim that $\E(\Pi_{T}(x)-x)^4 = O(\frac{\log n}{n})$. Thus finishes the proof of this lemma. \hfill$\square$
\subsection{Proof of Lemma~\ref{lem:phdmrm2}}\label{plem:phdmrm2}
First, we could separate the left hand side of (\ref{peq34}) as follows:
\begin{align}\label{peq35}
     &||(y_i \cdot \bm x_i - \Pi_{T}(y_i)\Pi_{T}(\bm x_i))_{\mathcal{S}}||_2 \notag\\
     &\le  ||(y_i \cdot \Pi_{T}(\bm x_i) - \Pi_{T}(y_i)\Pi_{T}(\bm x_i))_{\mathcal{S}}||_2 + ||(y_i \cdot \bm x_i - y_i\Pi_{T}(\bm x_i))_{\mathcal{S}}||_2 \notag\\
     &\le \underbrace{T\cdot \sqrt{s}\cdot|\Pi_{T}(y_i)-y_i|}_{(\ref{peq35}.1)} +\underbrace{||((\Pi_{T}(y_i)-y_i)(\Pi_{T}(\bm x_i)-\bm x_i))_{\mathcal{S}}||_2}_{(\ref{peq35}.2)} + \underbrace{T \cdot ||(\Pi_{T}(\bm x_i)-\bm x_i)_{\mathcal{S}}||_2}_{(\ref{peq35}.3)}.
\end{align}
For the term $(\ref{peq35}.1)$, denote the distribution of $y$ as $f_y$, then $y \sim N(0, \beta^\top \beta+\sigma^2)$, which follows a Gaussian distribution.Then by Lemma~\ref{lem:phdmrm1} we have, if we choose $T \asymp \sqrt{\log n}$, we have $\E[\Pi_{T}(\bm y_i)-\bm y_i]^2 = O(\frac{1}{n})$, thus $T^2 \cdot s\cdot \E[\Pi_{T}(\bm y_i)-\bm y_i]^2 = O(\frac{s \cdot \log n}{n})$.

Then, let us analyze on the term $(\ref{peq35}.3)$, we have for any $j \in \mathcal{S}$:
\begin{align}
    \E||(\Pi_{T}(\bm x_i)-\bm x_i)_{\mathcal{S}}||_2^2
    &= s\cdot \E(\Pi_{T}(\bm x_{ij})-\bm x_{ij})^2. 
\end{align}
By Lemma~\ref{lem:phdmrm1}, we have that when we choose $T \asymp \sqrt{\log n}$,  $\E(\Pi_{T}(\bm x_{ij})-\bm x_{ij})^2 = O(\frac{1}{n})$, which means that the term $ \E||(\Pi_{T}(\bm x_i)-\bm x_i)_{\mathcal{S}}||_2^2 = O(\frac{s}{n})$. Therefore, for the term $(\ref{peq35}.3)$ $T^2 \cdot \E||(\Pi_{T}(\bm x_i)-\bm x_i)_{\mathcal{S}_1}||_2^2= O(\frac{s \log n }{n})$.

Finally, let us analyze the term $(\ref{peq35}.2)$, for any $j \in \mathcal{S}$:
\begin{align}\label{peq36}
    \E||((\Pi_{T}(y_i)-y_i)(\Pi_{T}(\bm x_i)-\bm x_i))_{\mathcal{S}}||_2^2 
    &= s\cdot \E(\Pi_{T}(y_i)-y_i)^2(\Pi_{T}(\bm x_{ij})-\bm x_{ij})^2 \notag\\
    &\le s\cdot \sqrt{\E(\Pi_{T}(y_i)-y_i)^4 \cdot \E(\Pi_{T}(\bm x_{ij})-\bm x_{ij})^4}.
\end{align}
Then again, by Lemma~\ref{lem:phdmrm1}, we can also obtain that both $\E(\Pi_{T}(y_i)-y_i)^4$ and $ \E(\Pi_{T}(\bm x_{ij})-\bm x_{ij})^4$ are $O(\frac{\log n }{n})$, insert this analysis into (\ref{peq36}), we can claim that $\E||((\Pi_{T}(y_i)-y_i)(\Pi_{T}(\bm x_i)-\bm x_i))_{\mathcal{S}^{t+0.5}}||_2^2 = O(\frac{s \log n}{n})$.

Therefore, if we choose $\xi = O(\sqrt{\frac{s \log d}{n}} \cdot \log n)$, we have:
\begin{align}
    &\Pro(\frac{1}{n_0}\sum_{i=1}^{n_0} ||(y_i \cdot \bm x_i - \Pi_{T}(y_i)\Pi_{T}(\bm x_i))_{\mathcal{S}_1}||_2 > \xi/2) \notag\\
    &\le \Pro(\frac{1}{n_0}\sum_{i=1}^{n_0} [(\ref{peq35}.1)+(\ref{peq35}.2)+(\ref{peq35}.3)] > \xi/2) \notag\\
    &\le \Pro(\frac{1}{n_0}\sum_{i=1}^{n_0} [(\ref{peq35}.1)] > \xi/6)+\Pro(\frac{1}{n_0}\sum_{i=1}^{n_0} [(\ref{peq35}.2)] > \xi/6)+\Pro(\frac{1}{n_0}\sum_{i=1}^{n_0} [(\ref{peq35}.3)] > \xi/6) \notag\\
    &\le \frac{36 \E[(\ref{peq35}.1)^2]}{\xi^2}+\frac{36 \E[(\ref{peq35}.2)^2]}{\xi^2}+\frac{36 \E[(\ref{peq35}.3)^2]}{\xi^2} \notag\\
    &= O(\frac{1}{\log n \cdot \log d})
\end{align}
Thus finishes the proof. \hfill$\square$




\end{document}